\documentclass{article}

\PassOptionsToPackage{numbers, compress}{natbib}

\usepackage[preprint]{neurips_2026}
\PassOptionsToPackage{numbers,sort&compress}{natbib}
\usepackage{neurips_2026}
\usepackage{graphicx}
\usepackage[utf8]{inputenc} 
\usepackage[T1]{fontenc}    
\usepackage{hyperref}       
\usepackage{url}            
\usepackage{booktabs}       
\usepackage{amsfonts}       
\usepackage{nicefrac}       
\usepackage{microtype}      
\usepackage{xcolor}         
\usepackage{booktabs}
\usepackage{tabularx}
\usepackage{array}
\usepackage{makecell}
\usepackage{booktabs}
\usepackage{multirow}
\usepackage{makecell}
\usepackage[table]{xcolor}
\usepackage{array}
\usepackage[most]{tcolorbox}
\usepackage{xcolor}
\usepackage{graphicx}
\usepackage{subcaption}

\definecolor{lightpurple}{RGB}{245,240,255}
\definecolor{borderpurple}{RGB}{170,150,220}

\newtcolorbox{insightbox}{
  colback=lightpurple,
  colframe=borderpurple,
  boxrule=0.8pt,
  arc=2.5mm,
  left=1.2mm,
  right=1.2mm,
  top=1mm,
  bottom=1mm,
  enhanced
}

\newcolumntype{Y}{>{\raggedright\arraybackslash}X}

\title{EdgeFlowerTune: Evaluating Federated LLM Fine-Tuning Under Realistic Edge System Constraints}

\author{%
  $^\dagger$Jiaxiang Geng$^{1}$,
  $^\dagger$Yiyi Lu$^{1}$,
  $^\dagger$Lunyu Zhao$^{1}$,
  Yan Gao$^{2,3}$,
  Nicholas D. Lane$^{2,3}$,
  Bing Luo$^{1}$ \\
  $^{1}$Duke Kunshan University \\
  $^{2}$Flower Labs \\
  $^{3}$University of Cambridge
}

\begin{document}

\maketitle

\begingroup
\renewcommand{\thefootnote}{\fnsymbol{footnote}}
\footnotetext[2]{These authors contributed equally to this work.}
\endgroup

\begin{abstract}
Federated fine-tuning offers a promising paradigm for adapting large language models (LLMs) on edge devices by leveraging the rich, diverse, and continuously generated data from smartphones and IoT devices without compromising user data privacy. Such edge-side adaptation can improve model personalization, robustness, and responsiveness to local contexts. However, the practical feasibility of federated LLM fine-tuning on real edge devices remains unclear, as most existing work focuses on cross-silo or simulation-based settings, overlooking the resource and runtime constraints that determine whether a method is deployable on real edge systems.
We present \textbf{EdgeFlowerTune}, a deployment-oriented benchmark for federated LLM fine-tuning under realistic edge-system constraints. EdgeFlowerTune jointly evaluates model quality and system costs, including communication, wall-clock latency, memory usage, energy consumption, and robustness to dynamic edge conditions. To compare methods in terms of effectiveness, efficiency, and robustness, EdgeFlowerTune introduces three complementary protocols: \textit{Quality-under-Budget}, \textit{Cost-to-Target}, and \textit{Robustness}. We instantiate EdgeFlowerTune as a real-device platform built on Flower and MobileFineTuner, spanning commercial Android smartphones and NVIDIA edge development boards. Our benchmark results show that accuracy-only evaluation can lead to misleading conclusions: methods with similar final quality may differ substantially in deployability once realistic system constraints are considered. EdgeFlowerTune provides a reproducible benchmark for system-aware evaluation of federated LLM fine-tuning at the edge.
\end{abstract}

\vspace{-2mm}
\section{Introduction}
\vspace{-2mm}

In recent years, large language models (LLMs) have achieved remarkable performance across a wide range of language tasks and have become the dominant foundation for downstream adaptation and deployment \cite{achiam2023gpt4}. The scaling of these models faces a critical bottleneck: high-quality public datasets are projected to be exhausted between 2026 and 2032 \cite{10.5555/3692070.3694094}. Fortunately, vast amounts of valuable data are continuously generated on edge devices, such as smartphones and Internet-of-Things (IoT) devices \cite{10.1145/3487552.3487863}. However, directly collecting data from these edge devices for centralized training may expose sensitive personal information, leading to privacy risks and potentially violating data protection regulations such as the General Data Protection Regulation (GDPR) \cite{GDPR2016a}.

Federated learning (FL) offers a natural solution by enabling multiple clients to collaboratively train or adapt a shared model while keeping training data local \cite{kairouz2021advances,mcmahan2017communication}. This paradigm preserves edge data on local devices without sharing raw data with the server, while still allowing such data to contribute to model training. Building on this idea, recent works have begun to extend FL to LLM fine-tuning on edge devices \cite{cho-etal-2024-heterogeneous,wang2024flora}, commonly referred to as federated LLM fine-tuning. However, compared with traditional federated learning, federated LLM fine-tuning imposes substantially greater burdens on edge devices due to the much larger model size, leading to significantly higher demands on computation, communication, memory, and energy. These system costs have become a major bottleneck to the development of federated LLM fine-tuning on edge devices and constitute a central challenge faced by existing research efforts \cite{10944288}.

To facilitate the comparison of different federated LLM fine-tuning methods, OpenFedLLM and FederatedScope-LLM provide end-to-end frameworks and benchmarks for federated LLM fine-tuning \cite{ye2024openfedllm,kuang2024federatedscope}, while FlowerTune further expands the benchmarking landscape with a cross-domain leaderboard spanning general NLP, finance, medical, and code tasks \cite{gao2025flowertune}. However, these existing frameworks and benchmarks mainly focus on cross-silo settings and emphasize final task quality, such as accuracy, while largely overlooking the system constraints imposed by real edge environments. Considering federated LLM fine-tuning in edge environments raises several new challenges: \textit{(1)} A method may achieve strong final performance but still be impractical for edge deployment because its consumption of computation, communication, memory, or energy exceeds what edge devices can sustain. \textit{(2)} When comparing two methods, it remains unclear how to trade off higher final accuracy against better system efficiency. \textit{(3)} Edge environments are inherently dynamic: client dropout, hardware heterogeneity, and straggler effects may substantially affect the performance of a method during training.


To overcome these challenges, we present \textbf{EdgeFlowerTune}, a benchmark for federated fine-tuning of LLMs under realistic edge-system constraints. Rather than focusing only on which method achieves the highest final task score, EdgeFlowerTune evaluates the effectiveness, efficiency, and robustness of different methods under realistic edge deployment conditions. Furthermore, to enable credible system-aware evaluation, we instantiate EdgeFlowerTune as a real-device federated LLM fine-tuning platform built on top of \textit{Flower} and \textit{MobileFineTuner}, spanning heterogeneous edge hardware including mainstream commercial smartphones and NVIDIA edge computing development boards.

\begin{figure*}
\centering
\includegraphics[width=0.85\textwidth]{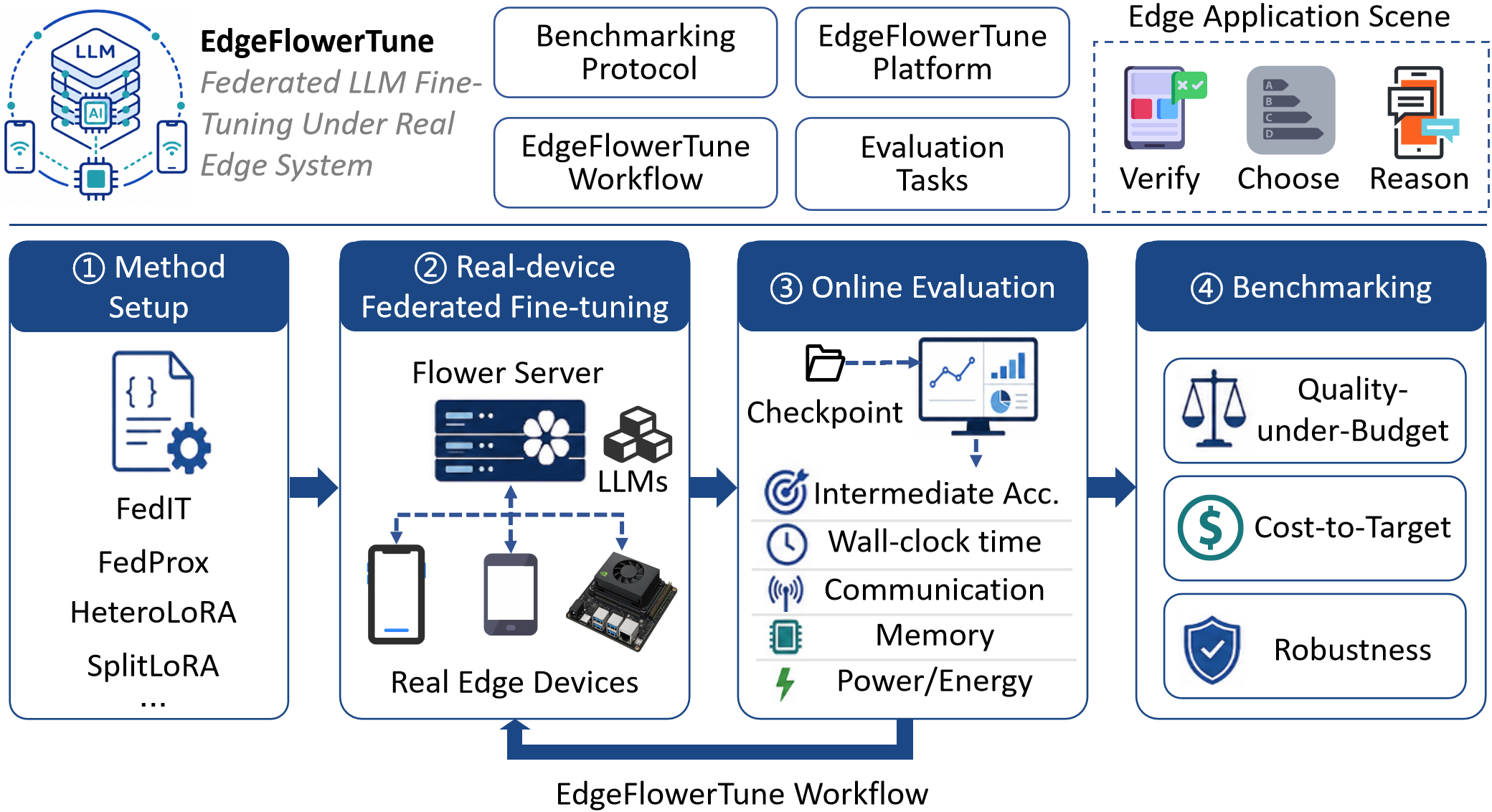}
\caption{Overview of EdgeFlowerTune. Candidate federated LLM fine-tuning methods are deployed and executed on a real-device edge platform, and monitored online for both model quality and system costs. The collected metrics are then evaluated by proposed benchmarking protocols.}
\label{fig:workflow}
\vspace{-6mm}
\end{figure*}

Our contributions are summarized as follows:

\begin{itemize}
    \item We present \textbf{EdgeFlowerTune}, a benchmark for federated fine-tuning of large language models under realistic edge-system constraints. Unlike existing federated LLM fine-tuning benchmarks that mainly emphasize final task quality, EdgeFlowerTune centers evaluation on deployability in terms of effectiveness, efficiency, and robustness. Our code is open-sourced at \url{https://github.com/Edge-Intelligence-Lab/EdgeFlowerTune}.

    \item We propose an end-to-end evaluation workflow and introduce three complementary deployment-oriented evaluation protocols, namely \textit{Quality-under-Budget}, \textit{Cost-to-Target}, and \textit{Robustness}, to systematically evaluate federated LLM fine-tuning methods under explicit constraints on communication, latency, memory, energy, and system variability.

    \item We instantiate EdgeFlowerTune as a real-device federated LLM fine-tuning evaluation platform built on top of \textit{Flower} and \textit{MobileFineTuner}, spanning heterogeneous edge hardware including commercial smartphones and NVIDIA edge computing development boards, thereby enabling credible system-aware benchmarking beyond simulation-only or server-only evaluation.

    \item We conduct benchmark studies on representative federated LLM fine-tuning methods and show that system-aware evaluation can substantially change method conclusions: methods that appear favorable under accuracy-only evaluation may become suboptimal once realistic edge constraints and system perturbations are taken into account.
\end{itemize}

\vspace{-2mm}
\section{EdgeFlowerTune Benchmark Design}
\label{sec:overview}
\vspace{-2mm}

In this section, we present the benchmark design of EdgeFlowerTune. 
We first introduce the end-to-end evaluation workflow, which describes how a candidate federated LLM fine-tuning method is deployed, executed on real edge devices, monitored during training, and finally evaluated by the benchmark. 
We then define three deployment-oriented benchmarking protocols for comparing different methods under realistic edge-system constraints. 

\vspace{-2mm}
\subsection{End-to-End Evaluation Workflow}
\vspace{-2mm}

Figure~\ref{fig:workflow} illustrates the end-to-end evaluation workflow of EdgeFlowerTune. 
It has three key characteristics compared to existing cross-silo benchmarks. 
First, each candidate method must be explicitly deployed into the EdgeFlowerTune benchmark environment, so that its client-side training logic, server-side coordination, communication pattern, and fine-tuning configuration can be executed in a unified evaluation stack. 
Second, fine-tuning is performed on a real heterogeneous edge system, where commercial mobile devices and edge computing boards participate as federated clients. 
Third, because system behavior evolves during training, EdgeFlowerTune records model-quality and system-cost metrics online throughout the fine-tuning process, instead of estimating them after training from offline measurements. 
The collected execution traces are then used by the benchmark protocols to compare different methods in terms of effectiveness, efficiency, and robustness. The stages of the workflow are listed as follows:

\noindent\textbf{Stage 1: Method deployment.}
A candidate federated LLM fine-tuning method is first instantiated in the EdgeFlowerTune benchmark environment. 
The method defines its local training procedure, server-side aggregation rule, exchanged objects between clients and server, communication schedule, and fine-tuning configuration. 
To ensure fair comparison, all methods are deployed under the same benchmark specification, including the same task, data partition, base model, client pool, training budget, and system constraints. 

\noindent\textbf{Stage 2: Real-device federated fine-tuning.}
The deployed method is then executed on the real EdgeFlowerTune system. 
In each communication round, the server coordinates client selection, distributes the required model states or fine-tuning parameters, collects method-specific updates, and performs aggregation. 
Selected clients perform local LLM fine-tuning on their private data using the prescribed method-specific procedure. 
Unlike server-only or simulation-based evaluation, this stage exposes each method to practical edge-system conditions, including heterogeneous device capabilities, real communication links, and runtime resource limitations.

\noindent\textbf{Stage 3: Online metric collection.}
During federated fine-tuning, EdgeFlowerTune continuously monitors both model-quality and system-cost metrics. 
Model-quality metrics include task-specific scores such as loss and accuracy. 
System-cost metrics include communication volume, client-side memory usage, energy consumption and wall-clock time. 
These metrics are collected online during actual benchmark execution.

\noindent\textbf{Stage 4: Protocol-based benchmarking.}
After execution, the collected quality and system traces are passed to the benchmark evaluation module. 
EdgeFlowerTune compares methods using deployment-oriented protocols (illustrated in Sec.~\ref{sec:protocols}) rather than only final task quality. 

\begin{figure*}
\centering
\includegraphics[width=0.85\textwidth]{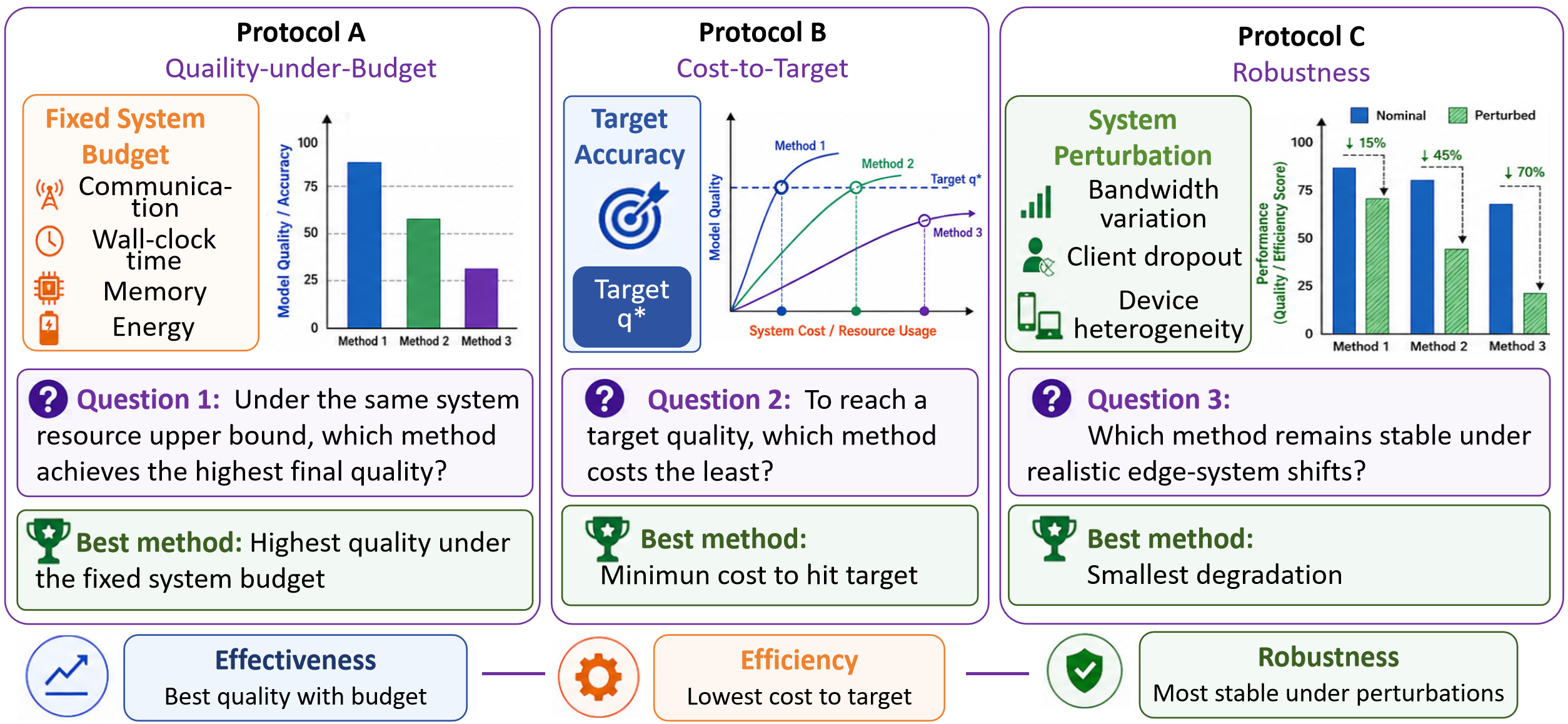}
\caption{EdgeFlowerTune benchmarking protocols. Protocol A evaluates the best model quality achievable under fixed system budgets; Protocol B measures the system cost required to reach a target quality; and Protocol C quantifies robustness under edge-system perturbations.}
\label{fig:protocol}
\vspace{-6mm}
\end{figure*}

\vspace{-2mm}
\subsection{EdgeFlowerTune Benchmarking Protocols}
\label{sec:protocols}
\vspace{-2mm}

Figure~\ref{fig:protocol} summarizes the three benchmarking protocols in EdgeFlowerTune. 
The goal of these protocols is to compare federated LLM fine-tuning methods from a deployment-oriented perspective. 
In realistic edge environments, a method is not necessarily better simply because it achieves a higher final accuracy or lower perplexity. 
It may require excessive communication, exceed the memory capacity of mobile devices, consume too much energy, or become unstable under network and device variability. 
Therefore, EdgeFlowerTune evaluates each method along two coupled dimensions: model quality and system cost. 
Model quality is measured using task-specific metrics, such as loss and accuracy. 
System cost is measured using real execution metrics, including communication volume, wall-clock time, client-side memory usage, power consumption, and energy consumption.

Based on these measurements, EdgeFlowerTune defines three complementary protocols. 
Each protocol corresponds to a different deployment question and produces a different notion of the ``best'' method.

\textbf{Protocol A: Quality-under-Budget.}  
This protocol evaluates method effectiveness under fixed system resource budgets. 
Given the same upper bounds on communication volume, wall-clock time, memory usage, and energy consumption, each candidate method is executed within the allowed budget, and EdgeFlowerTune measures the best model quality it can achieve. 
This protocol answers the question: \textit{under the same system resource upper bound, which method achieves the highest final quality?} 
The best method under this protocol is the one that delivers the highest task quality without violating the specified system budget. 

\textbf{Protocol B: Cost-to-Target.}  
This protocol evaluates method efficiency for reaching a required quality level. 
Instead of fixing the resource budget, EdgeFlowerTune specifies a target quality threshold, denoted as $q^\ast$, and measures the system cost required for each method to reach this target. 
The cost can be reported along multiple dimensions, including total communication volume, training latency, memory footprint, and energy consumption. 
This protocol answers the question: \textit{to reach a target quality, which method costs the least?} 
The best method under this protocol is the one that reaches $q^\ast$ with the minimum system cost. 

\textbf{Protocol C: Robustness.}  
This protocol evaluates method stability under realistic edge-system perturbations. 
EdgeFlowerTune compares each method under nominal conditions and perturbed conditions, including bandwidth variation, client dropout, and device heterogeneity. 
For each perturbation, the benchmark measures the degradation in both model quality and system efficiency. 
This protocol answers the question: \textit{which method remains stable under realistic edge-system shifts?} 
The best method under this protocol is the one with the smallest relative degradation from nominal execution to perturbed execution.

\vspace{-2mm}
\section{Experimental Settings}

\vspace{-2mm}
\subsection{EdgeFlowerTune Platform}
\label{sec:platform}
\vspace{-2mm}

\begin{figure*}
\centering
\includegraphics[width=0.85\textwidth]{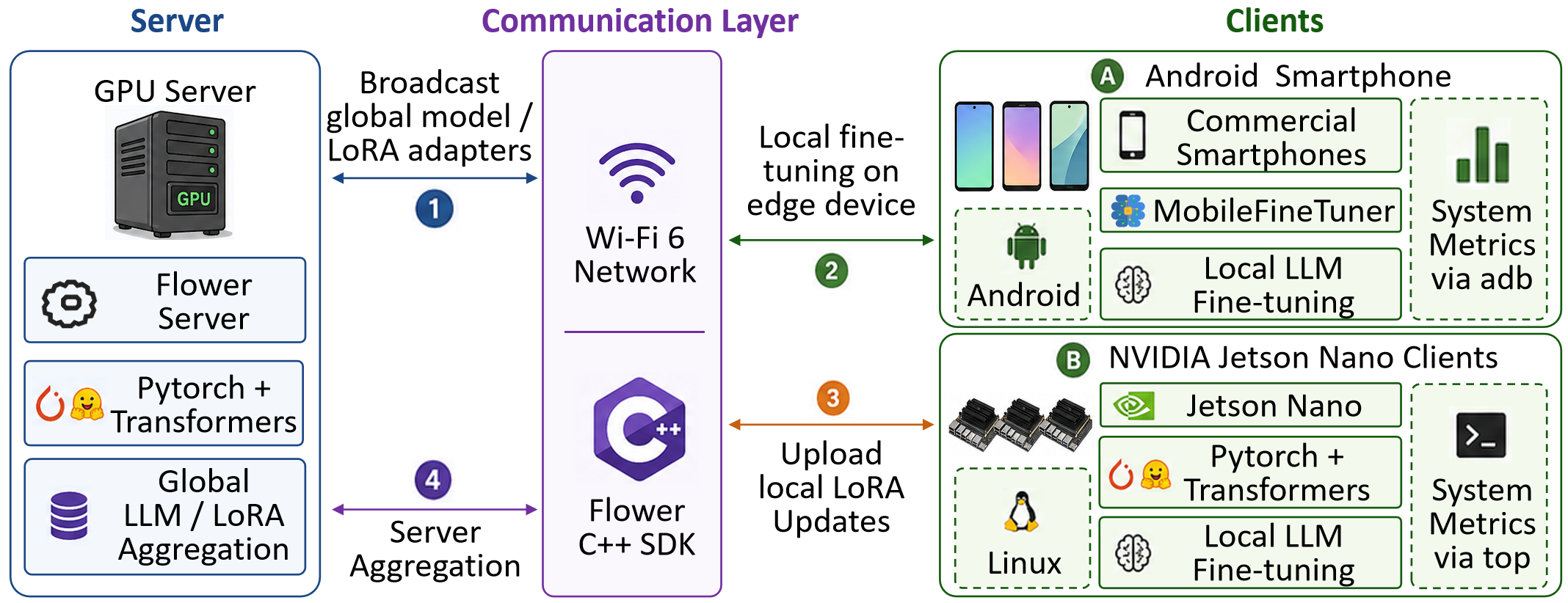}
\caption{EdgeFlowerTune Platform. The platform consists of one gpu server and several real edge devices including android smartphones and NVIDIA boards. }
\label{fig:platform}
\vspace{-4mm}
\end{figure*}

To support system-aware benchmarking, EdgeFlowerTune is built as a real-device federated LLM fine-tuning platform rather than a simulation-only environment. 
As shown in Figure~\ref{fig:platform}, the platform consists of a GPU server, a cross-platform communication layer, and heterogeneous edge clients. 
The server is a Dell PowerEdge T640 equipped with two NVIDIA A800 GPUs. 
It runs \textit{Flower} \cite{gao2025flowertune} for federated orchestration and uses \textit{PyTorch} and \textit{Transformers} \cite{DBLP:journals/corr/abs-1910-03771} for model execution and aggregation.

The client pool spans both Android smartphones and Linux edge computing boards \footnote{We exclude iPhones from our current testbed because iOS aggressively terminates memory-intensive background processes, making long-running on-device LLM fine-tuning unstable.}, as summarized in Table~\ref{tab:client_devices}. 
Android clients perform local LLM fine-tuning with \textit{MobileFineTuner} \cite{geng2025mobilefinetunerunifiedendtoendframework}, a C++ framework for end-to-end LLM adaptation on commercial mobile devices. 
NVIDIA Jetson Orin Nano clients run local training with \textit{PyTorch} and \textit{Transformers}. 

Client-server communication is implemented using the \textit{Flower C++ SDK} over a Wi-Fi 6 network. 
During benchmark execution, the platform records system behavior online. 
Specifically, \textit{adb} is used to collect system metrics from Android smartphones, the Linux \textit{top} command is used for Jetson devices, and \textit{Wireshark} is used to capture communication traffic. 

\begin{table*}[t]
\centering
\caption{Heterogeneous client devices in the EdgeFlowerTune platform. Devices are ordered from faster to slower execution speed in our testbed.}
\vspace{-2mm}
\label{tab:client_devices}
\footnotesize
\setlength{\tabcolsep}{5pt}
\renewcommand{\arraystretch}{1.12}
\resizebox{0.9\textwidth}{!}{
\begin{tabular}{lll}
\toprule
\textbf{Device Model} & \textbf{Processor / CPU} & \textbf{Memory} \\
\midrule
Jetson Orin Nano 
& NVIDIA Orin, 6-core Arm Cortex-A78AE, up to 1.5 GHz 
& 8 GB \\

iQOO 15 
& Snapdragon 8 Elite Gen 5, 2 $\times$ 4.6 GHz + 6 $\times$ 3.62 GHz 
& 16 GB + 16 GB virtual \\

HUAWEI P50 Pro 
& Kirin 9000, 1 $\times$ A77 3.13 GHz + 3 $\times$ A77 2.54 GHz + 4 $\times$ A55 2.05 GHz 
& 8 GB \\

HUAWEI Mate 20 
& Kirin 980, 2 $\times$ A76 2.6 GHz + 2 $\times$ A76 1.92 GHz + 4 $\times$ A55 1.8 GHz 
& 6 GB \\

HUAWEI nova 9 Pro 
& Snapdragon 778G, Kryo 670 CPU, up to 2.4 GHz 
& 8 GB \\
\bottomrule
\end{tabular}
}
\vspace{-6mm}
\end{table*}

\vspace{-2mm}
\subsection{Evaluation Tasks}
\label{sec:evaluation_tasks}

\vspace{-2mm}
\subsubsection{Dataset Selection}
\label{sec:dataset_selection}
\vspace{-2mm}

EdgeFlowerTune evaluates federated LLM fine-tuning on representative edge-oriented language tasks. 
In real edge applications, LLMs often operate over local and context-sensitive information, such as user messages, notifications, schedules, application content, device states, sensor summaries, and event logs. 
These scenarios require not only open-ended generation, but also lightweight verification, contextual selection, and reasoning over local information. 
Accordingly, we organize the evaluation datasets into three categories: \textbf{Verify}, \textbf{Choose}, and \textbf{Reason}.

\textbf{Verify} evaluates whether a model can determine if a fact, condition, or semantic relation holds. 
This category reflects edge scenarios such as checking whether a notification is urgent, whether a sensor event satisfies an alert condition, or whether a context supports a candidate conclusion. 
We instantiate this category with BoolQ~\cite{clark2019boolq} and QNLI~\cite{wang2018glue,rajpurkar2016squad}, which evaluate boolean question answering and question-answer entailment, respectively.

\textbf{Choose} evaluates whether a model can select the most appropriate option from multiple candidates. 
This category reflects scenarios where an edge assistant or local controller needs to choose among candidate replies, recommendations, actions, or explanations based on contextual information. 
We instantiate this category with PIQA~\cite{bisk2020piqa}, HellaSwag~\cite{zellers2019hellaswag}, and SocialIQA~\cite{sap2019socialiqa}, covering physical commonsense, event continuation, and social commonsense.

\textbf{Reason} evaluates whether a model can infer the correct answer from contextual clues and commonsense knowledge. 
This category reflects scenarios where a user asks about local content, a device explains its current state, or reasons over event descriptions. 
We instantiate this category with ARC-E~\cite{clark2018arc} and WinoGrande~\cite{sakaguchi2020winogrande}, which evaluate elementary science question answering and commonsense coreference-style reasoning.

\vspace{-2mm}
\subsubsection{Model selection}
\vspace{-2mm}
Large-scale LLMs with multi-billion parameters are infeasible to run or fine-tune on typical edge devices due to their high computational and memory requirements. Edge tasks, while diverse, generally involve lower complexity than centralized server-side workloads, making extremely large models unnecessary for practical deployment. To reflect realistic constraints while covering a practical range of model capacities, we select three lightweight LLMs representative of current edge-oriented large models: \textit{Gemma 3-270M}, \textit{Gemma 3-1B} \cite{gemma3_2025}, and \textit{Qwen2.5-0.5B} \cite{qwen2p5_2024}. These models are specifically designed for resource-constrained devices, balancing parameter scale with inference and fine-tuning efficiency, enabling evaluation of federated fine-tuning in realistic edge intelligence scenarios.

\vspace{-2mm}
\subsection{Method Selection}
\label{sec:method_selection}
\vspace{-2mm}

To cover the main algorithmic patterns in federated LLM fine-tuning, we select four representative method families: FedAvg + LoRA, FedProx + LoRA, HeteroLoRA, and SplitLoRA. 

\textbf{FedAvg + LoRA.}
This is the standard adapter-only federated fine-tuning baseline. The pretrained LLM is frozen, each client updates only LoRA parameters~\cite{hu2022lora}, and the server aggregates client LoRA updates using FedAvg~\cite{mcmahan2017communication}. This abstraction covers the basic FL-based instruction tuning setting explored by {FedIT}~\cite{zhang2023fedit} and the FedAvg baseline in {OpenFedLLM}~\cite{ye2024openfedllm}.

\textbf{FedProx + LoRA.}
This method extends FedAvg + LoRA by adding a proximal regularizer to the local LoRA objective. It follows the classical {FedProx} formulation~\cite{li2020fedprox}, while applying the regularization only to trainable LoRA parameters. This category represents local-correction baselines for federated LLM fine-tuning, as also included in {OpenFedLLM}~\cite{ye2024openfedllm}.

\textbf{HeteroLoRA.}
This family targets client resource heterogeneity by allowing different clients to use different LoRA ranks or adapter configurations. Existing examples include {FlexLoRA}~\cite{bai2024flexlora}, {FLoRA}~\cite{wang2024flora}, and {HLoRA}~\cite{liu2025hlora}. In our benchmark, HeteroLoRA represents this class by assigning heterogeneous LoRA ranks to clients and aligning them to a common shape before aggregation.

\textbf{SplitLoRA.}
This family combines split learning with LoRA-based fine-tuning. Instead of training the entire model on each client, {SplitLoRA}~\cite{lin2024splitlora} partitions the LLM into client-side and server-side submodels. Training exchanges activations and activation gradients between the two sides, while client-side LoRA adapters are periodically synchronized through federated aggregation.

\vspace{-2mm}
\subsection{Parameter Settings}
\vspace{-2mm}
We use the AdamW optimizer with a weight decay of $0.01$ across all settings.
The learning rate is set to $2\times10^{-4}$.
The sequence length is set to $64$ for ARC-Easy, HellaSwag, PIQA, QNLI, SocialIQA, and WinoGrande, while BoolQ uses a longer sequence length of $128$ due to its relatively longer input contexts.
Following FlowerTune~\cite{gao2025flowertune}, we partition the training data of each task into approximately equal-size client shards.
All evaluations are conducted in a zero-shot setting.
For accuracy evaluation, we adopt letter-token classification accuracy for multiple-choice tasks, and the final accuracy is computed as the fraction of examples for which the predicted letter matches the ground-truth answer. This follows the common likelihood-based multiple-choice evaluation protocol used for autoregressive language models~\cite{brown2020language,wang2024answerc}.

For method-specific settings, FedAvg+LoRA and FedProx+LoRA use a LoRA rank of $8$ for all clients.
For HeteroLoRA, the LoRA rank is set according to device capability: Jetson and iQOO clients use rank $8$, while the remaining clients use rank $4$.
For SplitLoRA, the LoRA rank is set to $8$, and each client-side model contains only the first hidden layer of the backbone model.
\

\vspace{-2mm}
\section{Results and Analysis}
\vspace{-2mm}

We report representative case studies in the main paper and provide the complete benchmark results in the Appendix~\ref{AppendixA}-\ref{AppendixD} . Across the full results covering 3 backbone models, 7 datasets, and 4 federated fine-tuning methods, we observe a consistent qualitative conclusion: accuracy-only rankings do not necessarily align with deployability-aware rankings once wall-clock time, communication, energy consumption, memory feasibility, and robustness are considered. In the main-paper results, for Protocols A and B, we select one representative task from each category: BoolQ for \textit{Verify}, SocialIQA for \textit{Choose}, and ARC-Easy for \textit{Reason}. For Protocol C, we additionally report HellaSwag for communication fluctuation and client dropout, and ARC-Easy for device heterogeneity. All detailed results in the main paper use Qwen2.5-0.5B as the backbone model.

\begin{table*}[t]
\centering
\caption{Testing quality across selected tasks and methods.}
\label{tab:testing_quality_selected}
\renewcommand{\arraystretch}{1.15}
\setlength{\tabcolsep}{5pt}
\resizebox{0.9\textwidth}{!}{
\begin{tabular}{lllccc ccc}
\toprule
\multirow{2}{*}{\textbf{Task Type}} 
& \multirow{2}{*}{\textbf{Task}} 
& \multirow{2}{*}{\textbf{Method}} 
& \multicolumn{3}{c}{\textbf{Testing Loss (lower is better)}} 
& \multicolumn{3}{c}{\textbf{Testing Accuracy (higher is better)}} \\
\cmidrule(lr){4-6} \cmidrule(lr){7-9}
& & 
& \textbf{Pretrained} 
& \textbf{Best Finetuned} 
& \textbf{Rank} 
& \textbf{Pretrained} 
& \textbf{Best Finetuned} 
& \textbf{Rank} \\
\midrule

\multirow{6}{*}{Verify}
& \multirow{6}{*}{BoolQ}
& \cellcolor{gray!10} Centroid     
& \multirow{6}{*}{0.6393} 
& \cellcolor{gray!10} 0.4788 
& \cellcolor{gray!10} Lower Bound 
& \multirow{6}{*}{63.21\%} 
& \cellcolor{gray!10} 80.24\% 
& \cellcolor{gray!10} Upper Bound \\
& & FedAvg+LoRA   
& & 0.5051 & 2 
& & 77.68\% & 2 \\
& & \cellcolor{gray!10} FedProx+LoRA  
& & \cellcolor{gray!10} 0.4939 
& \cellcolor{gray!10} 1 
& & \cellcolor{gray!10} 78.20\% 
& \cellcolor{gray!10} 1 \\
& & HeteroLoRA    
& & 0.5059 & 4 
& & 76.73\% & 4 \\
& & \cellcolor{gray!10} SplitLoRA     
& & \cellcolor{gray!10} 0.5161 
& \cellcolor{gray!10} 3 
& & \cellcolor{gray!10} 76.79\% 
& \cellcolor{gray!10} 3 \\
& & Local Only    
& & 0.6127 & Upper Bound 
& & 71.99\% & Lower Bound \\

\midrule

\multirow{6}{*}{Choose}
& \multirow{6}{*}{SocialIQA}
& \cellcolor{gray!10} Centroid     
& \multirow{6}{*}{0.9644} 
& \cellcolor{gray!10} 0.7769 
& \cellcolor{gray!10} Lower Bound 
& \multirow{6}{*}{55.99\%} 
& \cellcolor{gray!10} 68.07\% 
& \cellcolor{gray!10} Upper Bound \\
& & FedAvg+LoRA   
& & 0.7993 & 2 
& & 66.63\% & 2 \\
& & \cellcolor{gray!10} FedProx+LoRA  
& & \cellcolor{gray!10} 0.7863 
& \cellcolor{gray!10} 1 
& & \cellcolor{gray!10} 67.35\% 
& \cellcolor{gray!10} 1 \\
& & HeteroLoRA    
& & 0.8250 & 4 
& & 66.07\% & 4 \\
& & \cellcolor{gray!10} SplitLoRA     
& & \cellcolor{gray!10} 0.8044 
& \cellcolor{gray!10} 3 
& & \cellcolor{gray!10} 66.38\% 
& \cellcolor{gray!10} 3 \\
& & Local Only    
& & 0.9981 & Upper Bound 
& & 59.31\% & Lower Bound \\

\midrule

\multirow{6}{*}{Reason}
& \multirow{6}{*}{ARC-E}
& \cellcolor{gray!10} Centroid     
& \multirow{6}{*}{0.7180} 
& \cellcolor{gray!10} 0.5913 
& \cellcolor{gray!10} Lower Bound 
& \multirow{6}{*}{71.05\%} 
& \cellcolor{gray!10} 79.47\% 
& \cellcolor{gray!10} Upper Bound \\
& & FedAvg+LoRA   
& & 0.6054 & 2 
& & 78.60\% & 2 \\
& & \cellcolor{gray!10} FedProx+LoRA  
& & \cellcolor{gray!10} 0.6013 
& \cellcolor{gray!10} 1 
& & \cellcolor{gray!10} 78.77\% 
& \cellcolor{gray!10} 1 \\
& & HeteroLoRA    
& & 0.6274 & 4 
& & 76.67\% & 4 \\
& & \cellcolor{gray!10} SplitLoRA     
& & \cellcolor{gray!10} 0.6102 
& \cellcolor{gray!10} 3 
& & \cellcolor{gray!10} 77.02\% 
& \cellcolor{gray!10} 3 \\
& & Local Only    
& & 0.6693 & Upper Bound 
& & 74.53\% & Lower Bound \\

\bottomrule
\end{tabular}
}
\vspace{-5mm}
\end{table*}

\vspace{-2mm}
\subsection{Results of Protocol A: Quality-under-Budget}
\label{sec:4.1}

Under Protocol A, we evaluate the quality that each method can achieve under the current edge-system budget. 
We consider a total of 100 clients, with 20 clients instantiated for each device type listed in Table~\ref{tab:client_devices}. 
In each communication round, 10 clients are randomly selected to participate in federated fine-tuning. 
The experimental results are reported in Table~\ref{tab:testing_quality_selected}.

Across the three representative tasks, FedAvg+LoRA achieves the strongest or tied strongest accuracy among the federated baselines, while HeteroLoRA generally obtains the lowest accuracy. However, final quality alone does not fully characterize edge deployability. 
Although SplitLoRA ranks third in quality in Table~\ref{tab:testing_quality_selected}, it provides substantially better executability under larger model settings. 
Specifically, for Qwen2.5-0.5B and Gemma-3-270M, all selected methods can be executed on the evaluated devices. 
In contrast, when scaling to Gemma-3-1B, only SplitLoRA can be successfully executed, while the other methods encounter out-of-memory failures. 
The complete results for the larger model setting are provided in the Appendix. 
This observation shows that quality-under-budget evaluation should consider not only the final accuracy achieved by a method, but also whether the method can actually run within the memory and system constraints of edge devices.

\begin{insightbox}
\textbf{Insight 1.}
A method may achieve strong final task performance under feasible settings, but still be impractical for edge deployment if its system cost exceeds what edge devices can sustain.
\end{insightbox}

\vspace{-2mm}
\subsection{Results of Protocol B: Cost-to-Target}

For Protocol B, we use the same client configuration as in Sec.~\ref{sec:4.1}
For each task, we set three target accuracy levels to represent different stages of training progress.
These targets are derived from the improvement interval between the pretrained model and the centroid upper bound.
Specifically, the three targets correspond to reaching 50\%, 70\%, and 90\% of the accuracy improvement from the pretrained baseline to the centroid result.

Table~\ref{tab:cost_to_target_selected} reports four system metrics when each method first reaches the corresponding target accuracy.
The wall-clock time, communication volume, and energy consumption are recorded at the first point where the target accuracy is achieved.
Peak memory is computed as the average peak memory across all participating clients up to that point\footnote{We include peak memory because whether a fine-tuning method can be executed on edge devices directly depends on whether its memory footprint exceeds the device memory limit.}.
Since all metrics in this table represent system cost, lower values are better, and a smaller rank indicates a more efficient method.

The results show that although FedAvg+LoRA and FedProx+LoRA achieve competitive final accuracy, they often require much longer wall-clock time, higher energy consumption, and larger peak memory to reach the same target accuracy.
The ranking also varies across metrics and target accuracy levels; for example, methods with low communication cost at an early target may not remain the most efficient as the target accuracy increases.

\begin{insightbox}
\textbf{Insight 2.}
Methods with higher final accuracy do not necessarily provide better system efficiency.
Their relative efficiency can also change across target accuracy levels and system metrics.
\end{insightbox}

\begin{table*}[t]
\centering
\caption{System cost to reach target accuracy across selected tasks and methods.}
\label{tab:cost_to_target_selected}
\renewcommand{\arraystretch}{1.15}
\setlength{\tabcolsep}{4pt}
\resizebox{0.95\textwidth}{!}{
\begin{tabular}{llclcccccccc}
\toprule
\textbf{Task Type} 
& \textbf{Task} 
& \textbf{Target Accuracy} 
& \textbf{Methods} 
& \makecell{\textbf{Wall-clock}\\\textbf{time (hour)}} 
& \textbf{Rank} 
& \makecell{\textbf{Communication}\\\textbf{volume (MB)}} 
& \textbf{Rank} 
& \makecell{\textbf{Energy}\\\textbf{consumption (kJ)}} 
& \textbf{Rank} 
& \makecell{\textbf{Peak}\\\textbf{memory (MB)}} 
& \textbf{Rank} \\
\midrule

\multirow{12}{*}{Verify}
& \multirow{12}{*}{BoolQ}
& \multirow{4}{*}{71\%}
& \cellcolor{gray!10} FedAvg+LoRA  & \cellcolor{gray!10} 6.65 & \cellcolor{gray!10} 3 & \cellcolor{gray!10} 6624.38 & \cellcolor{gray!10} 1 & \cellcolor{gray!10} 300.67 & \cellcolor{gray!10} 3 & \cellcolor{gray!10} 3453.60 & \cellcolor{gray!10} 3 \\
& & & FedProx+LoRA & 6.67 & 4 & 6624.38 & 1 & 301.19 & 4 & 3470.35 & 4 \\
& & & \cellcolor{gray!10} HeteroLoRA & \cellcolor{gray!10} 3.07 & \cellcolor{gray!10} 2 & \cellcolor{gray!10} 9326.95 & \cellcolor{gray!10} 2 & \cellcolor{gray!10} 138.66 & \cellcolor{gray!10} 2 & \cellcolor{gray!10} 2499.21 & \cellcolor{gray!10} 2 \\
& & & SplitLoRA & 0.43 & 1 & 11744.04 & 3 & 19.42 & 1 & 1142.87 & 1 \\

\cmidrule(lr){3-12}

& & \multirow{4}{*}{75\%}
& \cellcolor{gray!10} FedAvg+LoRA  & \cellcolor{gray!10} 12.48 & \cellcolor{gray!10} 3 & \cellcolor{gray!10} 12420.70 & \cellcolor{gray!10} 1 & \cellcolor{gray!10} 563.76 & \cellcolor{gray!10} 3 & \cellcolor{gray!10} 3497.81 & \cellcolor{gray!10} 3 \\
& & & FedProx+LoRA & 12.50 & 4 & 12420.70 & 1 & 564.83 & 4 & 3514.77 & 4 \\
& & & \cellcolor{gray!10} HeteroLoRA & \cellcolor{gray!10} 5.32 & \cellcolor{gray!10} 2 & \cellcolor{gray!10} 16166.72 & \cellcolor{gray!10} 3 & \cellcolor{gray!10} 240.57 & \cellcolor{gray!10} 2 & \cellcolor{gray!10} 2531.20 & \cellcolor{gray!10} 2 \\
& & & SplitLoRA & 0.47 & 1 & 12811.68 & 2 & 21.18 & 1 & 1157.50 & 1 \\

\cmidrule(lr){3-12}

& & \multirow{4}{*}{78\%}
& \cellcolor{gray!10} FedAvg+LoRA  & \cellcolor{gray!10} 32.46 & \cellcolor{gray!10} 3 & \cellcolor{gray!10} 32293.83 & \cellcolor{gray!10} 2 & \cellcolor{gray!10} 1466.74 & \cellcolor{gray!10} 3 & \cellcolor{gray!10} 3520.44 & \cellcolor{gray!10} 3 \\
& & & FedProx+LoRA & 32.50 & 4 & 32293.83 & 2 & 1468.52 & 4 & 3537.51 & 4 \\
& & & \cellcolor{gray!10} HeteroLoRA & \cellcolor{gray!10} 13.33 & \cellcolor{gray!10} 2 & \cellcolor{gray!10} 40416.80 & \cellcolor{gray!10} 3 & \cellcolor{gray!10} 602.20 & \cellcolor{gray!10} 2 & \cellcolor{gray!10} 2547.57 & \cellcolor{gray!10} 2 \\
& & & SplitLoRA & 0.99 & 1 & 27224.82 & 1 & 44.94 & 1 & 1164.99 & 1 \\

\midrule

\multirow{12}{*}{Choose}
& \multirow{12}{*}{SocialIQA}
& \multirow{4}{*}{62\%}
& \cellcolor{gray!10} FedAvg+LoRA  & \cellcolor{gray!10} 15.79 & \cellcolor{gray!10} 4 & \cellcolor{gray!10} 15732.89 & \cellcolor{gray!10} 2 & \cellcolor{gray!10} 772.17 & \cellcolor{gray!10} 4 & \cellcolor{gray!10} 3418.89 & \cellcolor{gray!10} 3 \\
& & & FedProx+LoRA & 15.71 & 3 & 15732.89 & 2 & 768.41 & 3 & 3462.20 & 4 \\
& & & \cellcolor{gray!10} HeteroLoRA & \cellcolor{gray!10} 5.32 & \cellcolor{gray!10} 2 & \cellcolor{gray!10} 16166.72 & \cellcolor{gray!10} 3 & \cellcolor{gray!10} 259.95 & \cellcolor{gray!10} 2 & \cellcolor{gray!10} 2562.68 & \cellcolor{gray!10} 2 \\
& & & SplitLoRA & 0.44 & 1 & 10142.58 & 1 & 21.54 & 1 & 1147.95 & 1 \\

\cmidrule(lr){3-12}

& & \multirow{4}{*}{64\%}
& \cellcolor{gray!10} FedAvg+LoRA  & \cellcolor{gray!10} 34.91 & \cellcolor{gray!10} 4 & \cellcolor{gray!10} 34777.97 & \cellcolor{gray!10} 3 & \cellcolor{gray!10} 1707.33 & \cellcolor{gray!10} 4 & \cellcolor{gray!10} 3455.95 & \cellcolor{gray!10} 3 \\
& & & FedProx+LoRA & 34.72 & 3 & 34777.97 & 3 & 1697.93 & 3 & 3499.73 & 4 \\
& & & \cellcolor{gray!10} HeteroLoRA & \cellcolor{gray!10} 9.98 & \cellcolor{gray!10} 2 & \cellcolor{gray!10} 30468.05 & \cellcolor{gray!10} 2 & \cellcolor{gray!10} 488.07 & \cellcolor{gray!10} 2 & \cellcolor{gray!10} 2590.46 & \cellcolor{gray!10} 2 \\
& & & SplitLoRA & 0.81 & 1 & 18683.70 & 1 & 39.61 & 1 & 1160.39 & 1 \\

\cmidrule(lr){3-12}

& & \multirow{4}{*}{66\%}
& \cellcolor{gray!10} FedAvg+LoRA  & \cellcolor{gray!10} 94.94 & \cellcolor{gray!10} 4 & \cellcolor{gray!10} 94397.34 & \cellcolor{gray!10} 3 & \cellcolor{gray!10} 4643.29 & \cellcolor{gray!10} 4 & \cellcolor{gray!10} 3476.19 & \cellcolor{gray!10} 3 \\
& & & FedProx+LoRA & 94.25 & 3 & 94397.34 & 3 & 4609.67 & 3 & 3520.23 & 4 \\
& & & \cellcolor{gray!10} HeteroLoRA & \cellcolor{gray!10} 29.94 & \cellcolor{gray!10} 2 & \cellcolor{gray!10} 91404.14 & \cellcolor{gray!10} 2 & \cellcolor{gray!10} 1464.33 & \cellcolor{gray!10} 2 & \cellcolor{gray!10} 2605.63 & \cellcolor{gray!10} 2 \\
& & & SplitLoRA & 1.34 & 1 & 30961.56 & 1 & 65.53 & 1 & 1167.19 & 1 \\

\midrule

\multirow{12}{*}{Reason}
& \multirow{12}{*}{ARC-E}
& \multirow{4}{*}{75\%}
& \cellcolor{gray!10} FedAvg+LoRA  & \cellcolor{gray!10} 24.29 & \cellcolor{gray!10} 4 & \cellcolor{gray!10} 24013.36 & \cellcolor{gray!10} 2 & \cellcolor{gray!10} 1176.40 & \cellcolor{gray!10} 4 & \cellcolor{gray!10} 3404.12 & \cellcolor{gray!10} 3 \\
& & & FedProx+LoRA & 24.13 & 3 & 24013.36 & 2 & 1168.56 & 3 & 3441.75 & 4 \\
& & & \cellcolor{gray!10} HeteroLoRA & \cellcolor{gray!10} 4.00 & \cellcolor{gray!10} 2 & \cellcolor{gray!10} 24250.08 & \cellcolor{gray!10} 3 & \cellcolor{gray!10} 193.52 & \cellcolor{gray!10} 2 & \cellcolor{gray!10} 2540.26 & \cellcolor{gray!10} 2 \\
& & & SplitLoRA & 0.48 & 1 & 10676.40 & 1 & 23.09 & 1 & 1141.11 & 1 \\

\cmidrule(lr){3-12}

& & \multirow{4}{*}{76\%}
& \cellcolor{gray!10} FedAvg+LoRA  & \cellcolor{gray!10} 32.64 & \cellcolor{gray!10} 3 & \cellcolor{gray!10} 32293.83 & \cellcolor{gray!10} 2 & \cellcolor{gray!10} 1580.47 & \cellcolor{gray!10} 3 & \cellcolor{gray!10} 3445.54 & \cellcolor{gray!10} 3 \\
& & & FedProx+LoRA & 33.31 & 4 & 33121.88 & 3 & 1612.96 & 4 & 3483.62 & 4 \\
& & & \cellcolor{gray!10} HeteroLoRA & \cellcolor{gray!10} 7.89 & \cellcolor{gray!10} 2 & \cellcolor{gray!10} 47878.36 & \cellcolor{gray!10} 4 & \cellcolor{gray!10} 382.26 & \cellcolor{gray!10} 2 & \cellcolor{gray!10} 2571.17 & \cellcolor{gray!10} 2 \\
& & & SplitLoRA & 0.95 & 1 & 21352.80 & 1 & 46.20 & 1 & 1154.99 & 1 \\

\cmidrule(lr){3-12}

& & \multirow{4}{*}{77\%}
& \cellcolor{gray!10} FedAvg+LoRA  & \cellcolor{gray!10} 66.91 & \cellcolor{gray!10} 4 & \cellcolor{gray!10} 66243.75 & \cellcolor{gray!10} 2 & \cellcolor{gray!10} 3240.10 & \cellcolor{gray!10} 4 & \cellcolor{gray!10} 3458.55 & \cellcolor{gray!10} 3 \\
& & & FedProx+LoRA & 66.58 & 3 & 66243.75 & 2 & 3223.88 & 3 & 3496.77 & 4 \\
& & & \cellcolor{gray!10} HeteroLoRA & \cellcolor{gray!10} 11.17 & \cellcolor{gray!10} 2 & \cellcolor{gray!10} 67775.86 & \cellcolor{gray!10} 3 & \cellcolor{gray!10} 540.93 & \cellcolor{gray!10} 2 & \cellcolor{gray!10} 2580.88 & \cellcolor{gray!10} 2 \\
& & & SplitLoRA & 1.53 & 1 & 34164.48 & 1 & 73.86 & 1 & 1159.35 & 1 \\

\bottomrule
\end{tabular}
}
\vspace{-4mm}
\end{table*}

\begin{table*}[]
\centering
\caption{Robustness under communication fluctuation. Values in parentheses denote changes relative to the no-fluctuation setting. Ranks are computed by the absolute value of the change, where smaller change indicates better robustness. (HellaSwag@Qwen2.5-0.5B)}
\label{tab:protocol_c_comm_fluctuation}
\renewcommand{\arraystretch}{1.12}
\setlength{\tabcolsep}{3.2pt}
\scriptsize
\providecommand{\gc}[1]{\cellcolor{gray!10}#1}
\resizebox{0.95\textwidth}{!}{
\begin{tabular}{lcccccccccc}
\toprule
\textbf{Method}
& \textbf{Testing Accuracy}
& \textbf{Rank}
& \makecell{\textbf{Wall-clock}\\\textbf{time (h)}}
& \textbf{Rank}
& \makecell{\textbf{Communication}\\\textbf{volume (MB)}}
& \textbf{Rank}
& \makecell{\textbf{Energy}\\\textbf{consumption (kJ)}}
& \textbf{Rank}
& \makecell{\textbf{Peak}\\\textbf{memory (MB)}}
& \textbf{Rank} \\
\midrule

\gc{FedAvg+LoRA}
& \gc{33.68\% $(+0.00\%)$} & \gc{1}
& \gc{1.08 $(+0.02)$} & \gc{3}
& \gc{12006.68 $(+0.00)$} & \gc{1}
& \gc{38.61 $(+1.78)$} & \gc{1}
& \gc{3557.55 $(+0.00)$} & \gc{1} \\

FedProx+LoRA
& 33.68\% $(+0.00\%)$ & 1
& 1.15 $(+0.02)$ & 2
& 12006.68 $(+0.00)$ & 1
& 44.39 $(+2.08)$ & 2
& 3548.12 $(+0.00)$ & 1 \\

\gc{HeteroLoRA}
& \gc{33.12\% $(+0.00\%)$} & \gc{1}
& \gc{5.51 $(+0.00)$} & \gc{1}
& \gc{11751.96 $(+0.00)$} & \gc{1}
& \gc{233.79 $(+8.47)$} & \gc{3}
& \gc{3552.89 $(+0.00)$} & \gc{1} \\

SplitLoRA
& 33.67\% $(+0.00\%)$ & 1
& 12.08 $(+4.46)$ & 4
& 10676.40 $(+0.00)$ & 1
& 515.81 $(+203.62)$ & 4
& 3548.27 $(+0.00)$ & 1 \\

\bottomrule
\end{tabular}
}
\vspace{-5mm}
\end{table*}

\vspace{-2mm}
\subsection{Results of Protocol C: Robustness}

\textbf{Communication fluctuation:} Table~\ref{tab:protocol_c_comm_fluctuation} reports the results of Protocol C under dynamic communication fluctuation. 
To simulate unstable wireless connectivity in real edge deployments, we dynamically change the Wi-Fi bandwidth every $1/3$ hour, sequentially setting it to the full bandwidth, $1/2$ of the bandwidth, and $1/4$ of the bandwidth. 
The results show that communication fluctuation can affect system cost even when final model quality remains similar, and the magnitude of this impact varies across methods.

\textbf{Drop-out:} Table~\ref{tab:protocol_c_dropout} reports the results of Protocol C under different client drop-out ratios. 
We vary the drop-out ratio from 10\% to 50\% to simulate intermittent client availability in real edge deployments. 
The results show that client drop-out can affect both model quality and system cost, and the degree of degradation differs across methods.

\textbf{Heterogeneity:} Table~\ref{tab:protocol_c_heterogeneity} reports the results of Protocol C under different client heterogeneity settings. 
We keep the total number of clients fixed at 100 and vary the device composition to simulate different deployment scenarios. 
For each metric, the value outside the parentheses is the measured result under the corresponding client mix, while the value inside the parentheses denotes the change relative to the balanced setting, i.e., 20J+20I+20P+20M+20N. 
The results show that client heterogeneity affects model quality and system efficiency in a method-dependent manner, indicating that different methods exhibit different levels of robustness to shifts in device composition.

\begin{insightbox}
\textbf{Insight 3.}
Edge-system perturbations affect different methods differently, leading to method-dependent robustness in both model quality and system efficiency.
\end{insightbox}

\begin{table*}[t]
\centering
\caption{Robustness under different client dropout ratios. Values in parentheses denote changes relative to the no-dropout setting. Ranks are computed by the absolute value of the change, where smaller change indicates better robustness.(HellaSwag@Qwen2.5-0.5B)}
\label{tab:protocol_c_dropout}
\renewcommand{\arraystretch}{1.12}
\setlength{\tabcolsep}{3.2pt}
\scriptsize
\providecommand{\gc}[1]{\cellcolor{gray!10}#1}
\resizebox{0.95\textwidth}{!}{
\begin{tabular}{llcccccccccc}
\toprule
\textbf{Dropout Ratio} 
& \textbf{Method}
& \makecell{\textbf{Testing}\\\textbf{Accuracy}}
& \textbf{Rank}
& \makecell{\textbf{Wall-clock}\\\textbf{time (h)}}
& \textbf{Rank}
& \makecell{\textbf{Communication}\\\textbf{volume (MB)}}
& \textbf{Rank}
& \makecell{\textbf{Energy}\\\textbf{consumption (kJ)}}
& \textbf{Rank}
& \makecell{\textbf{Peak}\\\textbf{memory (MB)}}
& \textbf{Rank} \\
\midrule

\multirow{4}{*}{10\%}
& \gc{FedAvg+LoRA} 
& \gc{33.70\% $(-0.14\%)$} & \gc{4}
& \gc{184.96 $(+26.84)$} & \gc{3}
& \gc{8250.00 $(+412.50)$} & \gc{3}
& \gc{8956.40 $(+1299.76)$} & \gc{3}
& \gc{4551.18 $(-0.75)$} & \gc{4} \\
& FedProx+LoRA
& 33.72\% $(-0.12\%)$ & 3
& 185.16 $(+26.85)$ & 4
& 8250.00 $(+412.50)$ & 3
& 8966.19 $(+1300.06)$ & 4
& 4580.47 $(-0.11)$ & 1 \\
& \gc{HeteroLoRA}
& \gc{33.13\% $(-0.05\%)$} & \gc{1}
& \gc{182.40 $(+22.36)$} & \gc{2}
& \gc{8250.00 $(+206.25)$} & \gc{2}
& \gc{8832.25 $(+1082.90)$} & \gc{2}
& \gc{3927.04 $(-0.57)$} & \gc{2} \\
& SplitLoRA
& 33.59\% $(-0.08\%)$ & 2
& 5.22 $(+0.52)$ & 1
& 8250.00 $(+0.00)$ & 1
& 252.66 $(+25.27)$ & 1
& 2023.79 $(+0.58)$ & 3 \\

\midrule

\multirow{4}{*}{30\%}
& \gc{FedAvg+LoRA} 
& \gc{33.82\% $(-0.02\%)$} & \gc{1}
& \gc{237.81 $(+79.69)$} & \gc{3}
& \gc{8250.00 $(+412.50)$} & \gc{3}
& \gc{11515.38 $(+3858.73)$} & \gc{3}
& \gc{4551.10 $(-0.83)$} & \gc{1} \\
& FedProx+LoRA
& 33.81\% $(-0.03\%)$ & 2
& 238.07 $(+79.75)$ & 4
& 8250.00 $(+412.50)$ & 3
& 11527.95 $(+3861.83)$ & 4
& 4582.08 $(+1.50)$ & 3 \\
& \gc{HeteroLoRA}
& \gc{33.09\% $(-0.09\%)$} & \gc{3}
& \gc{234.51 $(+74.48)$} & \gc{2}
& \gc{8250.00 $(+206.25)$} & \gc{2}
& \gc{11355.75 $(+3606.40)$} & \gc{2}
& \gc{3928.49 $(+0.88)$} & \gc{2} \\
& SplitLoRA
& 33.57\% $(-0.10\%)$ & 4
& 6.71 $(+2.01)$ & 1
& 8250.00 $(+0.00)$ & 1
& 324.85 $(+97.45)$ & 1
& 2024.71 $(+1.50)$ & 3 \\

\midrule

\multirow{4}{*}{50\%}
& \gc{FedAvg+LoRA} 
& \gc{33.46\% $(-0.38\%)$} & \gc{4}
& \gc{289.47 $(+131.35)$} & \gc{3}
& \gc{7177.50 $(-660.00)$} & \gc{2}
& \gc{14017.18 $(+6360.54)$} & \gc{3}
& \gc{4553.43 $(+1.50)$} & \gc{1} \\
& FedProx+LoRA
& 33.73\% $(-0.11\%)$ & 1
& 271.62 $(+113.31)$ & 2
& 6723.75 $(-1113.75)$ & 3
& 13152.87 $(+5486.74)$ & 2
& 4582.08 $(+1.50)$ & 1 \\
& \gc{HeteroLoRA}
& \gc{33.06\% $(-0.12\%)$} & \gc{2}
& \gc{328.31 $(+168.28)$} & \gc{4}
& \gc{8250.00 $(+206.25)$} & \gc{1}
& \gc{15898.05 $(+8148.70)$} & \gc{4}
& \gc{3929.11 $(+1.50)$} & \gc{1} \\
& SplitLoRA
& 33.34\% $(-0.33\%)$ & 3
& 7.56 $(+2.87)$ & 1
& 6641.25 $(-1608.75)$ & 4
& 366.23 $(+138.84)$ & 1
& 2024.71 $(+1.50)$ & 1 \\

\bottomrule
\end{tabular}
}
\vspace{-3mm}
\end{table*}

\begin{table*}[t]
\centering
\caption{Robustness under different client heterogeneity settings. Values in parentheses denote changes relative to the balanced client mix. Ranks are computed by the absolute value of the change, where smaller change indicates better robustness. (ARC-E@Qwen2.5-0.5B)}
\label{tab:protocol_c_heterogeneity}
\renewcommand{\arraystretch}{1.12}
\setlength{\tabcolsep}{3.2pt}
\scriptsize
\providecommand{\gc}[1]{\cellcolor{gray!10}#1}
\resizebox{0.95\textwidth}{!}{
\begin{tabular}{llcccccccccc}
\toprule
\textbf{Client Mix} 
& \textbf{Method}
& \makecell{\textbf{Testing}\\\textbf{Accuracy}}
& \textbf{Rank}
& \makecell{\textbf{Wall-clock}\\\textbf{time (h)}}
& \textbf{Rank}
& \makecell{\textbf{Communication}\\\textbf{volume (MB)}}
& \textbf{Rank}
& \makecell{\textbf{Energy}\\\textbf{consumption (kJ)}}
& \textbf{Rank}
& \makecell{\textbf{Peak}\\\textbf{memory (MB)}}
& \textbf{Rank} \\
\midrule

\multirow{4}{*}{100J}
& \gc{FedAvg+LoRA} 
& \gc{78.60\% $(+0.00\%)$} & \gc{1}
& \gc{19.85 $(-136.51)$} & \gc{4}
& \gc{6624.38 $(+0.00)$} & \gc{1}
& \gc{960.95 $(-6609.90)$} & \gc{4}
& \gc{4569.06 $(+0.60)$} & \gc{3} \\
& FedProx+LoRA
& 78.77\% $(+0.00\%)$ & 1
& 23.92 $(-136.49)$ & 3
& 6624.38 $(+0.00)$ & 1
& 1158.02 $(-6608.81)$ & 3
& 4569.57 $(+0.50)$ & 2 \\
& \gc{HeteroLoRA}
& \gc{76.84\% $(+0.17\%)$} & \gc{2}
& \gc{31.57 $(-78.07)$} & \gc{2}
& \gc{3300.00 $(-1608.75)$} & \gc{2}
& \gc{1528.42 $(-3780.15)$} & \gc{2}
& \gc{3913.44 $(+0.40)$} & \gc{1} \\
& SplitLoRA
& 77.02\% $(+0.00\%)$ & 1
& 4.83 $(+0.04)$ & 1
& 3416.45 $(+0.00)$ & 1
& 233.71 $(+1.78)$ & 1
& 2018.80 $(+0.40)$ & 1 \\

\midrule

\multirow{4}{*}{\makecell{70J+20I\\+10P}}
& \gc{FedAvg+LoRA} 
& \gc{78.60\% $(+0.00\%)$} & \gc{1}
& \gc{52.48 $(-103.88)$} & \gc{3}
& \gc{6624.38 $(+0.00)$} & \gc{1}
& \gc{2540.99 $(-5029.87)$} & \gc{3}
& \gc{4568.76 $(+0.30)$} & \gc{3} \\
& FedProx+LoRA
& 78.77\% $(+0.00\%)$ & 1
& 52.34 $(-108.07)$ & 4
& 6624.38 $(+0.00)$ & 1
& 2534.20 $(-5232.63)$ & 4
& 4569.32 $(+0.25)$ & 2 \\
& \gc{HeteroLoRA}
& \gc{76.75\% $(+0.08\%)$} & \gc{2}
& \gc{51.46 $(-58.17)$} & \gc{2}
& \gc{6063.75 $(+1155.00)$} & \gc{2}
& \gc{2491.76 $(-2816.80)$} & \gc{2}
& \gc{3913.24 $(+0.20)$} & \gc{1} \\
& SplitLoRA
& 77.02\% $(+0.00\%)$ & 1
& 4.82 $(+0.03)$ & 1
& 3416.45 $(+0.00)$ & 1
& 233.19 $(+1.25)$ & 1
& 2018.60 $(+0.20)$ & 1 \\

\midrule

\multirow{4}{*}{\makecell{20J+20I+20P\\+20M+20N\\\textit{Reference}}}
& \gc{FedAvg+LoRA} 
& \gc{78.60\% $(+0.00\%)$} & \gc{1}
& \gc{156.36 $(+0.00)$} & \gc{1}
& \gc{6624.38 $(+0.00)$} & \gc{1}
& \gc{7570.86 $(+0.00)$} & \gc{1}
& \gc{4568.46 $(+0.00)$} & \gc{1} \\
& FedProx+LoRA
& 78.77\% $(+0.00\%)$ & 1
& 160.41 $(+0.00)$ & 1
& 6624.38 $(+0.00)$ & 1
& 7766.83 $(+0.00)$ & 1
& 4569.07 $(+0.00)$ & 1 \\
& \gc{HeteroLoRA}
& \gc{76.67\% $(+0.00\%)$} & \gc{1}
& \gc{109.64 $(+0.00)$} & \gc{1}
& \gc{4908.75 $(+0.00)$} & \gc{1}
& \gc{5308.56 $(+0.00)$} & \gc{1}
& \gc{3913.04 $(+0.00)$} & \gc{1} \\
& SplitLoRA
& 77.02\% $(+0.00\%)$ & 1
& 4.79 $(+0.00)$ & 1
& 3416.45 $(+0.00)$ & 1
& 231.94 $(+0.00)$ & 1
& 2018.40 $(+0.00)$ & 1 \\

\midrule

\multirow{4}{*}{\makecell{10J+20I+20P\\+20M+30N}}
& \gc{FedAvg+LoRA} 
& \gc{78.60\% $(+0.00\%)$} & \gc{1}
& \gc{166.01 $(+9.65)$} & \gc{4}
& \gc{6624.38 $(+0.00)$} & \gc{1}
& \gc{8038.00 $(+467.14)$} & \gc{4}
& \gc{4568.16 $(-0.30)$} & \gc{3} \\
& FedProx+LoRA
& 78.77\% $(+0.00\%)$ & 1
& 165.21 $(+4.81)$ & 3
& 6624.38 $(+0.00)$ & 1
& 7999.56 $(+232.73)$ & 3
& 4568.82 $(-0.25)$ & 2 \\
& \gc{HeteroLoRA}
& \gc{76.40\% $(-0.27\%)$} & \gc{2}
& \gc{113.26 $(+3.62)$} & \gc{2}
& \gc{4743.75 $(-165.00)$} & \gc{2}
& \gc{5483.93 $(+175.37)$} & \gc{2}
& \gc{3912.84 $(-0.20)$} & \gc{1} \\
& SplitLoRA
& 77.02\% $(+0.00\%)$ & 1
& 4.83 $(+0.04)$ & 1
& 3416.45 $(+0.00)$ & 1
& 233.79 $(+1.85)$ & 1
& 2018.20 $(-0.20)$ & 1 \\

\midrule

\multirow{4}{*}{\makecell{10I+10P\\+30M+50N}}
& \gc{FedAvg+LoRA} 
& \gc{78.60\% $(+0.00\%)$} & \gc{1}
& \gc{168.82 $(+12.47)$} & \gc{4}
& \gc{6624.38 $(+0.00)$} & \gc{1}
& \gc{8174.42 $(+603.56)$} & \gc{4}
& \gc{4567.86 $(-0.60)$} & \gc{3} \\
& FedProx+LoRA
& 78.77\% $(+0.00\%)$ & 1
& 167.20 $(+6.79)$ & 3
& 6624.38 $(+0.00)$ & 1
& 8095.70 $(+328.87)$ & 3
& 4568.57 $(-0.50)$ & 2 \\
& \gc{HeteroLoRA}
& \gc{76.32\% $(-0.35\%)$} & \gc{2}
& \gc{114.62 $(+4.98)$} & \gc{2}
& \gc{4166.25 $(-742.50)$} & \gc{2}
& \gc{5549.86 $(+241.30)$} & \gc{2}
& \gc{3912.64 $(-0.40)$} & \gc{1} \\
& SplitLoRA
& 77.02\% $(+0.00\%)$ & 1
& 4.84 $(+0.05)$ & 1
& 3416.45 $(+0.00)$ & 1
& 234.30 $(+2.36)$ & 1
& 2018.00 $(-0.40)$ & 1 \\

\bottomrule
\end{tabular}
}

\footnotesize{\textit{Note:} J, I, P, M, and N denote Jetson, iQOO, Huawei P50, Mate 20, and Nova clients, respectively.}
\vspace{-6mm}
\end{table*}

\vspace{-2mm}
\section{Limitations}
\vspace{-2mm}

The evaluated methods do not exhaustively represent all existing or emerging algorithms. Our current platform does not fully capture other edge deployments such as wearables and embedded IoT devices. Our task suite focuses on lightweight language understanding and reasoning, while open-ended generation, multimodal interaction, and long-context personalization remain future extensions. In addition, system metrics such as energy and runtime memory can be affected by background processes and measurement-tool granularity.

\vspace{-2mm}
\section{Conclusion}
\vspace{-2mm}

In this paper, we presented \textbf{EdgeFlowerTune}, a deployment-oriented benchmark for federated LLM fine-tuning under realistic edge-system constraints. 
EdgeFlowerTune jointly evaluates model quality, system efficiency, and robustness through three complementary protocols: \textit{Quality-under-Budget}, \textit{Cost-to-Target}, and \textit{Robustness}. 
Our results on real edge devices show that benchmark conclusions can change substantially once system constraints are considered: methods that appear favorable under accuracy-only evaluation may become suboptimal once realistic edge constraints and system perturbations are taken into account. 
These findings highlight the need for system-aware and robustness-aware evaluation when developing federated LLM fine-tuning methods for edge deployment. 
We view EdgeFlowerTune as an extensible benchmark and plan to release an open leaderboard for future method submissions.

\bibliographystyle{plainnat}
\bibliography{citations}

@article{achiam2023gpt4,
  title   = {GPT-4 Technical Report},
  author  = {{OpenAI}},
  journal = {arXiv preprint arXiv:2303.08774},
  year    = {2023}
}

@inproceedings{mcmahan2017communication,
  title     = {Communication-Efficient Learning of Deep Networks from Decentralized Data},
  author    = {McMahan, H. Brendan and Moore, Eider and Ramage, Daniel and Hampson, Seth and Arcas, Blaise Ag{\"u}era y},
  booktitle = {Proceedings of the 20th International Conference on Artificial Intelligence and Statistics (AISTATS)},
  pages     = {1273--1282},
  year      = {2017}
}

@article{kairouz2021advances,
  title   = {Advances and Open Problems in Federated Learning},
  author  = {Kairouz, Peter and McMahan, H. Brendan and Avent, Brendan and Bellet, Aur{\'e}lien and Bennis, Mehdi and Bhagoji, Arjun Nitin and Bonawitz, Keith and Cormode, Graham and Cummings, Rachel and D'Oliveira, Rodrigo and others},
  journal = {Foundations and Trends in Machine Learning},
  volume  = {14},
  number  = {1--2},
  pages   = {1--210},
  year    = {2021},
  doi     = {10.1561/2200000083}
}

@inproceedings{hu2022lora,
  title     = {LoRA: Low-Rank Adaptation of Large Language Models},
  author    = {Hu, Edward J. and Shen, Yelong and Wallis, Phillip and Allen-Zhu, Zeyuan and Li, Yuanzhi and Wang, Shean and Wang, Lu and Chen, Weizhu},
  booktitle = {International Conference on Learning Representations (ICLR)},
  year      = {2022}
}

@inproceedings{kuang2024federatedscope,
  title     = {FederatedScope-LLM: A Comprehensive Package for Fine-tuning Large Language Models in Federated Learning},
  author    = {Kuang, Weirui and Qian, Bingchen and Li, Zitao and Chen, Daoyuan and Gao, Dawei and Pan, Xuchen and Xie, Yuexiang and Li, Yaliang and Ding, Bolin and Zhou, Jingren},
  booktitle = {Proceedings of the 30th ACM SIGKDD Conference on Knowledge Discovery and Data Mining},
  pages     = {5260--5271},
  year      = {2024},
  doi       = {10.1145/3637528.3671573}
}

@inproceedings{ye2024openfedllm,
  title     = {OpenFedLLM: Training Large Language Models on Decentralized Private Data via Federated Learning},
  author    = {Ye, Rui and Wang, Wenhao and Chai, Jingyi and Li, Dihan and Li, Zexi and Xu, Yinda and Du, Yaxin and Wang, Yanfeng and Chen, Siheng},
  booktitle = {Proceedings of the 30th ACM SIGKDD Conference on Knowledge Discovery and Data Mining},
  pages     = {6137--6147},
  year      = {2024},
  doi       = {10.1145/3637528.3671582}
}

@inproceedings{gao2025flowertune,
  title     = {FlowerTune: A Cross-Domain Benchmark for Federated Fine-Tuning of Large Language Models},
  author    = {Gao, Yan and Scamarcia, Massimo Roberto and Fernandez-Marques, Javier and Naseri, Mohammad and Ng, Chong Shen and Stripelis, Dimitris and Li, Zexi and Shen, Tao and Bai, Jiamu and Chen, Daoyuan and Zhang, Zikai and Hu, Rui and Song, InSeo and Lee, KangYoon and Jia, Hong and Dang, Ting and Wang, Junyan and Liu, Zheyuan and Beutel, Daniel Janes and Lyu, Lingjuan and Lane, Nicholas D.},
  booktitle = {Advances in Neural Information Processing Systems},
  year      = {2025},
  note      = {Datasets and Benchmarks Track},
  url       = {https://openreview.net/forum?id=l8Nb6ecZjW}
}

@article{zhang2023fedit,
  title={Towards Building the Federated {GPT}: Federated Instruction Tuning},
  author={Zhang, Jianyi and Vahidian, Saeed and Kuo, Martin and Li, Chunyuan and Zhang, Ruiyi and Yu, Tong and Zhou, Yufan and Wang, Guoyin and Chen, Yiran},
  journal={arXiv preprint arXiv:2305.05644},
  year={2023}
}

@inproceedings{li2020fedprox,
  title={Federated Optimization in Heterogeneous Networks},
  author={Li, Tian and Sahu, Anit Kumar and Zaheer, Manzil and Sanjabi, Maziar and Talwalkar, Ameet and Smith, Virginia},
  booktitle={Proceedings of Machine Learning and Systems},
  volume={2},
  pages={429--450},
  year={2020}
}

@inproceedings{bai2024flexlora,
  title={Federated Fine-tuning of Large Language Models under Heterogeneous Tasks and Client Resources},
  author={Bai, Jiamu and Chen, Daoyuan and Qian, Bingchen and Yao, Liuyi and Li, Yaliang},
  booktitle={Advances in Neural Information Processing Systems},
  year={2024}
}

@inproceedings{wang2024flora,
  title={{FLoRA}: Federated Fine-Tuning Large Language Models with Heterogeneous Low-Rank Adaptations},
  author={Wang, Ziyao and Shen, Zheyu and He, Yexiao and Sun, Guoheng and Wang, Hongyi and Lyu, Lingjuan and Li, Ang},
  booktitle={Advances in Neural Information Processing Systems},
  year={2024}
}

@article{liu2025hlora,
  title={{HLoRA}: Efficient Federated Learning System for {LLM} Heterogeneous Fine-Tuning},
  author={Liu, Qianli and Zhang, Zhaorui and Yao, Xin and Liu, Benben},
  journal={arXiv preprint arXiv:2503.00813},
  year={2025}
}

@article{lin2024splitlora,
  title={{SplitLoRA}: A Split Parameter-Efficient Fine-Tuning Framework for Large Language Models},
  author={Lin, Zheng and Hu, Xuanjie and Zhang, Yuxin and Chen, Zhe and Fang, Zihan and Chen, Xianhao and Li, Ang and Vepakomma, Praneeth and Gao, Yue},
  journal={arXiv preprint arXiv:2407.00952},
  year={2024}
}

@inproceedings{10.5555/3692070.3694094,
author = {Villalobos, Pablo and Ho, Anson and Sevilla, Jaime and Besiroglu, Tamay and Heim, Lennart and Hobbhahn, Marius},
title = {Position: will we run out of data? limits of LLM scaling based on human-generated data},
year = {2024},
publisher = {JMLR.org},
booktitle = {Proceedings of the 41st International Conference on Machine Learning},
articleno = {2024},
numpages = {22},
location = {Vienna, Austria},
series = {ICML'24}
}

@inproceedings{10.1145/3487552.3487863,
author = {Almeida, Mario and Laskaridis, Stefanos and Mehrotra, Abhinav and Dudziak, Lukasz and Leontiadis, Ilias and Lane, Nicholas D.},
title = {Smart at what cost? characterising mobile deep neural networks in the wild},
year = {2021},
isbn = {9781450391290},
publisher = {Association for Computing Machinery},
address = {New York, NY, USA},
url = {https://doi.org/10.1145/3487552.3487863},
doi = {10.1145/3487552.3487863},
booktitle = {Proceedings of the 21st ACM Internet Measurement Conference},
pages = {658–672},
numpages = {15},
location = {Virtual Event},
series = {IMC '21}
}

@online{GDPR2016a,
  year       = {2016},
  location   = {OJ L 119, 4.5.2016, p. 1--88},
  title      = {Regulation ({EU}) 2016/679 of the {European} {Parliament} and of the {Council}},
  url        = {https://data.europa.eu/eli/reg/2016/679/oj},
  titleaddon = {of 27 {April} 2016 on the protection of natural persons with regard to the processing of personal data and on the free movement of such data, and repealing {Directive} 95/46/{EC} ({General} {Data} {Protection} {Regulation})},
  author     = {{European Parliament} and {Council of the European Union}},
  urldate    = {2023-04-13},
}

@ARTICLE{10944288,
  author={Fan, Tao and Gu, Hanlin and Cao, Xuemei and Chan, Chee Seng and Chen, Qian and Chen, Yiqiang and Feng, Yihui and Gu, Yang and Geng, Jiaxiang and Luo, Bing and Liu, Shuoling and Ong, Win Kent and Ren, Chao and Shao, Jiaqi and Sun, Chuan and Tang, Xiaoli and Tae, Hong Xi and Tong, Yongxin and Wei, Shuyue and Wu, Fan and Xi, Wei and Xu, Mingcong and Yang, He and Yang, Xin and Yan, Jiangpeng and Yu, Hao and Yu, Han and Zhang, Teng and Zhang, Yifei and Zhang, Xiaojin and Zheng, Zhenzhe and Fan, Lixin and Yang, Qiang},
  journal={IEEE Transactions on Knowledge and Data Engineering}, 
  title={Ten Challenging Problems in Federated Foundation Models}, 
  year={2025},
  volume={37},
  number={7},
  pages={4314-4337},
  doi={10.1109/TKDE.2025.3555328}}

@article{DBLP:journals/corr/abs-1910-03771,
  author       = {Thomas Wolf and
                  Lysandre Debut and
                  Victor Sanh and
                  Julien Chaumond and
                  Clement Delangue and
                  Anthony Moi and
                  Pierric Cistac and
                  Tim Rault and
                  R{\'{e}}mi Louf and
                  Morgan Funtowicz and
                  Jamie Brew},
  title        = {HuggingFace's Transformers: State-of-the-art Natural Language Processing},
  journal      = {CoRR},
  volume       = {abs/1910.03771},
  year         = {2019},
  eprinttype   = {arXiv},
  eprint       = {1910.03771},
}

@misc{geng2025mobilefinetunerunifiedendtoendframework,
      title={MobileFineTuner: A Unified End-to-End Framework for Fine-Tuning LLMs on Mobile Phones}, 
      author={Jiaxiang Geng and Lunyu Zhao and Yiyi Lu and Bing Luo},
      year={2025},
      eprint={2512.08211},
      archivePrefix={arXiv},
      primaryClass={cs.LG},
}

@inproceedings{clark2019boolq,
  title = {{BoolQ}: Exploring the Surprising Difficulty of Natural Yes/No Questions},
  author = {Clark, Christopher and Lee, Kenton and Chang, Ming-Wei and Kwiatkowski, Tom and Collins, Michael and Toutanova, Kristina},
  booktitle = {Proceedings of the 2019 Conference of the North American Chapter of the Association for Computational Linguistics: Human Language Technologies, Volume 1 (Long and Short Papers)},
  pages = {2924--2936},
  address = {Minneapolis, Minnesota},
  publisher = {Association for Computational Linguistics},
  year = {2019},
  doi = {10.18653/v1/N19-1300}
}

@inproceedings{wang2018glue,
  title = {{GLUE}: A Multi-Task Benchmark and Analysis Platform for Natural Language Understanding},
  author = {Wang, Alex and Singh, Amanpreet and Michael, Julian and Hill, Felix and Levy, Omer and Bowman, Samuel},
  booktitle = {Proceedings of the 2018 EMNLP Workshop BlackboxNLP: Analyzing and Interpreting Neural Networks for NLP},
  pages = {353--355},
  address = {Brussels, Belgium},
  publisher = {Association for Computational Linguistics},
  year = {2018},
  doi = {10.18653/v1/W18-5446}
}

@inproceedings{rajpurkar2016squad,
  title = {{SQuAD}: 100,000+ Questions for Machine Comprehension of Text},
  author = {Rajpurkar, Pranav and Zhang, Jian and Lopyrev, Konstantin and Liang, Percy},
  booktitle = {Proceedings of the 2016 Conference on Empirical Methods in Natural Language Processing},
  pages = {2383--2392},
  address = {Austin, Texas},
  publisher = {Association for Computational Linguistics},
  year = {2016},
  doi = {10.18653/v1/D16-1264}
}

@inproceedings{bisk2020piqa,
  title = {{PIQA}: Reasoning about Physical Commonsense in Natural Language},
  author = {Bisk, Yonatan and Zellers, Rowan and Le Bras, Ronan and Gao, Jianfeng and Choi, Yejin},
  booktitle = {Proceedings of the AAAI Conference on Artificial Intelligence},
  volume = {34},
  number = {05},
  pages = {7432--7439},
  year = {2020},
  doi = {10.1609/aaai.v34i05.6239}
}

@inproceedings{zellers2019hellaswag,
  title = {{HellaSwag}: Can a Machine Really Finish Your Sentence?},
  author = {Zellers, Rowan and Holtzman, Ari and Bisk, Yonatan and Farhadi, Ali and Choi, Yejin},
  booktitle = {Proceedings of the 57th Annual Meeting of the Association for Computational Linguistics},
  pages = {4791--4800},
  address = {Florence, Italy},
  publisher = {Association for Computational Linguistics},
  year = {2019},
  doi = {10.18653/v1/P19-1472}
}

@inproceedings{sap2019socialiqa,
  title = {Social {IQa}: Commonsense Reasoning about Social Interactions},
  author = {Sap, Maarten and Rashkin, Hannah and Chen, Derek and Le Bras, Ronan and Choi, Yejin},
  booktitle = {Proceedings of the 2019 Conference on Empirical Methods in Natural Language Processing and the 9th International Joint Conference on Natural Language Processing},
  pages = {4463--4473},
  address = {Hong Kong, China},
  publisher = {Association for Computational Linguistics},
  year = {2019},
  doi = {10.18653/v1/D19-1454}
}

@article{clark2018arc,
  title = {Think you have Solved Question Answering? Try {ARC}, the {AI2} Reasoning Challenge},
  author = {Clark, Peter and Cowhey, Isaac and Etzioni, Oren and Khot, Tushar and Sabharwal, Ashish and Schoenick, Carissa and Tafjord, Oyvind},
  journal = {arXiv preprint arXiv:1803.05457},
  year = {2018}
}

@inproceedings{sakaguchi2020winogrande,
  title = {{WinoGrande}: An Adversarial Winograd Schema Challenge at Scale},
  author = {Sakaguchi, Keisuke and Le Bras, Ronan and Bhagavatula, Chandra and Choi, Yejin},
  booktitle = {Proceedings of the AAAI Conference on Artificial Intelligence},
  volume = {34},
  number = {05},
  pages = {8732--8740},
  year = {2020},
  doi = {10.1609/aaai.v34i05.6399}
}

@article{gemma3_2025,
  title        = {{Gemma 3 Technical Report}},
  author       = {Gemma Team},
  journal      = {arXiv preprint},
  volume       = {arXiv:2503.19786},
  year         = {2025},
  note         = {Google DeepMind Gemma 3 model family},
  url          = {https://arxiv.org/abs/2503.19786}
}

@article{qwen2p5_2024,
  title        = {{Qwen2.5 Technical Report}},
  author       = {A.~Yang and the Qwen Team},
  journal      = {arXiv preprint},
  volume       = {arXiv:2412.15115},
  year         = {2024},
  note         = {Qwen2.5 model family including 0.5B variant},
  url          = {https://arxiv.org/abs/2412.15115}
}

@inproceedings{brown2020language,
  title     = {Language Models are Few-Shot Learners},
  author    = {Brown, Tom B. and Mann, Benjamin and Ryder, Nick and Subbiah, Melanie and Kaplan, Jared and Dhariwal, Prafulla and Neelakantan, Arvind and Shyam, Pranav and Sastry, Girish and Askell, Amanda and Agarwal, Sandhini and Herbert-Voss, Ariel and Krueger, Gretchen and Henighan, Tom and Child, Rewon and Ramesh, Aditya and Ziegler, Daniel M. and Wu, Jeffrey and Winter, Clemens and Hesse, Christopher and Chen, Mark and Sigler, Eric and Litwin, Mateusz and Gray, Scott and Chess, Benjamin and Clark, Jack and Berner, Christopher and McCandlish, Sam and Radford, Alec and Sutskever, Ilya and Amodei, Dario},
  booktitle = {Advances in Neural Information Processing Systems},
  volume    = {33},
  pages     = {1877--1901},
  year      = {2020}
}

@inproceedings{wang2024answerc,
  title     = {``My Answer is C'': First-Token Probabilities Do Not Match Text Answers in Instruction-Tuned Language Models},
  author    = {Wang, Xinpeng and Ma, Bolei and Hu, Chengzhi and Weber-Genzel, Leon and R{\"o}ttger, Paul and Kreuter, Frauke and Hovy, Dirk and Plank, Barbara},
  booktitle = {Findings of the Association for Computational Linguistics: ACL 2024},
  year      = {2024}
}

@inproceedings{cho-etal-2024-heterogeneous,
    title = "Heterogeneous {L}o{RA} for Federated Fine-tuning of On-Device Foundation Models",
    author = "Cho, Yae Jee  and
      Liu, Luyang  and
      Xu, Zheng  and
      Fahrezi, Aldi  and
      Joshi, Gauri",
    editor = "Al-Onaizan, Yaser  and
      Bansal, Mohit  and
      Chen, Yun-Nung",
    booktitle = "Proceedings of the 2024 Conference on Empirical Methods in Natural Language Processing",
    month = nov,
    year = "2024",
    address = "Miami, Florida, USA",
    publisher = "Association for Computational Linguistics",
    url = "https://aclanthology.org/2024.emnlp-main.717/",
    doi = "10.18653/v1/2024.emnlp-main.717",
    pages = "12903--12913"
}


\clearpage

\appendix

\section{Complete benchmark results of protocol A}
\label{AppendixA}

This section provides the complete benchmark results for Protocol A, i.e., \textit{Quality-under-Budget}. 
We report the achievable quality of Qwen2.5-0.5B, Gemma 3-270M, and Gemma 3-1B on seven evaluation datasets under the given system budgets. 
For each model and dataset, we evaluate four representative federated LLM fine-tuning methods: FedAvg+LoRA, FedProx+LoRA, HeteroLoRA, and SplitLoRA. 
The reported quality metrics include testing loss, where lower is better, and testing accuracy, where higher is better.

We consider a total of 100 clients, with 20 clients instantiated for each device type listed in Table~\ref{tab:client_devices}. 
In each communication round, 10 clients are randomly selected to participate in federated fine-tuning. 
In addition to the four federated fine-tuning methods, we also report the results of the pretrained model, centroid fine-tuning, and local-only training for reference. 
The pretrained result denotes the initial zero-shot quality before fine-tuning. 
The centroid result is obtained by centralized fine-tuning on the complete training dataset using the same experimental configuration, and is treated as the quality upper bound. 
The local-only result is obtained by training each client only on its local data without federated aggregation; its testing loss and testing accuracy are reported as the average values across clients.

\begin{table*}[h]
\centering
\caption{Testing quality across all selected tasks and methods under Protocol A with Qwen2.5-0.5B.}
\label{tab:testing_quality_selected}
\renewcommand{\arraystretch}{1.12}
\setlength{\tabcolsep}{4pt}
\resizebox{\textwidth}{!}{
\begin{tabular}{lllccc ccc}
\toprule
\multirow{2}{*}{\textbf{Task Type}} 
& \multirow{2}{*}{\textbf{Task}} 
& \multirow{2}{*}{\textbf{Method}} 
& \multicolumn{3}{c}{\textbf{Testing Loss (lower is better)}} 
& \multicolumn{3}{c}{\textbf{Testing Accuracy (higher is better)}} \\
\cmidrule(lr){4-6} \cmidrule(lr){7-9}
& & 
& \textbf{Pretrained} 
& \textbf{Best Finetuned} 
& \textbf{Rank} 
& \textbf{Pretrained} 
& \textbf{Best Finetuned} 
& \textbf{Rank} \\
\midrule

\multirow{12}{*}{Verify}
& \multirow{6}{*}{BoolQ}
& \cellcolor{gray!10} Centroid     
& \multirow{6}{*}{0.6393} 
& \cellcolor{gray!10} 0.4788 
& \cellcolor{gray!10} Lower Bound 
& \multirow{6}{*}{63.21\%} 
& \cellcolor{gray!10} 80.24\% 
& \cellcolor{gray!10} Upper Bound \\
& & FedAvg+LoRA   
& & 0.5051 & 2 
& & 77.68\% & 2 \\
& & \cellcolor{gray!10} FedProx+LoRA  
& & \cellcolor{gray!10} 0.4939 
& \cellcolor{gray!10} 1 
& & \cellcolor{gray!10} 78.20\% 
& \cellcolor{gray!10} 1 \\
& & HeteroLoRA    
& & 0.5059 & 3 
& & 76.73\% & 4 \\
& & \cellcolor{gray!10} SplitLoRA     
& & \cellcolor{gray!10} 0.5161 
& \cellcolor{gray!10} 4 
& & \cellcolor{gray!10} 76.79\% 
& \cellcolor{gray!10} 3 \\
& & Local Only    
& & 0.6127 & Upper Bound 
& & 71.99\% & Lower Bound \\

\cmidrule(lr){2-9}

& \multirow{6}{*}{QNLI}
& \cellcolor{gray!10} Centroid     
& \multirow{6}{*}{1.2153} 
& \cellcolor{gray!10} 0.4611 
& \cellcolor{gray!10} Lower Bound 
& \multirow{6}{*}{58.17\%} 
& \cellcolor{gray!10} 85.76\% 
& \cellcolor{gray!10} Upper Bound \\
& & FedAvg+LoRA   
& & 0.7994 & 2 
& & 64.84\% & 3 \\
& & \cellcolor{gray!10} FedProx+LoRA  
& & \cellcolor{gray!10} 0.7357 
& \cellcolor{gray!10} 1 
& & \cellcolor{gray!10} 64.96\% 
& \cellcolor{gray!10} 2 \\
& & HeteroLoRA    
& & 0.9490 & 4 
& & 64.29\% & 4 \\
& & \cellcolor{gray!10} SplitLoRA     
& & \cellcolor{gray!10} 0.8051 
& \cellcolor{gray!10} 3 
& & \cellcolor{gray!10} 65.20\% 
& \cellcolor{gray!10} 1 \\
& & Local Only    
& & 0.8657 & Upper Bound 
& & 63.88\% & Lower Bound \\

\midrule

\multirow{18}{*}{Choose}
& \multirow{6}{*}{PIQA}
& \cellcolor{gray!10} Centroid     
& \multirow{6}{*}{0.7382} 
& \cellcolor{gray!10} 0.6778 
& \cellcolor{gray!10} Lower Bound 
& \multirow{6}{*}{59.36\%} 
& \cellcolor{gray!10} 65.45\% 
& \cellcolor{gray!10} Upper Bound \\
& & FedAvg+LoRA   
& & 0.6781 & 2 
& & 65.29\% & 2 \\
& & \cellcolor{gray!10} FedProx+LoRA  
& & \cellcolor{gray!10} 0.6779 
& \cellcolor{gray!10} 1 
& & \cellcolor{gray!10} 65.29\% 
& \cellcolor{gray!10} 2 \\
& & HeteroLoRA    
& & 0.6943 & 4 
& & 65.34\% & 1 \\
& & \cellcolor{gray!10} SplitLoRA     
& & \cellcolor{gray!10} 0.6927 
& \cellcolor{gray!10} 3 
& & \cellcolor{gray!10} 64.47\% 
& \cellcolor{gray!10} 3 \\
& & Local Only    
& & 0.7514 & Upper Bound 
& & 61.22\% & Lower Bound \\

\cmidrule(lr){2-9}

& \multirow{6}{*}{HellaSwag}
& \cellcolor{gray!10} Centroid     
& \multirow{6}{*}{1.8658} 
& \cellcolor{gray!10} 1.7026 
& \cellcolor{gray!10} Lower Bound 
& \multirow{6}{*}{26.87\%} 
& \cellcolor{gray!10} 35.13\% 
& \cellcolor{gray!10} Upper Bound \\
& & FedAvg+LoRA   
& & 1.7446 & 1 
& & 33.84\% & 1 \\
& & \cellcolor{gray!10} FedProx+LoRA  
& & \cellcolor{gray!10} 1.7446 
& \cellcolor{gray!10} 1 
& & \cellcolor{gray!10} 33.84\% 
& \cellcolor{gray!10} 1 \\
& & HeteroLoRA    
& & 1.7618 & 3 
& & 33.18\% & 3 \\
& & \cellcolor{gray!10} SplitLoRA     
& & \cellcolor{gray!10} 1.7563 
& \cellcolor{gray!10} 2 
& & \cellcolor{gray!10} 33.67\% 
& \cellcolor{gray!10} 2 \\
& & Local Only    
& & 2.0041 & Upper Bound 
& & 26.06\% & Lower Bound \\

\cmidrule(lr){2-9}

& \multirow{6}{*}{SocialIQA}
& \cellcolor{gray!10} Centroid     
& \multirow{6}{*}{0.9644} 
& \cellcolor{gray!10} 0.7769 
& \cellcolor{gray!10} Lower Bound 
& \multirow{6}{*}{55.99\%} 
& \cellcolor{gray!10} 68.07\% 
& \cellcolor{gray!10} Upper Bound \\
& & FedAvg+LoRA   
& & 0.7993 & 2 
& & 66.63\% & 2 \\
& & \cellcolor{gray!10} FedProx+LoRA  
& & \cellcolor{gray!10} 0.7863 
& \cellcolor{gray!10} 1 
& & \cellcolor{gray!10} 67.35\% 
& \cellcolor{gray!10} 1 \\
& & HeteroLoRA    
& & 0.8250 & 4 
& & 66.07\% & 4 \\
& & \cellcolor{gray!10} SplitLoRA     
& & \cellcolor{gray!10} 0.8044 
& \cellcolor{gray!10} 3 
& & \cellcolor{gray!10} 66.38\% 
& \cellcolor{gray!10} 3 \\
& & Local Only    
& & 0.9981 & Upper Bound 
& & 59.31\% & Lower Bound \\

\midrule

\multirow{12}{*}{Reason}
& \multirow{6}{*}{ARC-E}
& \cellcolor{gray!10} Centroid     
& \multirow{6}{*}{0.7180} 
& \cellcolor{gray!10} 0.5913 
& \cellcolor{gray!10} Lower Bound 
& \multirow{6}{*}{71.05\%} 
& \cellcolor{gray!10} 79.47\% 
& \cellcolor{gray!10} Upper Bound \\
& & FedAvg+LoRA   
& & 0.6054 & 2 
& & 78.60\% & 2 \\
& & \cellcolor{gray!10} FedProx+LoRA  
& & \cellcolor{gray!10} 0.6013 
& \cellcolor{gray!10} 1 
& & \cellcolor{gray!10} 78.77\% 
& \cellcolor{gray!10} 1 \\
& & HeteroLoRA    
& & 0.6274 & 4 
& & 76.67\% & 4 \\
& & \cellcolor{gray!10} SplitLoRA     
& & \cellcolor{gray!10} 0.6102 
& \cellcolor{gray!10} 3 
& & \cellcolor{gray!10} 77.02\% 
& \cellcolor{gray!10} 3 \\
& & Local Only    
& & 0.6693 & Upper Bound 
& & 74.53\% & Lower Bound \\

\cmidrule(lr){2-9}

& \multirow{6}{*}{WinoGrande}
& \cellcolor{gray!10} Centroid     
& \multirow{6}{*}{0.7032} 
& \cellcolor{gray!10} 0.6156 
& \cellcolor{gray!10} Lower Bound 
& \multirow{6}{*}{51.07\%} 
& \cellcolor{gray!10} 63.93\% 
& \cellcolor{gray!10} Upper Bound \\
& & FedAvg+LoRA   
& & 0.6905 & 2 
& & 63.46\% & 2 \\
& & \cellcolor{gray!10} FedProx+LoRA  
& & \cellcolor{gray!10} 0.6906 
& \cellcolor{gray!10} 3 
& & \cellcolor{gray!10} 63.61\% 
& \cellcolor{gray!10} 1 \\
& & HeteroLoRA    
& & 0.6960 & 4 
& & 61.17\% & 4 \\
& & \cellcolor{gray!10} SplitLoRA     
& & \cellcolor{gray!10} 0.6889 
& \cellcolor{gray!10} 1 
& & \cellcolor{gray!10} 62.43\% 
& \cellcolor{gray!10} 3 \\
& & Local Only    
& & 0.7073 & Upper Bound 
& & 56.69\% & Lower Bound \\

\bottomrule
\end{tabular}
}
\vspace{-5mm}
\end{table*}

\begin{table*}[h]
\centering
\caption{Testing quality across all selected tasks and methods under Protocol A with Gemma 3-270M.}
\label{tab:protocol_a_gemma270m}
\renewcommand{\arraystretch}{1.12}
\setlength{\tabcolsep}{4pt}
\resizebox{\textwidth}{!}{
\begin{tabular}{lllccc ccc}
\toprule
\multirow{2}{*}{\textbf{Task Type}} 
& \multirow{2}{*}{\textbf{Task}} 
& \multirow{2}{*}{\textbf{Method}} 
& \multicolumn{3}{c}{\textbf{Testing Loss (lower is better)}} 
& \multicolumn{3}{c}{\textbf{Testing Accuracy (higher is better)}} \\
\cmidrule(lr){4-6} \cmidrule(lr){7-9}
& & 
& \textbf{Pretrained} 
& \textbf{Best Finetuned} 
& \textbf{Rank} 
& \textbf{Pretrained} 
& \textbf{Best Finetuned} 
& \textbf{Rank} \\
\midrule

\multirow{12}{*}{Verify}
& \multirow{6}{*}{BoolQ}
& \cellcolor{gray!10} Centroid     
& \multirow{6}{*}{0.8543} 
& \cellcolor{gray!10} 0.6199 
& \cellcolor{gray!10} Lower Bound 
& \multirow{6}{*}{60.40\%} 
& \cellcolor{gray!10} 69.66\% 
& \cellcolor{gray!10} Upper Bound \\
& & FedAvg+LoRA   
& & 0.7356 & 3 
& & 62.94\% & 2 \\
& & \cellcolor{gray!10} FedProx+LoRA  
& & \cellcolor{gray!10} 0.7271 
& \cellcolor{gray!10} 2 
& & \cellcolor{gray!10} 64.53\% 
& \cellcolor{gray!10} 1 \\
& & HeteroLoRA    
& & 0.7441 & 4 
& & 62.20\% & 4 \\
& & \cellcolor{gray!10} SplitLoRA     
& & \cellcolor{gray!10} 0.7017 
& \cellcolor{gray!10} 1 
& & \cellcolor{gray!10} 62.72\% 
& \cellcolor{gray!10} 3 \\
& & Local Only    
& & 0.7396 & Upper Bound 
& & 60.28\% & Lower Bound \\

\cmidrule(lr){2-9}

& \multirow{6}{*}{QNLI}
& \cellcolor{gray!10} Centroid     
& \multirow{6}{*}{1.0986} 
& \cellcolor{gray!10} 0.5454 
& \cellcolor{gray!10} Lower Bound 
& \multirow{6}{*}{51.51\%} 
& \cellcolor{gray!10} 78.29\% 
& \cellcolor{gray!10} Upper Bound \\
& & FedAvg+LoRA   
& & 0.6720 & 2 
& & 68.26\% & 2 \\
& & \cellcolor{gray!10} FedProx+LoRA  
& & \cellcolor{gray!10} 0.6741 
& \cellcolor{gray!10} 3 
& & \cellcolor{gray!10} 66.98\% 
& \cellcolor{gray!10} 3 \\
& & HeteroLoRA    
& & 0.7411 & 4 
& & 63.87\% & 4 \\
& & \cellcolor{gray!10} SplitLoRA     
& & \cellcolor{gray!10} 0.5718 
& \cellcolor{gray!10} 1 
& & \cellcolor{gray!10} 71.46\% 
& \cellcolor{gray!10} 1 \\
& & Local Only    
& & 0.7056 & Upper Bound 
& & 62.93\% & Lower Bound \\

\midrule

\multirow{18}{*}{Choose}
& \multirow{6}{*}{PIQA}
& \cellcolor{gray!10} Centroid     
& \multirow{6}{*}{1.3732} 
& \cellcolor{gray!10} 1.2522 
& \cellcolor{gray!10} Lower Bound 
& \multirow{6}{*}{51.47\%} 
& \cellcolor{gray!10} 51.87\% 
& \cellcolor{gray!10} Upper Bound \\
& & FedAvg+LoRA   
& & 1.3194 & 2 
& & 50.47\% & 2 \\
& & \cellcolor{gray!10} FedProx+LoRA  
& & \cellcolor{gray!10} 1.3194 
& \cellcolor{gray!10} 2 
& & \cellcolor{gray!10} 51.47\% 
& \cellcolor{gray!10} 1 \\
& & HeteroLoRA    
& & 1.3443 & 3 
& & 51.47\% & 1 \\
& & \cellcolor{gray!10} SplitLoRA     
& & \cellcolor{gray!10} 1.3188 
& \cellcolor{gray!10} 1 
& & \cellcolor{gray!10} 51.47\% 
& \cellcolor{gray!10} 1 \\
& & Local Only    
& & 1.3534 & Upper Bound 
& & 50.24\% & Lower Bound \\

\cmidrule(lr){2-9}

& \multirow{6}{*}{HellaSwag}
& \cellcolor{gray!10} Centroid     
& \multirow{6}{*}{2.9327} 
& \cellcolor{gray!10} 0.2513 
& \cellcolor{gray!10} Lower Bound 
& \multirow{6}{*}{24.54\%} 
& \cellcolor{gray!10} 25.20\% 
& \cellcolor{gray!10} Upper Bound \\
& & FedAvg+LoRA   
& & 2.6755 & 1 
& & 24.84\% & 3 \\
& & \cellcolor{gray!10} FedProx+LoRA  
& & \cellcolor{gray!10} 2.6755 
& \cellcolor{gray!10} 1 
& & \cellcolor{gray!10} 24.84\% 
& \cellcolor{gray!10} 3 \\
& & HeteroLoRA    
& & 2.7378 & 3 
& & 24.93\% & 1 \\
& & \cellcolor{gray!10} SplitLoRA     
& & \cellcolor{gray!10} 2.6765 
& \cellcolor{gray!10} 2 
& & \cellcolor{gray!10} 24.88\% 
& \cellcolor{gray!10} 2 \\
& & Local Only    
& & 2.6982 & Upper Bound 
& & 24.73\% & Lower Bound \\

\cmidrule(lr){2-9}

& \multirow{6}{*}{SocialIQA}
& \cellcolor{gray!10} Centroid     
& \multirow{6}{*}{1.8536} 
& \cellcolor{gray!10} 1.2164 
& \cellcolor{gray!10} Lower Bound 
& \multirow{6}{*}{34.34\%} 
& \cellcolor{gray!10} 44.98\% 
& \cellcolor{gray!10} Upper Bound \\
& & FedAvg+LoRA   
& & 1.3476 & 3 
& & 40.94\% & 1 \\
& & \cellcolor{gray!10} FedProx+LoRA  
& & \cellcolor{gray!10} 1.3368 
& \cellcolor{gray!10} 2 
& & \cellcolor{gray!10} 40.38\% 
& \cellcolor{gray!10} 3 \\
& & HeteroLoRA    
& & 1.4414 & 4 
& & 38.79\% & 4 \\
& & \cellcolor{gray!10} SplitLoRA     
& & \cellcolor{gray!10} 1.2748 
& \cellcolor{gray!10} 1 
& & \cellcolor{gray!10} 40.84\% 
& \cellcolor{gray!10} 2 \\
& & Local Only    
& & 1.7288 & Upper Bound 
& & 37.86\% & Lower Bound \\

\midrule

\multirow{12}{*}{Reason}
& \multirow{6}{*}{ARC-E}
& \cellcolor{gray!10} Centroid     
& \multirow{6}{*}{2.9670} 
& \cellcolor{gray!10} 2.3865 
& \cellcolor{gray!10} Lower Bound 
& \multirow{6}{*}{26.67\%} 
& \cellcolor{gray!10} 30.90\% 
& \cellcolor{gray!10} Upper Bound \\
& & FedAvg+LoRA   
& & 2.4061 & 1 
& & 30.18\% & 1 \\
& & \cellcolor{gray!10} FedProx+LoRA  
& & \cellcolor{gray!10} 2.4061 
& \cellcolor{gray!10} 1 
& & \cellcolor{gray!10} 30.18\% 
& \cellcolor{gray!10} 1 \\
& & HeteroLoRA    
& & 2.5769 & 3 
& & 30.00\% & 2 \\
& & \cellcolor{gray!10} SplitLoRA     
& & \cellcolor{gray!10} 2.4278 
& \cellcolor{gray!10} 2 
& & \cellcolor{gray!10} 29.30\% 
& \cellcolor{gray!10} 3 \\
& & Local Only    
& & 2.5307 & Upper Bound 
& & 29.26\% & Lower Bound \\

\cmidrule(lr){2-9}

& \multirow{6}{*}{WinoGrande}
& \cellcolor{gray!10} Centroid     
& \multirow{6}{*}{1.5455} 
& \cellcolor{gray!10} 1.3361 
& \cellcolor{gray!10} Lower Bound 
& \multirow{6}{*}{49.88\%} 
& \cellcolor{gray!10} 50.59\% 
& \cellcolor{gray!10} Upper Bound \\
& & FedAvg+LoRA   
& & 1.4291 & 3 
& & 50.51\% & 2 \\
& & \cellcolor{gray!10} FedProx+LoRA  
& & \cellcolor{gray!10} 1.3735 
& \cellcolor{gray!10} 1 
& & \cellcolor{gray!10} 51.30\% 
& \cellcolor{gray!10} 1 \\
& & HeteroLoRA    
& & 1.4621 & 4 
& & 50.51\% & 2 \\
& & \cellcolor{gray!10} SplitLoRA     
& & \cellcolor{gray!10} 1.4263 
& \cellcolor{gray!10} 2 
& & \cellcolor{gray!10} 50.43\% 
& \cellcolor{gray!10} 3 \\
& & Local Only    
& & 1.4315 & Upper Bound 
& & 50.43\% & Lower Bound \\

\bottomrule
\end{tabular}
}
\vspace{-6mm}
\end{table*}

\vspace{-2mm}
\subsection{Results of Qwen2.5-0.5B}
\vspace{-2mm}

Table~\ref{tab:testing_quality_selected} summarizes the Protocol A results of Qwen2.5-0.5B across all seven tasks. 
The centroid setting provides the centralized upper bound, while the pretrained and local-only results serve as reference baselines. 
Overall, federated fine-tuning improves testing accuracy over the pretrained model on most tasks, with FedAvg+LoRA and FedProx+LoRA generally achieving strong final quality.

Comparing the four federated fine-tuning methods, FedProx+LoRA achieves the strongest accuracy on BoolQ, SocialIQA, ARC-E, and WinoGrande, and ties with FedAvg+LoRA on PIQA and HellaSwag. 
FedAvg+LoRA remains a competitive baseline, usually ranking second or tied first in accuracy, but it does not consistently outperform FedProx+LoRA. 
SplitLoRA achieves the best accuracy on QNLI and the lowest testing loss on WinoGrande.
HeteroLoRA obtains the best accuracy on PIQA but generally ranks lower on the other tasks, suggesting that heterogeneous LoRA ranks may introduce a quality trade-off. 

We note several behaviors. 
The local-only baseline can show higher testing loss than the pretrained model because each client trains only on a small local shard without aggregation, which may lead to local overfitting and weaker global generalization. 
In addition, testing loss and testing accuracy do not always induce the same ranking: loss measures target-token likelihood, whereas accuracy only evaluates whether the predicted answer letter is correct. 
Thus, a method can achieve better accuracy but worse loss, or vice versa.

\begin{table*}[h]
\centering
\caption{Testing quality across all selected tasks and methods under Protocol A with Gemma 3-1B. ``-'' indicates that the method is not executable under the system budget due to out-of-memory.}
\label{tab:protocol_a_gemma1b}
\renewcommand{\arraystretch}{1.12}
\setlength{\tabcolsep}{4pt}
\resizebox{\textwidth}{!}{
\begin{tabular}{lllccc ccc}
\toprule
\multirow{2}{*}{\textbf{Task Type}} 
& \multirow{2}{*}{\textbf{Task}} 
& \multirow{2}{*}{\textbf{Method}} 
& \multicolumn{3}{c}{\textbf{Testing Loss (lower is better)}} 
& \multicolumn{3}{c}{\textbf{Testing Accuracy (higher is better)}} \\
\cmidrule(lr){4-6} \cmidrule(lr){7-9}
& & 
& \textbf{Pretrained} 
& \textbf{Best Finetuned} 
& \textbf{Rank} 
& \textbf{Pretrained} 
& \textbf{Best Finetuned} 
& \textbf{Rank} \\
\midrule

\multirow{12}{*}{Verify}
& \multirow{6}{*}{BoolQ}
& \cellcolor{gray!10} Centroid     
& \multirow{6}{*}{1.0778} 
& \cellcolor{gray!10} 0.5800 
& \cellcolor{gray!10} Lower Bound 
& \multirow{6}{*}{59.51\%} 
& \cellcolor{gray!10} 70.00\% 
& \cellcolor{gray!10} Upper Bound \\
& & FedAvg+LoRA   
& & - & - 
& & - & - \\
& & \cellcolor{gray!10} FedProx+LoRA  
& & \cellcolor{gray!10} - 
& \cellcolor{gray!10} - 
& & \cellcolor{gray!10} - 
& \cellcolor{gray!10} - \\
& & HeteroLoRA    
& & - & - 
& & - & - \\
& & \cellcolor{gray!10} SplitLoRA     
& & \cellcolor{gray!10} 0.7590 
& \cellcolor{gray!10} 1 
& & \cellcolor{gray!10} 64.25\% 
& \cellcolor{gray!10} 1 \\
& & Local Only    
& & - & - 
& & - & - \\

\cmidrule(lr){2-9}

& \multirow{6}{*}{QNLI}
& \cellcolor{gray!10} Centroid     
& \multirow{6}{*}{1.4075} 
& \cellcolor{gray!10} 0.6451 
& \cellcolor{gray!10} Lower Bound 
& \multirow{6}{*}{49.81\%} 
& \cellcolor{gray!10} 68.00\% 
& \cellcolor{gray!10} Upper Bound \\
& & FedAvg+LoRA   
& & - & - 
& & - & - \\
& & \cellcolor{gray!10} FedProx+LoRA  
& & \cellcolor{gray!10} - 
& \cellcolor{gray!10} - 
& & \cellcolor{gray!10} - 
& \cellcolor{gray!10} - \\
& & HeteroLoRA    
& & - & - 
& & - & - \\
& & \cellcolor{gray!10} SplitLoRA     
& & \cellcolor{gray!10} 0.7367 
& \cellcolor{gray!10} 1 
& & \cellcolor{gray!10} 64.52\% 
& \cellcolor{gray!10} 1 \\
& & Local Only    
& & - & - 
& & - & - \\

\midrule

\multirow{18}{*}{Choose}
& \multirow{6}{*}{PIQA}
& \cellcolor{gray!10} Centroid     
& \multirow{6}{*}{1.1466} 
& \cellcolor{gray!10} 0.7800 
& \cellcolor{gray!10} Lower Bound 
& \multirow{6}{*}{51.52\%} 
& \cellcolor{gray!10} 57.02\% 
& \cellcolor{gray!10} Upper Bound \\
& & FedAvg+LoRA   
& & - & - 
& & - & - \\
& & \cellcolor{gray!10} FedProx+LoRA  
& & \cellcolor{gray!10} - 
& \cellcolor{gray!10} - 
& & \cellcolor{gray!10} - 
& \cellcolor{gray!10} - \\
& & HeteroLoRA    
& & - & - 
& & - & - \\
& & \cellcolor{gray!10} SplitLoRA     
& & \cellcolor{gray!10} 0.9847 
& \cellcolor{gray!10} 1 
& & \cellcolor{gray!10} 53.81\% 
& \cellcolor{gray!10} 1 \\
& & Local Only    
& & - & - 
& & - & - \\

\cmidrule(lr){2-9}

& \multirow{6}{*}{HellaSwag}
& \cellcolor{gray!10} Centroid     
& \multirow{6}{*}{2.7088} 
& \cellcolor{gray!10} 2.0173 
& \cellcolor{gray!10} Lower Bound 
& \multirow{6}{*}{25.03\%} 
& \cellcolor{gray!10} 31.65\% 
& \cellcolor{gray!10} Upper Bound \\
& & FedAvg+LoRA   
& & - & - 
& & - & - \\
& & \cellcolor{gray!10} FedProx+LoRA  
& & \cellcolor{gray!10} - 
& \cellcolor{gray!10} - 
& & \cellcolor{gray!10} - 
& \cellcolor{gray!10} - \\
& & HeteroLoRA    
& & - & - 
& & - & - \\
& & \cellcolor{gray!10} SplitLoRA     
& & \cellcolor{gray!10} 2.6109 
& \cellcolor{gray!10} 1 
& & \cellcolor{gray!10} 25.33\% 
& \cellcolor{gray!10} 1 \\
& & Local Only    
& & - & - 
& & - & - \\

\cmidrule(lr){2-9}

& \multirow{6}{*}{SocialIQA}
& \cellcolor{gray!10} Centroid     
& \multirow{6}{*}{1.8257} 
& \cellcolor{gray!10} 1.1800 
& \cellcolor{gray!10} Lower Bound 
& \multirow{6}{*}{34.44\%} 
& \cellcolor{gray!10} 49.03\% 
& \cellcolor{gray!10} Upper Bound \\
& & FedAvg+LoRA   
& & - & - 
& & - & - \\
& & \cellcolor{gray!10} FedProx+LoRA  
& & \cellcolor{gray!10} - 
& \cellcolor{gray!10} - 
& & \cellcolor{gray!10} - 
& \cellcolor{gray!10} - \\
& & HeteroLoRA    
& & - & - 
& & - & - \\
& & \cellcolor{gray!10} SplitLoRA     
& & \cellcolor{gray!10} 1.2247 
& \cellcolor{gray!10} 1 
& & \cellcolor{gray!10} 45.70\% 
& \cellcolor{gray!10} 1 \\
& & Local Only    
& & - & - 
& & - & - \\

\midrule

\multirow{12}{*}{Reason}
& \multirow{6}{*}{ARC-E}
& \cellcolor{gray!10} Centroid     
& \multirow{6}{*}{2.6107} 
& \cellcolor{gray!10} 1.9000 
& \cellcolor{gray!10} Lower Bound 
& \multirow{6}{*}{26.14\%} 
& \cellcolor{gray!10} 33.71\% 
& \cellcolor{gray!10} Upper Bound \\
& & FedAvg+LoRA   
& & - & - 
& & - & - \\
& & \cellcolor{gray!10} FedProx+LoRA  
& & \cellcolor{gray!10} - 
& \cellcolor{gray!10} - 
& & \cellcolor{gray!10} - 
& \cellcolor{gray!10} - \\
& & HeteroLoRA    
& & - & - 
& & - & - \\
& & \cellcolor{gray!10} SplitLoRA     
& & \cellcolor{gray!10} 2.2811 
& \cellcolor{gray!10} 1 
& & \cellcolor{gray!10} 28.07\% 
& \cellcolor{gray!10} 1 \\
& & Local Only    
& & - & - 
& & - & - \\

\cmidrule(lr){2-9}

& \multirow{6}{*}{WinoGrande}
& \cellcolor{gray!10} Centroid     
& \multirow{6}{*}{1.4381} 
& \cellcolor{gray!10} 0.6427 
& \cellcolor{gray!10} Lower Bound 
& \multirow{6}{*}{50.36\%} 
& \cellcolor{gray!10} 57.73\% 
& \cellcolor{gray!10} Upper Bound \\
& & FedAvg+LoRA   
& & - & - 
& & - & - \\
& & \cellcolor{gray!10} FedProx+LoRA  
& & \cellcolor{gray!10} - 
& \cellcolor{gray!10} - 
& & \cellcolor{gray!10} - 
& \cellcolor{gray!10} - \\
& & HeteroLoRA    
& & - & - 
& & - & - \\
& & \cellcolor{gray!10} SplitLoRA     
& & \cellcolor{gray!10} 0.8169 
& \cellcolor{gray!10} 1 
& & \cellcolor{gray!10} 51.62\% 
& \cellcolor{gray!10} 1 \\
& & Local Only    
& & - & - 
& & - & - \\

\bottomrule
\end{tabular}
}
\vspace{-5mm}
\end{table*}

\subsection{Results of Gemma 3-270M}

Table~\ref{tab:protocol_a_gemma270m} compares the four federated fine-tuning methods under Protocol A with Gemma 3-270M. 
FedProx+LoRA achieves the strongest or tied strongest accuracy on BoolQ, PIQA, ARC-E, and WinoGrande. 
FedAvg+LoRA is generally competitive and obtains the best accuracy on SocialIQA and tied-best accuracy on ARC-E. 
SplitLoRA often achieves the lowest testing loss, especially on BoolQ, QNLI, PIQA, and SocialIQA.

\subsection{Results of Gemma 3-1B}

Table~\ref{tab:protocol_a_gemma1b} reports the Protocol A results with Gemma 3-1B. 
Different from Qwen2.5-0.5B and Gemma 3-270M, most federated fine-tuning methods cannot be executed under the current edge-system budget when scaling to Gemma 3-1B. 
A full LoRA fine-tuning run of Gemma 3-1B requires approximately 10 GB of client-side memory, which can only be supported by the iQOO device in our testbed. 
The remaining client devices exceed their memory limits and encounter out-of-memory failures. 
As a result, FedAvg+LoRA, FedProx+LoRA, HeteroLoRA, and Local Only cannot be deployed under the current system configuration, and their entries are marked as ``-''.

In contrast, SplitLoRA remains executable because each client only hosts the first hidden layer of the backbone model, while the remaining model computation is offloaded to the server through split learning. 
Although SplitLoRA does not always achieve the highest final accuracy on smaller models, as shown in the Qwen2.5-0.5B and Gemma 3-270M results, it provides a clear deployability advantage when the model size increases. 
This result highlights an important system-level trade-off: methods with stronger quality on smaller models may become infeasible under realistic edge memory constraints, whereas split-model fine-tuning can still support larger backbone deployment within the available client-side budget.

\subsection{Overall ranking of methods under protocol A}

\begin{figure*}
\centering
\includegraphics[width=\textwidth]{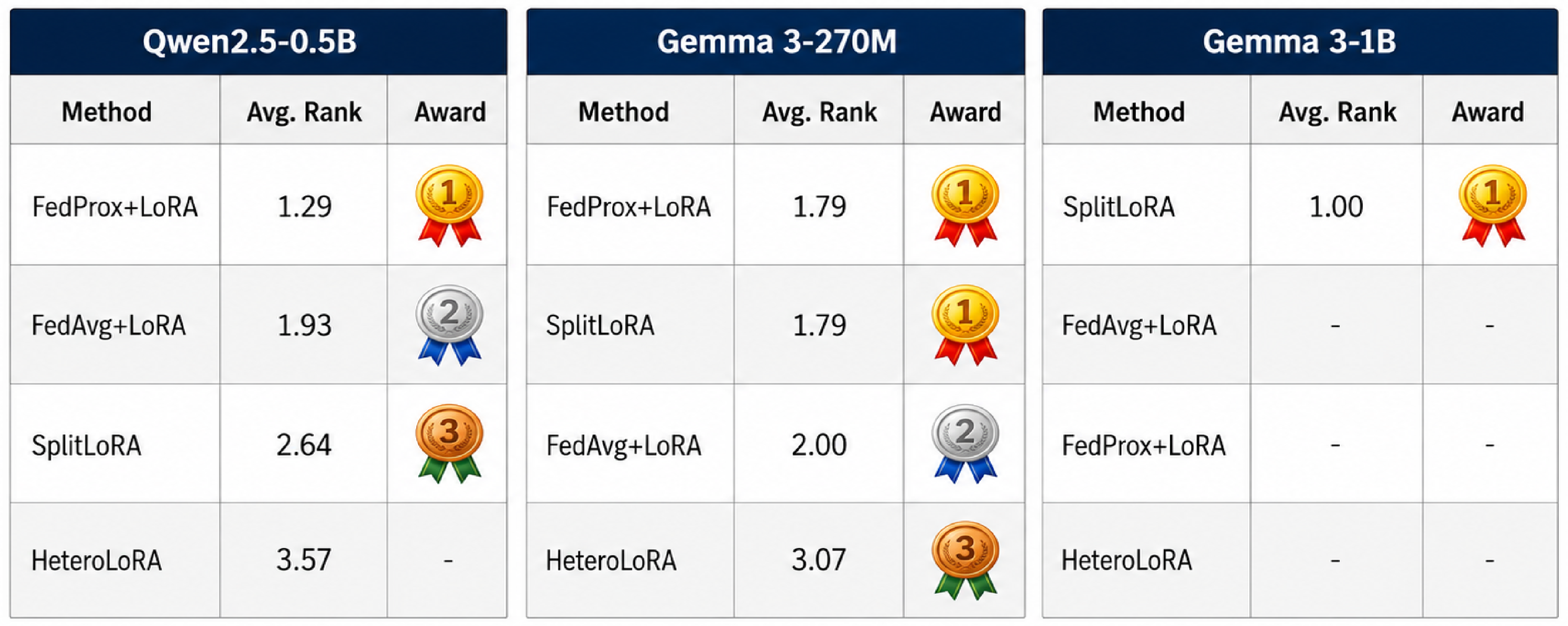}
\caption{Overall ranking of methods under protocol A. }
\label{fig:protocolAranking}
\end{figure*}

Figure~\ref{fig:protocolAranking} summarizes the overall ranking of the four federated fine-tuning methods under Protocol A. 
For Qwen2.5-0.5B and Gemma 3-270M, FedProx+LoRA achieves the best overall rank. 
This is mainly because FedProx+LoRA adds a proximal regularizer to the local LoRA objective, which helps constrain local updates and improves training stability under client-side data partitioning. 
FedAvg+LoRA also remains competitive, but it lacks this local regularization term and therefore shows slightly weaker average ranking across tasks.

HeteroLoRA obtains the lowest overall rank among the executable methods. 
Although heterogeneous LoRA ranks are designed to adapt to different device capabilities, the aggregation process requires aligning low-rank adapters to a common shape, typically through zero-padding. 
This padding-based alignment can introduce additional noise or inactive dimensions during aggregation, which may degrade the quality of the aggregated adapter. 
This observation is consistent with the findings in FLoRA~\cite{wang2024flora}, where heterogeneous adapter aggregation can introduce nontrivial quality trade-offs.

SplitLoRA does not achieve the best overall quality on Qwen2.5-0.5B and Gemma 3-270M because it only aggregates only a lightweight client-side submodel and relies on activation exchange with the server, which can affect final fine-tuning quality compared with full client-side LoRA updates. 
However, SplitLoRA shows a clear deployability advantage when scaling to Gemma 3-1B. 
Under the current edge-system budget, full LoRA-based methods cannot be executed due to client-side memory limitations, whereas SplitLoRA remains executable by keeping only the first hidden layer on the client side. 

Therefore, Figure~\ref{fig:protocolAranking} highlights a key Protocol A insight: \textit{A method may achieve strong final task performance under feasible settings, but still be impractical for edge deployment if its system cost exceeds what edge devices can sustain.}

\clearpage

\section{Complete benchmark results of protocol B}
\label{AppendixB}

This section provides the complete benchmark results for Protocol B, i.e., \textit{Cost-to-Target}. 
We use the same experimental setup as Protocol A, including the same client pool, data partitioning strategy, communication-round configuration, backbone models, datasets, and federated fine-tuning methods. 
Specifically, we consider a total of 100 clients, with 20 clients instantiated for each device type listed in Table~\ref{tab:client_devices}, and randomly select 10 clients to participate in each communication round.

Different from Protocol A, which reports the best achievable quality under a fixed system budget, Protocol B evaluates the system cost required for each method to reach a given target accuracy. 
For each task, we set three target accuracy levels to represent different stages of training progress. 
These targets are derived from the improvement interval between the pretrained model and the centroid upper bound, corresponding to 50\%, 70\%, and 90\% of the accuracy improvement from the pretrained baseline to the centroid result. 
For each target accuracy, we report four system metrics when each method first reaches the corresponding target: wall-clock time, communication volume, energy consumption, and peak memory. 
The wall-clock time, communication volume, and energy consumption are recorded at the first point where the target accuracy is achieved, while peak memory is computed as the average peak memory across all participating clients up to that point. 
Since all metrics in Protocol B represent system cost, lower values are better, and a smaller rank indicates a more efficient method.

\begin{table*}[h]
\centering
\caption{Complete Protocol B results of Qwen2.5-0.5B on Choose tasks.}
\label{tab:protocol_b_qwen25_choose}
\renewcommand{\arraystretch}{1.08}
\setlength{\tabcolsep}{3pt}
\scriptsize
\providecommand{\gc}[1]{\cellcolor{gray!10}#1}
\resizebox{\textwidth}{!}{
\begin{tabular}{lclcccccccc}
\toprule
\textbf{Task} 
& \textbf{Target Accuracy} 
& \textbf{Methods} 
& \makecell{\textbf{Wall-clock}\\\textbf{time (hour)}} 
& \textbf{Rank} 
& \makecell{\textbf{Communication}\\\textbf{volume (MB)}} 
& \textbf{Rank} 
& \makecell{\textbf{Energy}\\\textbf{consumption (kJ)}} 
& \textbf{Rank} 
& \makecell{\textbf{Peak}\\\textbf{memory (MB)}} 
& \textbf{Rank} \\
\midrule

\multirow{12}{*}{PIQA}
& \multirow{4}{*}{63\%}
& \gc{FedAvg+LoRA}  & \gc{5.02} & \gc{4} & \gc{4968.28} & \gc{2} & \gc{232.97} & \gc{4} & \gc{3432.59} & \gc{4} \\
& & FedProx+LoRA & 5.01 & 3 & 4968.28 & 2 & 232.21 & 3 & 3398.65 & 3 \\
& & \gc{HeteroLoRA} & \gc{2.05} & \gc{2} & \gc{6217.97} & \gc{3} & \gc{95.18} & \gc{2} & \gc{2531.45} & \gc{2} \\
& & SplitLoRA & 0.18 & 1 & 4804.38 & 1 & 8.47 & 1 & 1137.96 & 1 \\

\cmidrule(lr){2-11}

& \multirow{4}{*}{64\%}
& \gc{FedAvg+LoRA}  & \gc{7.52} & \gc{3} & \gc{7452.42} & \gc{1} & \gc{349.03} & \gc{3} & \gc{3473.55} & \gc{4} \\
& & FedProx+LoRA & 8.33 & 4 & 8280.47 & 3 & 386.66 & 4 & 3439.22 & 3 \\
& & \gc{HeteroLoRA} & \gc{5.94} & \gc{2} & \gc{18032.11} & \gc{4} & \gc{275.42} & \gc{2} & \gc{2561.67} & \gc{2} \\
& & SplitLoRA & 0.30 & 1 & 8007.30 & 2 & 14.12 & 1 & 1151.54 & 1 \\

\cmidrule(lr){2-11}

& \multirow{4}{*}{65\%}
& \gc{FedAvg+LoRA}  & \gc{11.68} & \gc{2} & \gc{11592.66} & \gc{1} & \gc{541.71} & \gc{2} & \gc{3492.19} & \gc{4} \\
& & FedProx+LoRA & 17.58 & 4 & 17388.98 & 2 & 815.72 & 4 & 3457.67 & 3 \\
& & \gc{HeteroLoRA} & \gc{12.70} & \gc{3} & \gc{38551.41} & \gc{4} & \gc{589.29} & \gc{3} & \gc{2575.41} & \gc{2} \\
& & SplitLoRA & 0.79 & 1 & 20818.98 & 3 & 36.72 & 1 & 1157.72 & 1 \\

\midrule

\multirow{12}{*}{HellaSwag}
& \multirow{4}{*}{30\%}
& \gc{FedAvg+LoRA}  & \gc{15.84} & \gc{3} & \gc{15732.89} & \gc{2} & \gc{770.04} & \gc{3} & \gc{3368.58} & \gc{3} \\
& & FedProx+LoRA & 15.87 & 4 & 15732.89 & 2 & 771.13 & 4 & 3539.18 & 4 \\
& & \gc{HeteroLoRA} & \gc{7.38} & \gc{2} & \gc{22384.69} & \gc{3} & \gc{358.58} & \gc{2} & \gc{2587.70} & \gc{2} \\
& & SplitLoRA & 0.45 & 1 & 10142.58 & 1 & 21.70 & 1 & 1150.11 & 1 \\

\cmidrule(lr){2-11}

& \multirow{4}{*}{32\%}
& \gc{FedAvg+LoRA}  & \gc{30.03} & \gc{4} & \gc{29809.69} & \gc{2} & \gc{1459.38} & \gc{4} & \gc{3408.78} & \gc{3} \\
& & FedProx+LoRA & 30.02 & 3 & 29809.69 & 2 & 1458.96 & 3 & 3581.42 & 4 \\
& & \gc{HeteroLoRA} & \gc{12.31} & \gc{2} & \gc{37307.81} & \gc{3} & \gc{598.08} & \gc{2} & \gc{2618.58} & \gc{2} \\
& & SplitLoRA & 0.85 & 1 & 19217.52 & 1 & 41.15 & 1 & 1163.83 & 1 \\

\cmidrule(lr){2-11}

& \multirow{4}{*}{33\%}
& \gc{FedAvg+LoRA}  & \gc{73.99} & \gc{3} & \gc{73696.17} & \gc{2} & \gc{3596.01} & \gc{3} & \gc{3427.08} & \gc{3} \\
& & FedProx+LoRA & 74.10 & 4 & 73696.17 & 2 & 3601.39 & 4 & 3600.64 & 4 \\
& & \gc{HeteroLoRA} & \gc{25.23} & \gc{2} & \gc{76481.02} & \gc{3} & \gc{1226.35} & \gc{2} & \gc{2632.64} & \gc{2} \\
& & SplitLoRA & 2.11 & 1 & 48043.80 & 1 & 102.63 & 1 & 1170.08 & 1 \\

\midrule

\multirow{12}{*}{SocialIQA}
& \multirow{4}{*}{62\%}
& \gc{FedAvg+LoRA}  & \gc{15.79} & \gc{4} & \gc{15732.89} & \gc{2} & \gc{772.17} & \gc{4} & \gc{3418.89} & \gc{3} \\
& & FedProx+LoRA & 15.71 & 3 & 15732.89 & 2 & 768.41 & 3 & 3462.20 & 4 \\
& & \gc{HeteroLoRA} & \gc{5.32} & \gc{2} & \gc{16166.72} & \gc{3} & \gc{259.95} & \gc{2} & \gc{2562.68} & \gc{2} \\
& & SplitLoRA & 0.44 & 1 & 10142.58 & 1 & 21.54 & 1 & 1147.95 & 1 \\

\cmidrule(lr){2-11}

& \multirow{4}{*}{64\%}
& \gc{FedAvg+LoRA}  & \gc{34.91} & \gc{4} & \gc{34777.97} & \gc{3} & \gc{1707.33} & \gc{4} & \gc{3455.95} & \gc{3} \\
& & FedProx+LoRA & 34.72 & 3 & 34777.97 & 3 & 1697.93 & 3 & 3499.73 & 4 \\
& & \gc{HeteroLoRA} & \gc{9.98} & \gc{2} & \gc{30468.05} & \gc{2} & \gc{488.07} & \gc{2} & \gc{2590.46} & \gc{2} \\
& & SplitLoRA & 0.81 & 1 & 18683.70 & 1 & 39.61 & 1 & 1160.39 & 1 \\

\cmidrule(lr){2-11}

& \multirow{4}{*}{66\%}
& \gc{FedAvg+LoRA}  & \gc{94.94} & \gc{4} & \gc{94397.34} & \gc{3} & \gc{4643.29} & \gc{4} & \gc{3476.19} & \gc{3} \\
& & FedProx+LoRA & 94.25 & 3 & 94397.34 & 3 & 4609.67 & 3 & 3520.23 & 4 \\
& & \gc{HeteroLoRA} & \gc{29.94} & \gc{2} & \gc{91404.14} & \gc{2} & \gc{1464.33} & \gc{2} & \gc{2605.63} & \gc{2} \\
& & SplitLoRA & 1.34 & 1 & 30961.56 & 1 & 65.53 & 1 & 1167.19 & 1 \\

\bottomrule
\end{tabular}
}
\vspace{5mm}
\end{table*}

\begin{table*}[h]
\centering
\caption{Complete Protocol B results of Qwen2.5-0.5B on Verify tasks.}
\label{tab:protocol_b_qwen25_verify}
\renewcommand{\arraystretch}{1.08}
\setlength{\tabcolsep}{3pt}
\scriptsize
\providecommand{\gc}[1]{\cellcolor{gray!10}#1}
\resizebox{\textwidth}{!}{
\begin{tabular}{lclcccccccc}
\toprule
\textbf{Task} 
& \textbf{Target Accuracy} 
& \textbf{Methods} 
& \makecell{\textbf{Wall-clock}\\\textbf{time (hour)}} 
& \textbf{Rank} 
& \makecell{\textbf{Communication}\\\textbf{volume (MB)}} 
& \textbf{Rank} 
& \makecell{\textbf{Energy}\\\textbf{consumption (kJ)}} 
& \textbf{Rank} 
& \makecell{\textbf{Peak}\\\textbf{memory (MB)}} 
& \textbf{Rank} \\
\midrule

\multirow{12}{*}{BoolQ}
& \multirow{4}{*}{71\%}
& \gc{FedAvg+LoRA}  & \gc{6.65} & \gc{3} & \gc{6624.38} & \gc{1} & \gc{300.67} & \gc{3} & \gc{3453.60} & \gc{3} \\
& & FedProx+LoRA & 6.67 & 4 & 6624.38 & 1 & 301.19 & 4 & 3470.35 & 4 \\
& & \gc{HeteroLoRA} & \gc{3.07} & \gc{2} & \gc{9326.95} & \gc{2} & \gc{138.66} & \gc{2} & \gc{2499.21} & \gc{2} \\
& & SplitLoRA & 0.43 & 1 & 11744.04 & 3 & 19.42 & 1 & 1142.87 & 1 \\

\cmidrule(lr){2-11}

& \multirow{4}{*}{75\%}
& \gc{FedAvg+LoRA}  & \gc{12.48} & \gc{3} & \gc{12420.70} & \gc{1} & \gc{563.76} & \gc{3} & \gc{3497.81} & \gc{3} \\
& & FedProx+LoRA & 12.50 & 4 & 12420.70 & 1 & 564.83 & 4 & 3514.77 & 4 \\
& & \gc{HeteroLoRA} & \gc{5.32} & \gc{2} & \gc{16166.72} & \gc{3} & \gc{240.57} & \gc{2} & \gc{2531.20} & \gc{2} \\
& & SplitLoRA & 0.47 & 1 & 12811.68 & 2 & 21.18 & 1 & 1157.50 & 1 \\

\cmidrule(lr){2-11}

& \multirow{4}{*}{78\%}
& \gc{FedAvg+LoRA}  & \gc{32.46} & \gc{3} & \gc{32293.83} & \gc{2} & \gc{1466.74} & \gc{3} & \gc{3520.44} & \gc{3} \\
& & FedProx+LoRA & 32.50 & 4 & 32293.83 & 2 & 1468.52 & 4 & 3537.51 & 4 \\
& & \gc{HeteroLoRA} & \gc{13.33} & \gc{2} & \gc{40416.80} & \gc{3} & \gc{602.20} & \gc{2} & \gc{2547.57} & \gc{2} \\
& & SplitLoRA & 0.99 & 1 & 27224.82 & 1 & 44.94 & 1 & 1164.99 & 1 \\

\midrule

\multirow{12}{*}{QNLI}
& \multirow{4}{*}{62\%}
& \gc{FedAvg+LoRA}  & \gc{2.50} & \gc{4} & \gc{2484.14} & \gc{2} & \gc{116.24} & \gc{4} & \gc{3375.58} & \gc{3} \\
& & FedProx+LoRA & 2.49 & 3 & 2484.14 & 2 & 115.91 & 3 & 3497.83 & 4 \\
& & \gc{HeteroLoRA} & \gc{1.03} & \gc{2} & \gc{3108.98} & \gc{3} & \gc{48.17} & \gc{2} & \gc{2510.74} & \gc{2} \\
& & SplitLoRA & 0.07 & 1 & 1601.46 & 1 & 3.43 & 1 & 1152.89 & 1 \\

\cmidrule(lr){2-11}

& \multirow{4}{*}{63\%}
& \gc{FedAvg+LoRA}  & \gc{18.28} & \gc{4} & \gc{18217.03} & \gc{2} & \gc{851.18} & \gc{4} & \gc{3415.86} & \gc{3} \\
& & FedProx+LoRA & 18.22 & 3 & 18217.03 & 2 & 848.50 & 3 & 3539.57 & 4 \\
& & \gc{HeteroLoRA} & \gc{3.71} & \gc{2} & \gc{11192.34} & \gc{1} & \gc{172.76} & \gc{2} & \gc{2540.71} & \gc{2} \\
& & SplitLoRA & 1.97 & 1 & 42705.60 & 3 & 91.54 & 1 & 1166.65 & 1 \\

\cmidrule(lr){2-11}

& \multirow{4}{*}{64\%}
& \gc{FedAvg+LoRA}  & \gc{53.12} & \gc{3} & \gc{52995.00} & \gc{2} & \gc{2474.11} & \gc{3} & \gc{3434.19} & \gc{3} \\
& & FedProx+LoRA & 73.59 & 4 & 73696.17 & 3 & 3427.23 & 4 & 3558.57 & 4 \\
& & \gc{HeteroLoRA} & \gc{16.29} & \gc{2} & \gc{49121.95} & \gc{1} & \gc{758.47} & \gc{2} & \gc{2554.34} & \gc{2} \\
& & SplitLoRA & 3.78 & 1 & 82208.27 & 4 & 176.19 & 1 & 1172.91 & 1 \\

\bottomrule
\end{tabular}
}
\end{table*}

\begin{table*}[h]
\centering
\caption{Complete Protocol B results of Qwen2.5-0.5B on Reason tasks.}
\label{tab:protocol_b_qwen25_reason}
\renewcommand{\arraystretch}{1.08}
\setlength{\tabcolsep}{3pt}
\scriptsize
\providecommand{\gc}[1]{\cellcolor{gray!10}#1}
\resizebox{\textwidth}{!}{
\begin{tabular}{lclcccccccc}
\toprule
\textbf{Task} 
& \textbf{Target Accuracy} 
& \textbf{Methods} 
& \makecell{\textbf{Wall-clock}\\\textbf{time (hour)}} 
& \textbf{Rank} 
& \makecell{\textbf{Communication}\\\textbf{volume (MB)}} 
& \textbf{Rank} 
& \makecell{\textbf{Energy}\\\textbf{consumption (kJ)}} 
& \textbf{Rank} 
& \makecell{\textbf{Peak}\\\textbf{memory (MB)}} 
& \textbf{Rank} \\
\midrule

\multirow{12}{*}{ARC-E}
& \multirow{4}{*}{75\%}
& \gc{FedAvg+LoRA}  & \gc{24.29} & \gc{4} & \gc{24013.36} & \gc{2} & \gc{1176.40} & \gc{4} & \gc{3404.12} & \gc{3} \\
& & FedProx+LoRA & 24.13 & 3 & 24013.36 & 2 & 1168.56 & 3 & 3441.75 & 4 \\
& & \gc{HeteroLoRA} & \gc{4.00} & \gc{2} & \gc{24250.08} & \gc{3} & \gc{193.52} & \gc{2} & \gc{2540.26} & \gc{2} \\
& & SplitLoRA & 0.48 & 1 & 10676.40 & 1 & 23.09 & 1 & 1141.11 & 1 \\

\cmidrule(lr){2-11}

& \multirow{4}{*}{76\%}
& \gc{FedAvg+LoRA}  & \gc{32.64} & \gc{3} & \gc{32293.83} & \gc{2} & \gc{1580.47} & \gc{3} & \gc{3445.54} & \gc{3} \\
& & FedProx+LoRA & 33.31 & 4 & 33121.88 & 3 & 1612.96 & 4 & 3483.62 & 4 \\
& & \gc{HeteroLoRA} & \gc{7.89} & \gc{2} & \gc{47878.36} & \gc{4} & \gc{382.26} & \gc{2} & \gc{2571.17} & \gc{2} \\
& & SplitLoRA & 0.95 & 1 & 21352.80 & 1 & 46.20 & 1 & 1154.99 & 1 \\

\cmidrule(lr){2-11}

& \multirow{4}{*}{77\%}
& \gc{FedAvg+LoRA}  & \gc{66.91} & \gc{4} & \gc{66243.75} & \gc{2} & \gc{3240.10} & \gc{4} & \gc{3458.55} & \gc{3} \\
& & FedProx+LoRA & 66.58 & 3 & 66243.75 & 2 & 3223.88 & 3 & 3496.77 & 4 \\
& & \gc{HeteroLoRA} & \gc{11.17} & \gc{2} & \gc{67775.86} & \gc{3} & \gc{540.93} & \gc{2} & \gc{2580.88} & \gc{2} \\
& & SplitLoRA & 1.53 & 1 & 34164.48 & 1 & 73.86 & 1 & 1159.35 & 1 \\

\midrule

\multirow{12}{*}{WinoGrande}
& \multirow{4}{*}{57\%}
& \gc{FedAvg+LoRA}  & \gc{10.80} & \gc{3} & \gc{10764.61} & \gc{3} & \gc{546.09} & \gc{3} & \gc{3373.35} & \gc{3} \\
& & FedProx+LoRA & 10.83 & 4 & 10764.61 & 3 & 547.75 & 4 & 3437.81 & 4 \\
& & \gc{HeteroLoRA} & \gc{2.46} & \gc{2} & \gc{7461.56} & \gc{2} & \gc{124.64} & \gc{2} & \gc{2520.73} & \gc{2} \\
& & SplitLoRA & 0.18 & 1 & 5338.20 & 1 & 9.32 & 1 & 1133.06 & 1 \\

\cmidrule(lr){2-11}

& \multirow{4}{*}{60\%}
& \gc{FedAvg+LoRA}  & \gc{39.91} & \gc{3} & \gc{39746.25} & \gc{3} & \gc{2018.83} & \gc{3} & \gc{3413.61} & \gc{3} \\
& & FedProx+LoRA & 46.43 & 4 & 46370.62 & 4 & 2348.84 & 4 & 3478.84 & 4 \\
& & \gc{HeteroLoRA} & \gc{9.46} & \gc{2} & \gc{28602.66} & \gc{2} & \gc{478.32} & \gc{2} & \gc{2550.81} & \gc{2} \\
& & SplitLoRA & 0.74 & 1 & 21352.80 & 1 & 37.30 & 1 & 1146.59 & 1 \\

\cmidrule(lr){2-11}

& \multirow{4}{*}{62\%}
& \gc{FedAvg+LoRA}  & \gc{99.43} & \gc{4} & \gc{99365.62} & \gc{2} & \gc{5029.80} & \gc{4} & \gc{3431.93} & \gc{3} \\
& & FedProx+LoRA & 99.39 & 3 & 99365.62 & 2 & 5027.48 & 3 & 3497.51 & 4 \\
& & \gc{HeteroLoRA} & \gc{34.68} & \gc{2} & \gc{105083.67} & \gc{3} & \gc{1754.18} & \gc{2} & \gc{2564.50} & \gc{2} \\
& & SplitLoRA & 1.38 & 1 & 40036.50 & 1 & 69.84 & 1 & 1152.74 & 1 \\

\bottomrule
\end{tabular}
}
\end{table*}

\subsection{Results of Qwen2.5-0.5B}

Tables~\ref{tab:protocol_b_qwen25_choose}--\ref{tab:protocol_b_qwen25_reason} report the complete Protocol B results of Qwen2.5-0.5B across all selected tasks and target accuracy levels. 
Overall, SplitLoRA consistently reaches the target accuracy with the lowest wall-clock time and energy consumption across Verify, Choose, and Reason tasks, and it also achieves the lowest communication volume in most settings. 
This indicates that, although SplitLoRA is not always the strongest method in final quality under Protocol A, it is highly efficient in reaching intermediate target accuracies under realistic edge-system budgets. 
FedAvg+LoRA and FedProx+LoRA sometimes achieve competitive communication cost because they only exchange LoRA updates, but they usually require substantially longer wall-clock time and higher energy consumption to reach the same target accuracy. 
HeteroLoRA often reduces wall-clock time compared with FedAvg+LoRA and FedProx+LoRA, but its communication and energy costs can still be high, especially at higher target accuracy levels.

\begin{table*}[h]
\centering
\caption{Complete Protocol B results of Gemma 3-270M on Verify tasks.}
\label{tab:protocol_b_gemma270m_verify}
\renewcommand{\arraystretch}{1.08}
\setlength{\tabcolsep}{3pt}
\scriptsize
\providecommand{\gc}[1]{\cellcolor{gray!10}#1}
\resizebox{\textwidth}{!}{
\begin{tabular}{lclcccccccc}
\toprule
\textbf{Task} 
& \textbf{Target Accuracy} 
& \textbf{Methods} 
& \makecell{\textbf{Wall-clock}\\\textbf{time (hour)}} 
& \textbf{Rank} 
& \makecell{\textbf{Communication}\\\textbf{volume (MB)}} 
& \textbf{Rank} 
& \makecell{\textbf{Energy}\\\textbf{consumption (kJ)}} 
& \textbf{Rank} 
& \makecell{\textbf{Peak}\\\textbf{memory (MB)}} 
& \textbf{Rank} \\
\midrule

\multirow{12}{*}{BoolQ}
& \multirow{4}{*}{62\%}
& \gc{FedAvg+LoRA}  & \gc{21.88} & \gc{3} & \gc{10166.13} & \gc{1} & \gc{1137.31} & \gc{3} & \gc{1339.46} & \gc{2} \\
& & FedProx+LoRA & 35.02 & 4 & 41229.32 & 3 & 1819.79 & 4 & 2324.80 & 4 \\
& & \gc{HeteroLoRA} & \gc{14.03} & \gc{2} & \gc{50899.22} & \gc{4} & \gc{729.35} & \gc{2} & \gc{1403.02} & \gc{3} \\
& & SplitLoRA & 0.62 & 1 & 14977.36 & 2 & 31.98 & 1 & 1002.91 & 1 \\

\cmidrule(lr){2-11}

& \multirow{4}{*}{63\%}
& \gc{FedAvg+LoRA}  & \gc{89.88} & \gc{4} & \gc{41794.10} & \gc{2} & \gc{4671.20} & \gc{4} & \gc{1356.61} & \gc{2} \\
& & FedProx+LoRA & 50.36 & 3 & 59302.44 & 3 & 2617.11 & 3 & 2354.56 & 4 \\
& & \gc{HeteroLoRA} & \gc{16.37} & \gc{2} & \gc{59382.42} & \gc{4} & \gc{850.83} & \gc{2} & \gc{1420.98} & \gc{3} \\
& & SplitLoRA & 0.71 & 1 & 17223.96 & 1 & 36.75 & 1 & 1015.75 & 1 \\

\cmidrule(lr){2-11}

& \multirow{4}{*}{64\%}
& \gc{FedAvg+LoRA}  & \gc{104.47} & \gc{4} & \gc{48571.52} & \gc{2} & \gc{5429.43} & \gc{4} & \gc{1365.38} & \gc{2} \\
& & FedProx+LoRA & 70.98 & 3 & 83588.20 & 4 & 3688.79 & 3 & 2369.79 & 4 \\
& & \gc{HeteroLoRA} & \gc{16.37} & \gc{2} & \gc{59382.42} & \gc{3} & \gc{850.83} & \gc{2} & \gc{1430.17} & \gc{3} \\
& & SplitLoRA & 0.71 & 1 & 17223.96 & 1 & 36.75 & 1 & 1022.32 & 1 \\

\midrule

\multirow{12}{*}{QNLI}
& \multirow{4}{*}{61\%}
& \gc{FedAvg+LoRA}  & \gc{15.75} & \gc{4} & \gc{18637.91} & \gc{3} & \gc{822.83} & \gc{4} & \gc{2365.36} & \gc{3} \\
& & FedProx+LoRA & 14.31 & 3 & 16943.55 & 2 & 747.48 & 3 & 2382.52 & 4 \\
& & \gc{HeteroLoRA} & \gc{4.09} & \gc{2} & \gc{14845.61} & \gc{1} & \gc{213.65} & \gc{2} & \gc{1450.77} & \gc{2} \\
& & SplitLoRA & 1.15 & 1 & 22507.10 & 4 & 60.05 & 1 & 1018.20 & 1 \\

\cmidrule(lr){2-11}

& \multirow{4}{*}{65\%}
& \gc{FedAvg+LoRA}  & \gc{31.02} & \gc{3} & \gc{36711.04} & \gc{3} & \gc{1620.23} & \gc{3} & \gc{2393.59} & \gc{3} \\
& & FedProx+LoRA & 32.43 & 4 & 38405.39 & 4 & 1693.70 & 4 & 2410.95 & 4 \\
& & \gc{HeteroLoRA} & \gc{6.42} & \gc{2} & \gc{23328.81} & \gc{1} & \gc{335.27} & \gc{2} & \gc{1468.08} & \gc{2} \\
& & SplitLoRA & 1.81 & 1 & 35477.29 & 2 & 94.73 & 1 & 1030.35 & 1 \\

\cmidrule(lr){2-11}

& \multirow{4}{*}{69\%}
& \gc{FedAvg+LoRA}  & \gc{62.42} & \gc{4} & \gc{73986.86} & \gc{4} & \gc{3260.63} & \gc{4} & \gc{2406.44} & \gc{3} \\
& & FedProx+LoRA & 45.35 & 3 & 53654.59 & 2 & 2368.79 & 3 & 2423.89 & 4 \\
& & \gc{HeteroLoRA} & \gc{17.72} & \gc{2} & \gc{64472.34} & \gc{3} & \gc{925.60} & \gc{2} & \gc{1475.96} & \gc{2} \\
& & SplitLoRA & 2.07 & 1 & 40436.48 & 1 & 107.93 & 1 & 1035.88 & 1 \\

\bottomrule
\end{tabular}
}
\end{table*}

\begin{table*}[h]
\centering
\caption{Complete Protocol B results of Gemma 3-270M on Reason tasks.}
\label{tab:protocol_b_gemma270m_reason}
\renewcommand{\arraystretch}{1.08}
\setlength{\tabcolsep}{3pt}
\scriptsize
\providecommand{\gc}[1]{\cellcolor{gray!10}#1}
\resizebox{\textwidth}{!}{
\begin{tabular}{lclcccccccc}
\toprule
\textbf{Task} 
& \textbf{Target Accuracy} 
& \textbf{Methods} 
& \makecell{\textbf{Wall-clock}\\\textbf{time (hour)}} 
& \textbf{Rank} 
& \makecell{\textbf{Communication}\\\textbf{volume (MB)}} 
& \textbf{Rank} 
& \makecell{\textbf{Energy}\\\textbf{consumption (kJ)}} 
& \textbf{Rank} 
& \makecell{\textbf{Peak}\\\textbf{memory (MB)}} 
& \textbf{Rank} \\
\midrule

\multirow{12}{*}{ARC-E}
& \multirow{4}{*}{28\%}
& \gc{FedAvg+LoRA}  & \gc{29.95} & \gc{3} & \gc{35581.46} & \gc{2} & \gc{1331.65} & \gc{3} & \gc{2238.33} & \gc{3} \\
& & FedProx+LoRA & 30.15 & 4 & 35581.46 & 2 & 1340.68 & 4 & 2274.42 & 4 \\
& & \gc{HeteroLoRA} & \gc{21.09} & \gc{2} & \gc{37326.09} & \gc{3} & \gc{938.00} & \gc{2} & \gc{1363.60} & \gc{2} \\
& & SplitLoRA & 1.78 & 1 & 24795.95 & 1 & 79.11 & 1 & 957.83 & 1 \\

\cmidrule(lr){2-11}

& \multirow{4}{*}{29\%}
& \gc{FedAvg+LoRA}  & \gc{32.80} & \gc{3} & \gc{38970.18} & \gc{2} & \gc{1458.55} & \gc{3} & \gc{2265.56} & \gc{3} \\
& & FedProx+LoRA & 33.01 & 4 & 38970.18 & 2 & 1467.81 & 4 & 2302.09 & 4 \\
& & \gc{HeteroLoRA} & \gc{26.35} & \gc{2} & \gc{46657.62} & \gc{3} & \gc{1171.86} & \gc{2} & \gc{1380.19} & \gc{2} \\
& & SplitLoRA & 1.86 & 1 & 25940.38 & 1 & 82.74 & 1 & 969.48 & 1 \\

\cmidrule(lr){2-11}

& \multirow{4}{*}{30\%}
& \gc{FedAvg+LoRA}  & \gc{37.11} & \gc{2} & \gc{44053.24} & \gc{2} & \gc{1650.23} & \gc{2} & \gc{2274.11} & \gc{3} \\
& & FedProx+LoRA & 37.33 & 3 & 44053.24 & 2 & 1660.18 & 3 & 2310.78 & 4 \\
& & \gc{HeteroLoRA} & \gc{39.01} & \gc{4} & \gc{69138.11} & \gc{3} & \gc{1734.87} & \gc{4} & \gc{1385.40} & \gc{2} \\
& & SplitLoRA & 1.97 & 1 & 27466.29 & 1 & 87.61 & 1 & 973.14 & 1 \\

\midrule

\multirow{12}{*}{WinoGrande}
& \multirow{4}{*}{51\%}
& \gc{FedAvg+LoRA}  & \gc{4.30} & \gc{3} & \gc{5083.07} & \gc{2} & \gc{194.76} & \gc{3} & \gc{2327.35} & \gc{3} \\
& & FedProx+LoRA & 21.11 & 4 & 24850.55 & 4 & 956.24 & 4 & 2330.59 & 4 \\
& & \gc{HeteroLoRA} & \gc{1.75} & \gc{2} & \gc{6362.40} & \gc{3} & \gc{79.40} & \gc{2} & \gc{1365.71} & \gc{2} \\
& & SplitLoRA & 0.19 & 1 & 3433.29 & 1 & 8.73 & 1 & 961.43 & 1 \\

\cmidrule(lr){2-11}

& \multirow{4}{*}{51\%}
& \gc{FedAvg+LoRA}  & \gc{4.78} & \gc{3} & \gc{5647.85} & \gc{2} & \gc{216.42} & \gc{3} & \gc{2355.12} & \gc{3} \\
& & FedProx+LoRA & 23.03 & 4 & 27109.69 & 4 & 1043.38 & 4 & 2358.40 & 4 \\
& & \gc{HeteroLoRA} & \gc{2.22} & \gc{2} & \gc{8059.04} & \gc{3} & \gc{100.49} & \gc{2} & \gc{1382.01} & \gc{2} \\
& & SplitLoRA & 0.19 & 1 & 3433.29 & 1 & 8.73 & 1 & 972.90 & 1 \\

\cmidrule(lr){2-11}

& \multirow{4}{*}{51\%}
& \gc{FedAvg+LoRA}  & \gc{5.73} & \gc{3} & \gc{6777.42} & \gc{2} & \gc{259.77} & \gc{3} & \gc{2367.76} & \gc{3} \\
& & FedProx+LoRA & 38.34 & 4 & 45182.81 & 4 & 1737.24 & 4 & 2371.06 & 4 \\
& & \gc{HeteroLoRA} & \gc{2.68} & \gc{2} & \gc{9755.68} & \gc{3} & \gc{121.50} & \gc{2} & \gc{1389.43} & \gc{2} \\
& & SplitLoRA & 0.28 & 1 & 4959.19 & 1 & 12.62 & 1 & 978.12 & 1 \\

\bottomrule
\end{tabular}
}
\end{table*}

\subsection{Results of Gemma 3-270M}

Tables~\ref{tab:protocol_b_gemma270m_verify}--\ref{tab:protocol_b_gemma270m_choose} report the complete Protocol B results of Gemma 3-270M across all selected tasks and target accuracy levels. 
Overall, SplitLoRA shows the strongest cost-to-target efficiency: it consistently reaches the target accuracy with the lowest wall-clock time, energy consumption, and peak memory across Verify, Choose, and Reason tasks. 
This advantage is especially clear as the target accuracy increases, where FedAvg+LoRA and FedProx+LoRA often require substantially longer training time and much higher energy consumption to reach the same target. 
HeteroLoRA usually ranks between SplitLoRA and the FedAvg/FedProx baselines in wall-clock time and energy consumption, but its communication cost is less stable. 
FedAvg+LoRA and FedProx+LoRA occasionally achieve lower communication volume, particularly at some lower target levels, but this communication advantage does not translate into lower overall system cost due to their longer training time and higher energy usage.

\begin{table*}[h]
\centering
\caption{Complete Protocol B results of Gemma 3-270M on Choose tasks.}
\label{tab:protocol_b_gemma270m_choose}
\renewcommand{\arraystretch}{1.08}
\setlength{\tabcolsep}{3pt}
\scriptsize
\providecommand{\gc}[1]{\cellcolor{gray!10}#1}
\resizebox{\textwidth}{!}{
\begin{tabular}{lclcccccccc}
\toprule
\textbf{Task} 
& \textbf{Target Accuracy} 
& \textbf{Methods} 
& \makecell{\textbf{Wall-clock}\\\textbf{time (hour)}} 
& \textbf{Rank} 
& \makecell{\textbf{Communication}\\\textbf{volume (MB)}} 
& \textbf{Rank} 
& \makecell{\textbf{Energy}\\\textbf{consumption (kJ)}} 
& \textbf{Rank} 
& \makecell{\textbf{Peak}\\\textbf{memory (MB)}} 
& \textbf{Rank} \\
\midrule

\multirow{12}{*}{PIQA}
& \multirow{4}{*}{52\%}
& \gc{FedAvg+LoRA}  & \gc{54.78} & \gc{4} & \gc{64851.75} & \gc{4} & \gc{2536.17} & \gc{4} & \gc{2340.37} & \gc{4} \\
& & FedProx+LoRA & 1.50 & 3 & 1127.86 & 1 & 69.32 & 3 & 2313.91 & 3 \\
& & \gc{HeteroLoRA} & \gc{1.17} & \gc{2} & \gc{6004.91} & \gc{3} & \gc{54.24} & \gc{2} & \gc{1342.21} & \gc{2} \\
& & SplitLoRA & 0.18 & 1 & 3814.76 & 2 & 8.37 & 1 & 1006.46 & 1 \\

\cmidrule(lr){2-11}

& \multirow{4}{*}{53\%}
& \gc{FedAvg+LoRA}  & \gc{56.68} & \gc{4} & \gc{67107.46} & \gc{4} & \gc{2624.33} & \gc{4} & \gc{2368.30} & \gc{4} \\
& & FedProx+LoRA & 1.50 & 2 & 1127.86 & 1 & 69.32 & 2 & 2341.52 & 3 \\
& & \gc{HeteroLoRA} & \gc{3.74} & \gc{3} & \gc{19142.58} & \gc{3} & \gc{173.16} & \gc{3} & \gc{1358.23} & \gc{2} \\
& & SplitLoRA & 0.36 & 1 & 7629.52 & 2 & 16.71 & 1 & 1018.47 & 1 \\

\cmidrule(lr){2-11}

& \multirow{4}{*}{53\%}
& \gc{FedAvg+LoRA}  & \gc{57.64} & \gc{3} & \gc{68235.31} & \gc{2} & \gc{2668.47} & \gc{3} & \gc{2381.01} & \gc{4} \\
& & FedProx+LoRA & 87.52 & 4 & 68799.24 & 3 & 4052.06 & 4 & 2354.09 & 3 \\
& & \gc{HeteroLoRA} & \gc{16.96} & \gc{2} & \gc{89194.59} & \gc{4} & \gc{785.27} & \gc{2} & \gc{1365.52} & \gc{2} \\
& & SplitLoRA & 0.36 & 1 & 7629.52 & 1 & 16.71 & 1 & 1023.94 & 1 \\

\midrule

\multirow{12}{*}{HellaSwag}
& \multirow{4}{*}{25\%}
& \gc{FedAvg+LoRA}  & \gc{17.19} & \gc{3} & \gc{20332.27} & \gc{2} & \gc{794.60} & \gc{3} & \gc{2323.27} & \gc{3} \\
& & FedProx+LoRA & 17.28 & 4 & 20332.27 & 2 & 798.81 & 4 & 2336.17 & 4 \\
& & \gc{HeteroLoRA} & \gc{7.90} & \gc{2} & \gc{28842.89} & \gc{3} & \gc{365.03} & \gc{2} & \gc{1369.23} & \gc{2} \\
& & SplitLoRA & 2.43 & 1 & 13351.67 & 1 & 112.42 & 1 & 940.64 & 1 \\

\cmidrule(lr){2-11}

& \multirow{4}{*}{25\%}
& \gc{FedAvg+LoRA}  & \gc{21.02} & \gc{3} & \gc{24850.55} & \gc{1} & \gc{971.35} & \gc{3} & \gc{2350.99} & \gc{3} \\
& & FedProx+LoRA & 21.13 & 4 & 24850.55 & 1 & 976.53 & 4 & 2364.05 & 4 \\
& & \gc{HeteroLoRA} & \gc{18.12} & \gc{2} & \gc{66168.98} & \gc{3} & \gc{837.60} & \gc{2} & \gc{1385.57} & \gc{2} \\
& & SplitLoRA & 5.21 & 1 & 28610.72 & 2 & 240.86 & 1 & 951.87 & 1 \\

\cmidrule(lr){2-11}

& \multirow{4}{*}{25\%}
& \gc{FedAvg+LoRA}  & \gc{24.84} & \gc{3} & \gc{29368.83} & \gc{1} & \gc{1148.13} & \gc{3} & \gc{2363.61} & \gc{3} \\
& & FedProx+LoRA & 24.93 & 4 & 29368.83 & 1 & 1152.17 & 4 & 2376.74 & 4 \\
& & \gc{HeteroLoRA} & \gc{18.82} & \gc{2} & \gc{68713.95} & \gc{3} & \gc{869.83} & \gc{2} & \gc{1393.01} & \gc{2} \\
& & SplitLoRA & 10.36 & 1 & 56839.96 & 2 & 479.00 & 1 & 956.98 & 1 \\

\midrule

\multirow{12}{*}{SocialIQA}
& \multirow{4}{*}{38\%}
& \gc{FedAvg+LoRA}  & \gc{47.18} & \gc{4} & \gc{55348.95} & \gc{4} & \gc{2091.23} & \gc{4} & \gc{2317.00} & \gc{4} \\
& & FedProx+LoRA & 44.22 & 3 & 51960.23 & 3 & 1959.83 & 3 & 2308.28 & 3 \\
& & \gc{HeteroLoRA} & \gc{11.17} & \gc{2} & \gc{40719.38} & \gc{2} & \gc{494.95} & \gc{2} & \gc{1336.75} & \gc{2} \\
& & SplitLoRA & 1.46 & 1 & 20599.72 & 1 & 64.80 & 1 & 964.00 & 1 \\

\cmidrule(lr){2-11}

& \multirow{4}{*}{39\%}
& \gc{FedAvg+LoRA}  & \gc{69.76} & \gc{4} & \gc{81893.85} & \gc{4} & \gc{3091.84} & \gc{4} & \gc{2342.12} & \gc{4} \\
& & FedProx+LoRA & 53.82 & 3 & 63255.94 & 3 & 2385.60 & 3 & 2333.29 & 3 \\
& & \gc{HeteroLoRA} & \gc{13.38} & \gc{2} & \gc{48778.42} & \gc{2} & \gc{593.26} & \gc{2} & \gc{1351.24} & \gc{2} \\
& & SplitLoRA & 2.03 & 1 & 28610.72 & 1 & 90.03 & 1 & 974.45 & 1 \\

\cmidrule(lr){2-11}

& \multirow{4}{*}{40\%}
& \gc{FedAvg+LoRA}  & \gc{86.13} & \gc{4} & \gc{101096.54} & \gc{4} & \gc{3817.46} & \gc{4} & \gc{2355.84} & \gc{4} \\
& & FedProx+LoRA & 67.74 & 3 & 79634.71 & 2 & 3002.48 & 3 & 2346.96 & 3 \\
& & \gc{HeteroLoRA} & \gc{21.99} & \gc{2} & \gc{80166.27} & \gc{3} & \gc{974.85} & \gc{2} & \gc{1359.15} & \gc{2} \\
& & SplitLoRA & 3.76 & 1 & 53025.19 & 1 & 166.86 & 1 & 980.16 & 1 \\

\bottomrule
\end{tabular}
}
\vspace{-5mm}
\end{table*}

\subsection{Results of Gemma 3-1B}

Tables~\ref{tab:protocol_b_gemma1b_verify}--\ref{tab:protocol_b_gemma1b_choose} report the Protocol B results of Gemma 3-1B. 
Different from Qwen2.5-0.5B and Gemma 3-270M, FedAvg+LoRA, FedProx+LoRA, and HeteroLoRA cannot reach any target accuracy under the current edge-system budget because they trigger out-of-memory failures on most client devices. 
In contrast, SplitLoRA remains executable across all Verify, Choose, and Reason tasks. 
Its peak memory remains small, ranging from about $1313$ MB to $1347$ MB across all target levels, which is far below the memory footprint required by full client-side LoRA fine-tuning of Gemma 3-1B.

This result highlights the deployability advantage of SplitLoRA for larger backbone models. 
Since SplitLoRA places only the first hidden layer on the client side and offloads the remaining model computation to the server, increasing the backbone size does not proportionally increase the client-side memory footprint. 
Therefore, although SplitLoRA may not always achieve the best final quality on smaller models, it becomes the only feasible method when scaling to Gemma 3-1B under realistic edge memory constraints.

\begin{table*}[h]
\centering
\caption{Complete Protocol B results of Gemma 3-1B on Verify tasks. ``-'' indicates that the method is not executable under the system budget due to out-of-memory.}
\label{tab:protocol_b_gemma1b_verify}
\renewcommand{\arraystretch}{1.08}
\setlength{\tabcolsep}{3pt}
\scriptsize
\providecommand{\gc}[1]{\cellcolor{gray!10}#1}
\resizebox{\textwidth}{!}{
\begin{tabular}{lclcccccccc}
\toprule
\textbf{Task} 
& \textbf{Target Accuracy} 
& \textbf{Methods} 
& \makecell{\textbf{Wall-clock}\\\textbf{time (hour)}} 
& \textbf{Rank} 
& \makecell{\textbf{Communication}\\\textbf{volume (MB)}} 
& \textbf{Rank} 
& \makecell{\textbf{Energy}\\\textbf{consumption (kJ)}} 
& \textbf{Rank} 
& \makecell{\textbf{Peak}\\\textbf{memory (MB)}} 
& \textbf{Rank} \\
\midrule

\multirow{12}{*}{BoolQ}
& \multirow{4}{*}{62\%}
& \gc{FedAvg+LoRA}  & \gc{-} & \gc{-} & \gc{-} & \gc{-} & \gc{-} & \gc{-} & \gc{-} & \gc{-} \\
& & FedProx+LoRA & - & - & - & - & - & - & - & - \\
& & \gc{HeteroLoRA} & \gc{-} & \gc{-} & \gc{-} & \gc{-} & \gc{-} & \gc{-} & \gc{-} & \gc{-} \\
& & SplitLoRA & 0.09 & 1 & 4025.52 & 1 & 2.83 & 1 & 1313.15 & 1 \\

\cmidrule(lr){2-11}

& \multirow{4}{*}{63\%}
& \gc{FedAvg+LoRA}  & \gc{-} & \gc{-} & \gc{-} & \gc{-} & \gc{-} & \gc{-} & \gc{-} & \gc{-} \\
& & FedProx+LoRA & - & - & - & - & - & - & - & - \\
& & \gc{HeteroLoRA} & \gc{-} & \gc{-} & \gc{-} & \gc{-} & \gc{-} & \gc{-} & \gc{-} & \gc{-} \\
& & SplitLoRA & 1.42 & 1 & 60382.81 & 1 & 42.43 & 1 & 1329.96 & 1 \\

\cmidrule(lr){2-11}

& \multirow{4}{*}{64\%}
& \gc{FedAvg+LoRA}  & \gc{-} & \gc{-} & \gc{-} & \gc{-} & \gc{-} & \gc{-} & \gc{-} & \gc{-} \\
& & FedProx+LoRA & - & - & - & - & - & - & - & - \\
& & \gc{HeteroLoRA} & \gc{-} & \gc{-} & \gc{-} & \gc{-} & \gc{-} & \gc{-} & \gc{-} & \gc{-} \\
& & SplitLoRA & 1.49 & 1 & 63401.95 & 1 & 44.58 & 1 & 1338.56 & 1 \\

\midrule

\multirow{12}{*}{QNLI}
& \multirow{4}{*}{57\%}
& \gc{FedAvg+LoRA}  & \gc{-} & \gc{-} & \gc{-} & \gc{-} & \gc{-} & \gc{-} & \gc{-} & \gc{-} \\
& & FedProx+LoRA & - & - & - & - & - & - & - & - \\
& & \gc{HeteroLoRA} & \gc{-} & \gc{-} & \gc{-} & \gc{-} & \gc{-} & \gc{-} & \gc{-} & \gc{-} \\
& & SplitLoRA & 0.15 & 1 & 6038.28 & 1 & 4.86 & 1 & 1323.72 & 1 \\

\cmidrule(lr){2-11}

& \multirow{4}{*}{60\%}
& \gc{FedAvg+LoRA}  & \gc{-} & \gc{-} & \gc{-} & \gc{-} & \gc{-} & \gc{-} & \gc{-} & \gc{-} \\
& & FedProx+LoRA & - & - & - & - & - & - & - & - \\
& & \gc{HeteroLoRA} & \gc{-} & \gc{-} & \gc{-} & \gc{-} & \gc{-} & \gc{-} & \gc{-} & \gc{-} \\
& & SplitLoRA & 0.18 & 1 & 7044.66 & 1 & 5.67 & 1 & 1339.52 & 1 \\

\cmidrule(lr){2-11}

& \multirow{4}{*}{63\%}
& \gc{FedAvg+LoRA}  & \gc{-} & \gc{-} & \gc{-} & \gc{-} & \gc{-} & \gc{-} & \gc{-} & \gc{-} \\
& & FedProx+LoRA & - & - & - & - & - & - & - & - \\
& & \gc{HeteroLoRA} & \gc{-} & \gc{-} & \gc{-} & \gc{-} & \gc{-} & \gc{-} & \gc{-} & \gc{-} \\
& & SplitLoRA & 0.20 & 1 & 8051.04 & 1 & 6.48 & 1 & 1346.70 & 1 \\

\bottomrule
\end{tabular}
}
\end{table*}

\begin{table*}[h]
\centering
\caption{Complete Protocol B results of Gemma 3-1B on Reason tasks. ``-'' indicates that the method is not executable under the system budget due to out-of-memory.}
\label{tab:protocol_b_gemma1b_reason}
\renewcommand{\arraystretch}{1.08}
\setlength{\tabcolsep}{3pt}
\scriptsize
\providecommand{\gc}[1]{\cellcolor{gray!10}#1}
\resizebox{\textwidth}{!}{
\begin{tabular}{lclcccccccc}
\toprule
\textbf{Task} 
& \textbf{Target Accuracy} 
& \textbf{Methods} 
& \makecell{\textbf{Wall-clock}\\\textbf{time (hour)}} 
& \textbf{Rank} 
& \makecell{\textbf{Communication}\\\textbf{volume (MB)}} 
& \textbf{Rank} 
& \makecell{\textbf{Energy}\\\textbf{consumption (kJ)}} 
& \textbf{Rank} 
& \makecell{\textbf{Peak}\\\textbf{memory (MB)}} 
& \textbf{Rank} \\
\midrule

\multirow{12}{*}{ARC-E}
& \multirow{4}{*}{27\%}
& \gc{FedAvg+LoRA}  & \gc{-} & \gc{-} & \gc{-} & \gc{-} & \gc{-} & \gc{-} & \gc{-} & \gc{-} \\
& & FedProx+LoRA & - & - & - & - & - & - & - & - \\
& & \gc{HeteroLoRA} & \gc{-} & \gc{-} & \gc{-} & \gc{-} & \gc{-} & \gc{-} & \gc{-} & \gc{-} \\
& & SplitLoRA & 0.15 & 1 & 6541.47 & 1 & 4.52 & 1 & 1316.12 & 1 \\

\cmidrule(lr){2-11}

& \multirow{4}{*}{27\%}
& \gc{FedAvg+LoRA}  & \gc{-} & \gc{-} & \gc{-} & \gc{-} & \gc{-} & \gc{-} & \gc{-} & \gc{-} \\
& & FedProx+LoRA & - & - & - & - & - & - & - & - \\
& & \gc{HeteroLoRA} & \gc{-} & \gc{-} & \gc{-} & \gc{-} & \gc{-} & \gc{-} & \gc{-} & \gc{-} \\
& & SplitLoRA & 0.15 & 1 & 6541.47 & 1 & 4.52 & 1 & 1332.13 & 1 \\

\cmidrule(lr){2-11}

& \multirow{4}{*}{28\%}
& \gc{FedAvg+LoRA}  & \gc{-} & \gc{-} & \gc{-} & \gc{-} & \gc{-} & \gc{-} & \gc{-} & \gc{-} \\
& & FedProx+LoRA & - & - & - & - & - & - & - & - \\
& & \gc{HeteroLoRA} & \gc{-} & \gc{-} & \gc{-} & \gc{-} & \gc{-} & \gc{-} & \gc{-} & \gc{-} \\
& & SplitLoRA & 0.19 & 1 & 8051.04 & 1 & 5.56 & 1 & 1337.16 & 1 \\

\midrule

\multirow{12}{*}{WinoGrande}
& \multirow{4}{*}{51\%}
& \gc{FedAvg+LoRA}  & \gc{-} & \gc{-} & \gc{-} & \gc{-} & \gc{-} & \gc{-} & \gc{-} & \gc{-} \\
& & FedProx+LoRA & - & - & - & - & - & - & - & - \\
& & \gc{HeteroLoRA} & \gc{-} & \gc{-} & \gc{-} & \gc{-} & \gc{-} & \gc{-} & \gc{-} & \gc{-} \\
& & SplitLoRA & 1.58 & 1 & 68433.85 & 1 & 47.09 & 1 & 1318.06 & 1 \\

\cmidrule(lr){2-11}

& \multirow{4}{*}{51\%}
& \gc{FedAvg+LoRA}  & \gc{-} & \gc{-} & \gc{-} & \gc{-} & \gc{-} & \gc{-} & \gc{-} & \gc{-} \\
& & FedProx+LoRA & - & - & - & - & - & - & - & - \\
& & \gc{HeteroLoRA} & \gc{-} & \gc{-} & \gc{-} & \gc{-} & \gc{-} & \gc{-} & \gc{-} & \gc{-} \\
& & SplitLoRA & 1.68 & 1 & 72962.56 & 1 & 50.21 & 1 & 1333.79 & 1 \\

\cmidrule(lr){2-11}

& \multirow{4}{*}{51\%}
& \gc{FedAvg+LoRA}  & \gc{-} & \gc{-} & \gc{-} & \gc{-} & \gc{-} & \gc{-} & \gc{-} & \gc{-} \\
& & FedProx+LoRA & - & - & - & - & - & - & - & - \\
& & \gc{HeteroLoRA} & \gc{-} & \gc{-} & \gc{-} & \gc{-} & \gc{-} & \gc{-} & \gc{-} & \gc{-} \\
& & SplitLoRA & 1.73 & 1 & 74975.32 & 1 & 51.61 & 1 & 1340.95 & 1 \\

\bottomrule
\end{tabular}
}
\end{table*}

\clearpage

\begin{table*}[h]
\centering
\caption{Complete Protocol B results of Gemma 3-1B on Choose tasks. ``-'' indicates that the method is not executable under the system budget due to out-of-memory.}
\label{tab:protocol_b_gemma1b_choose}
\renewcommand{\arraystretch}{1.08}
\setlength{\tabcolsep}{3pt}
\scriptsize
\providecommand{\gc}[1]{\cellcolor{gray!10}#1}
\resizebox{\textwidth}{!}{
\begin{tabular}{lclcccccccc}
\toprule
\textbf{Task} 
& \textbf{Target Accuracy} 
& \textbf{Methods} 
& \makecell{\textbf{Wall-clock}\\\textbf{time (hour)}} 
& \textbf{Rank} 
& \makecell{\textbf{Communication}\\\textbf{volume (MB)}} 
& \textbf{Rank} 
& \makecell{\textbf{Energy}\\\textbf{consumption (kJ)}} 
& \textbf{Rank} 
& \makecell{\textbf{Peak}\\\textbf{memory (MB)}} 
& \textbf{Rank} \\
\midrule

\multirow{12}{*}{PIQA}
& \multirow{4}{*}{53\%}
& \gc{FedAvg+LoRA}  & \gc{-} & \gc{-} & \gc{-} & \gc{-} & \gc{-} & \gc{-} & \gc{-} & \gc{-} \\
& & FedProx+LoRA & - & - & - & - & - & - & - & - \\
& & \gc{HeteroLoRA} & \gc{-} & \gc{-} & \gc{-} & \gc{-} & \gc{-} & \gc{-} & \gc{-} & \gc{-} \\
& & SplitLoRA & 0.87 & 1 & 37236.07 & 1 & 25.74 & 1 & 1316.86 & 1 \\

\cmidrule(lr){2-11}

& \multirow{4}{*}{53\%}
& \gc{FedAvg+LoRA}  & \gc{-} & \gc{-} & \gc{-} & \gc{-} & \gc{-} & \gc{-} & \gc{-} & \gc{-} \\
& & FedProx+LoRA & - & - & - & - & - & - & - & - \\
& & \gc{HeteroLoRA} & \gc{-} & \gc{-} & \gc{-} & \gc{-} & \gc{-} & \gc{-} & \gc{-} & \gc{-} \\
& & SplitLoRA & 0.87 & 1 & 37236.07 & 1 & 25.74 & 1 & 1332.58 & 1 \\

\cmidrule(lr){2-11}

& \multirow{4}{*}{54\%}
& \gc{FedAvg+LoRA}  & \gc{-} & \gc{-} & \gc{-} & \gc{-} & \gc{-} & \gc{-} & \gc{-} & \gc{-} \\
& & FedProx+LoRA & - & - & - & - & - & - & - & - \\
& & \gc{HeteroLoRA} & \gc{-} & \gc{-} & \gc{-} & \gc{-} & \gc{-} & \gc{-} & \gc{-} & \gc{-} \\
& & SplitLoRA & 1.15 & 1 & 49312.63 & 1 & 34.10 & 1 & 1339.73 & 1 \\

\midrule

\multirow{12}{*}{HellaSwag}
& \multirow{4}{*}{25\%}
& \gc{FedAvg+LoRA}  & \gc{-} & \gc{-} & \gc{-} & \gc{-} & \gc{-} & \gc{-} & \gc{-} & \gc{-} \\
& & FedProx+LoRA & - & - & - & - & - & - & - & - \\
& & \gc{HeteroLoRA} & \gc{-} & \gc{-} & \gc{-} & \gc{-} & \gc{-} & \gc{-} & \gc{-} & \gc{-} \\
& & SplitLoRA & 0.13 & 1 & 5535.09 & 1 & 4.00 & 1 & 1319.53 & 1 \\

\cmidrule(lr){2-11}

& \multirow{4}{*}{25\%}
& \gc{FedAvg+LoRA}  & \gc{-} & \gc{-} & \gc{-} & \gc{-} & \gc{-} & \gc{-} & \gc{-} & \gc{-} \\
& & FedProx+LoRA & - & - & - & - & - & - & - & - \\
& & \gc{HeteroLoRA} & \gc{-} & \gc{-} & \gc{-} & \gc{-} & \gc{-} & \gc{-} & \gc{-} & \gc{-} \\
& & SplitLoRA & 0.13 & 1 & 5535.09 & 1 & 4.00 & 1 & 1335.28 & 1 \\

\cmidrule(lr){2-11}

& \multirow{4}{*}{25\%}
& \gc{FedAvg+LoRA}  & \gc{-} & \gc{-} & \gc{-} & \gc{-} & \gc{-} & \gc{-} & \gc{-} & \gc{-} \\
& & FedProx+LoRA & - & - & - & - & - & - & - & - \\
& & \gc{HeteroLoRA} & \gc{-} & \gc{-} & \gc{-} & \gc{-} & \gc{-} & \gc{-} & \gc{-} & \gc{-} \\
& & SplitLoRA & 1.31 & 1 & 55854.10 & 1 & 40.27 & 1 & 1342.45 & 1 \\

\midrule

\multirow{12}{*}{SocialIQA}
& \multirow{4}{*}{40\%}
& \gc{FedAvg+LoRA}  & \gc{-} & \gc{-} & \gc{-} & \gc{-} & \gc{-} & \gc{-} & \gc{-} & \gc{-} \\
& & FedProx+LoRA & - & - & - & - & - & - & - & - \\
& & \gc{HeteroLoRA} & \gc{-} & \gc{-} & \gc{-} & \gc{-} & \gc{-} & \gc{-} & \gc{-} & \gc{-} \\
& & SplitLoRA & 0.84 & 1 & 34216.93 & 1 & 25.27 & 1 & 1321.08 & 1 \\

\cmidrule(lr){2-11}

& \multirow{4}{*}{42\%}
& \gc{FedAvg+LoRA}  & \gc{-} & \gc{-} & \gc{-} & \gc{-} & \gc{-} & \gc{-} & \gc{-} & \gc{-} \\
& & FedProx+LoRA & - & - & - & - & - & - & - & - \\
& & \gc{HeteroLoRA} & \gc{-} & \gc{-} & \gc{-} & \gc{-} & \gc{-} & \gc{-} & \gc{-} & \gc{-} \\
& & SplitLoRA & 0.98 & 1 & 40255.21 & 1 & 29.70 & 1 & 1335.40 & 1 \\

\cmidrule(lr){2-11}

& \multirow{4}{*}{45\%}
& \gc{FedAvg+LoRA}  & \gc{-} & \gc{-} & \gc{-} & \gc{-} & \gc{-} & \gc{-} & \gc{-} & \gc{-} \\
& & FedProx+LoRA & - & - & - & - & - & - & - & - \\
& & \gc{HeteroLoRA} & \gc{-} & \gc{-} & \gc{-} & \gc{-} & \gc{-} & \gc{-} & \gc{-} & \gc{-} \\
& & SplitLoRA & 1.36 & 1 & 55854.10 & 1 & 41.19 & 1 & 1343.23 & 1 \\

\bottomrule
\end{tabular}
}

\end{table*}

\begin{figure*}[h]
\centering
\includegraphics[width=\textwidth]{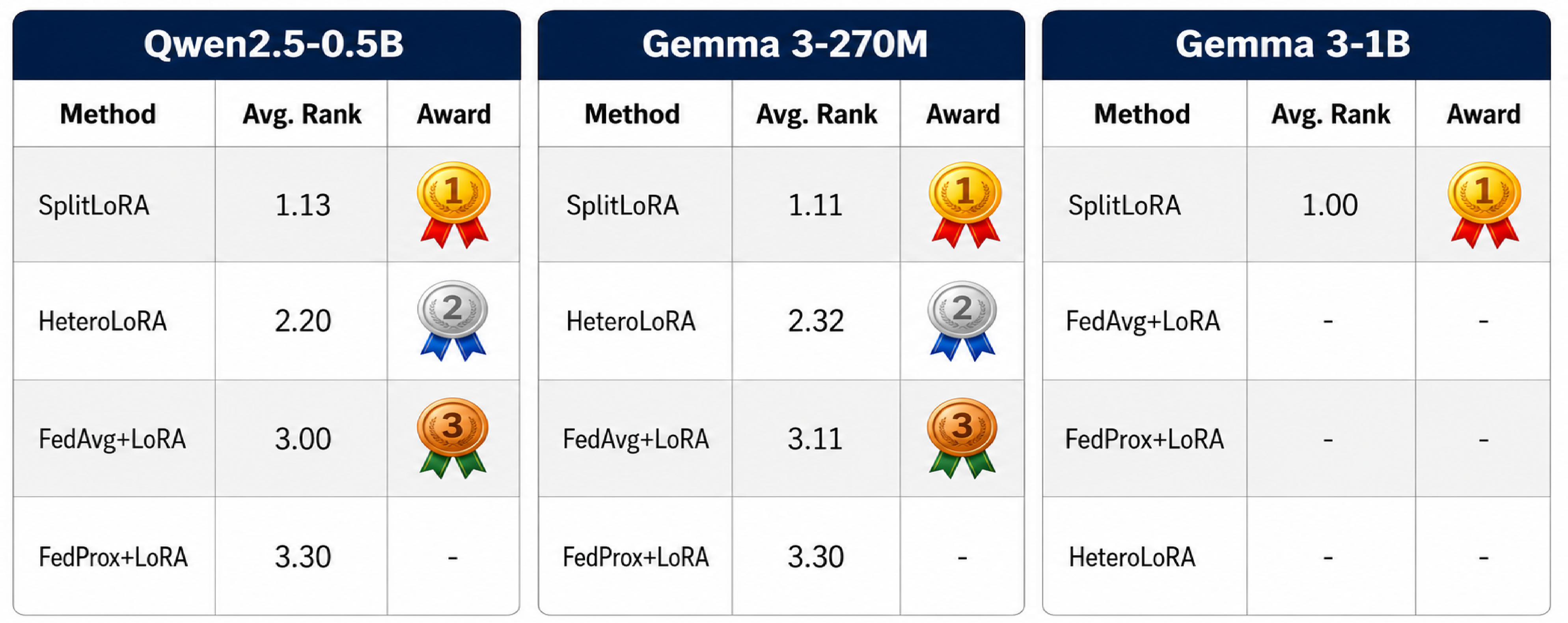}
\caption{Overall ranking of methods under protocol B. }
\label{fig:protocolBranking}
\vspace{-5mm}
\end{figure*}

\subsection{Overall ranking of methods under protocol B}

The overall ranking under Protocol B shows that SplitLoRA is the most efficient and deployable method across all three model scales. 
This result should be interpreted together with the quality results in Protocol A: SplitLoRA is not always the method with the highest final accuracy on smaller models, where FedProx+LoRA often achieves the best accuracy due to its proximal regularization on local LoRA updates. 
However, Protocol B focuses on the system cost required to reach target accuracy levels, and under this cost-to-target criterion SplitLoRA consistently achieves the best average rank for Qwen2.5-0.5B and Gemma 3-270M. 
It reaches target accuracies with substantially lower wall-clock time, energy consumption, and peak memory, showing a stronger efficiency--quality trade-off under realistic edge-system constraints.

HeteroLoRA ranks second overall because its heterogeneous LoRA design adapts to different device capabilities. 
FedAvg+LoRA ranks third, while FedProx+LoRA ranks last under Protocol B. 
This indicates that better final accuracy does not necessarily imply better cost-to-target efficiency: FedProx+LoRA can achieve strong accuracy, but it often requires longer training time and higher energy consumption to reach the same target. 
For Gemma 3-1B, only SplitLoRA is feasible, while the other methods are infeasible due to out-of-memory errors. 
This further highlights the key advantage of SplitLoRA: even when it is not the most accurate method on smaller models, it provides the best efficiency and deployability when realistic edge constraints and larger backbone models are considered.

Therefore, Figure~\ref{fig:protocolBranking} highlights a key Protocol B insight: \textit{Methods with higher final accuracy do not necessarily provide better system efficiency.
Their relative efficiency can also change across target accuracy levels and system metrics.}

\section{Complete benchmark results of protocol C}
\label{AppendixC}

This section provides the complete benchmark results for Protocol C, i.e., \textit{Robustness}. 
We use the same basic experimental setup as Protocols A and B, including the same client pool, device types, data partitioning strategy, communication-round configuration, backbone models, datasets, and federated fine-tuning methods. 
The key difference is that Protocol C introduces system perturbations to evaluate how robust each method is under realistic edge deployment conditions.

Here, we show the results of evaluating robustness under dynamic communication fluctuation. 
To simulate unstable wireless connectivity in real edge deployments, we dynamically change the Wi-Fi bandwidth every $1/3$ hour, sequentially setting it to the full bandwidth, $1/2$ of the bandwidth, and $1/4$ of the bandwidth. 
For each method, we report both model quality and system metrics under the fluctuating communication setting.

\subsection{Results of Qwen2.5-0.5B}

Table~\ref{tab:protocol_c_qwen25_comm_fluctuation} reports the robustness results of Qwen2.5-0.5B under dynamic communication fluctuation. 
Across all seven tasks, the testing accuracy remains unchanged for all methods, indicating that bandwidth fluctuation mainly affects system cost rather than final model quality. 
The communication volume and peak memory also remain unchanged, because the transmitted payload size and client-side model footprint are not altered by bandwidth variation. 
However, wall-clock time and energy consumption increase significantly under fluctuating bandwidth. 
Among the four methods, SplitLoRA consistently shows the smallest increase in wall-clock time and energy consumption, achieving the best robustness rank across all tasks. 
In contrast, FedAvg+LoRA, FedProx+LoRA, and HeteroLoRA suffer much larger increases in time and energy, because their training process requires longer client-side execution and is more exposed to bandwidth degradation. 

\begin{table*}[h]
\centering
\caption{Complete Protocol C results of Qwen2.5-0.5B under communication fluctuation. Values in parentheses denote changes relative to the non-fluctuation setting. Ranks are computed by the absolute value of the change, where smaller change indicates better robustness}
\label{tab:protocol_c_qwen25_comm_fluctuation}
\renewcommand{\arraystretch}{1.08}
\setlength{\tabcolsep}{3pt}
\scriptsize
\providecommand{\gc}[1]{\cellcolor{gray!10}#1}
\resizebox{\textwidth}{!}{
\begin{tabular}{lllcccccccccc}
\toprule
\textbf{Task Type} 
& \textbf{Task} 
& \textbf{Methods} 
& \textbf{Testing Accuracy} 
& \textbf{Rank} 
& \makecell{\textbf{Wall-clock}\\\textbf{time (h)}} 
& \textbf{Rank} 
& \makecell{\textbf{Communication}\\\textbf{volume (MB)}} 
& \textbf{Rank} 
& \makecell{\textbf{Energy}\\\textbf{consumption (kJ)}} 
& \textbf{Rank} 
& \makecell{\textbf{Peak}\\\textbf{memory (MB)}} 
& \textbf{Rank} \\
\midrule

\multirow{8}{*}{Verify}
& \multirow{4}{*}{BoolQ}
& \gc{SplitLoRA}    
& \gc{78.69\% (+0.00\%)} & \gc{1} 
& \gc{6.14 (+2.50)} & \gc{1} 
& \gc{99824.33 (+0.00)} & \gc{1} 
& \gc{277.44 (+112.97)} & \gc{1} 
& \gc{1151.97 (+0.00)} & \gc{1} \\
& & HeteroLoRA    
& 79.02\% (+0.00\%) & 1 
& 141.91 (+59.12) & 2 
& 62801.48 (+0.00) & 1 
& 6412.58 (+2671.55) & 2 
& 2519.11 (+0.00) & 1 \\
& & \gc{FedAvg+LoRA}   
& \gc{79.39\% (+0.00\%)} & \gc{1} 
& \gc{168.13 (+70.00)} & \gc{3} 
& \gc{97709.53 (+0.00)} & \gc{1} 
& \gc{7597.85 (+3163.24)} & \gc{3} 
& \gc{3481.10 (+0.00)} & \gc{1} \\
& & FedProx+LoRA  
& 79.33\% (+0.00\%) & 1 
& 169.31 (+70.42) & 4 
& 98537.58 (+0.00) & 1 
& 7651.04 (+3182.07) & 4 
& 3497.98 (+0.00) & 1 \\

\cmidrule(lr){2-13}

& \multirow{4}{*}{QNLI}
& \gc{SplitLoRA}    
& \gc{65.20\% (+0.00\%)} & \gc{1} 
& \gc{6.28 (+2.50)} & \gc{1} 
& \gc{82208.27 (+0.00)} & \gc{1} 
& \gc{292.62 (+116.43)} & \gc{1} 
& \gc{1159.46 (+0.00)} & \gc{1} \\
& & HeteroLoRA    
& 64.29\% (+0.00\%) & 1 
& 275.53 (+114.73) & 3 
& 121250.39 (+0.00) & 1 
& 12832.37 (+5343.39) & 3 
& 2525.05 (+0.00) & 1 \\
& & \gc{FedAvg+LoRA}   
& \gc{64.84\% (+0.00\%)} & \gc{1} 
& \gc{225.03 (+93.75)} & \gc{2} 
& \gc{130831.41 (+0.00)} & \gc{1} 
& \gc{10480.39 (+4366.28)} & \gc{2} 
& \gc{3394.81 (+0.00)} & \gc{1} \\
& & FedProx+LoRA  
& 64.96\% (+0.00\%) & 1 
& 279.66 (+116.50) & 4 
& 163125.23 (+0.00) & 1 
& 13024.89 (+5425.68) & 4 
& 3517.76 (+0.00) & 1 \\

\midrule

\multirow{12}{*}{Choose}
& \multirow{4}{*}{PIQA}
& \gc{SplitLoRA}    
& \gc{64.47\% (+0.00\%)} & \gc{1} 
& \gc{1.25 (+0.42)} & \gc{1} 
& \gc{21886.62 (+0.00)} & \gc{1} 
& \gc{57.93 (+19.33)} & \gc{1} 
& \gc{1144.44 (+0.00)} & \gc{1} \\
& & HeteroLoRA    
& 66.05\% (+0.00\%) & 1 
& 106.81 (+44.49) & 4 
& 47256.56 (+0.00) & 1 
& 4955.24 (+2063.96) & 4 
& 2545.88 (+0.00) & 1 \\
& & \gc{FedAvg+LoRA}   
& \gc{65.29\% (+0.00\%)} & \gc{1} 
& \gc{93.03 (+38.75)} & \gc{3} 
& \gc{53823.05 (+0.00)} & \gc{1} 
& \gc{4316.03 (+1797.67)} & \gc{3} 
& \gc{3452.15 (+0.00)} & \gc{1} \\
& & FedProx+LoRA  
& 65.29\% (+0.00\%) & 1 
& 61.70 (+25.68) & 2 
& 35606.02 (+0.00) & 1 
& 2862.36 (+1191.49) & 2 
& 3418.02 (+0.00) & 1 \\

\cmidrule(lr){2-13}

& \multirow{4}{*}{HellaSwag}
& \gc{SplitLoRA}    
& \gc{33.67\% (+0.00\%)} & \gc{1} 
& \gc{8.03 (+3.33)} & \gc{1} 
& \gc{106763.99 (+0.00)} & \gc{1} 
& \gc{390.22 (+162.00)} & \gc{1} 
& \gc{1156.66 (+0.00)} & \gc{1} \\
& & HeteroLoRA    
& 33.18\% (+0.00\%) & 1 
& 274.20 (+114.17) & 3 
& 121250.39 (+0.00) & 1 
& 13326.07 (+5548.47) & 3 
& 2602.45 (+0.00) & 1 \\
& & \gc{FedAvg+LoRA}   
& \gc{33.84\% (+0.00\%)} & \gc{1} 
& \gc{271.04 (+112.92)} & \gc{2} 
& \gc{157328.91 (+0.00)} & \gc{1} 
& \gc{13172.27 (+5487.72)} & \gc{2} 
& \gc{3387.78 (+0.00)} & \gc{1} \\
& & FedProx+LoRA  
& 33.84\% (+0.00\%) & 1 
& 271.23 (+112.92) & 2 
& 157328.91 (+0.00) & 1 
& 13181.80 (+5487.72) & 2 
& 3559.35 (+0.00) & 1 \\

\cmidrule(lr){2-13}

& \multirow{4}{*}{SocialIQA}
& \gc{SplitLoRA}    
& \gc{67.45\% (+0.00\%)} & \gc{1} 
& \gc{5.17 (+2.08)} & \gc{1} 
& \gc{71531.87 (+0.00)} & \gc{1} 
& \gc{253.07 (+101.89)} & \gc{1} 
& \gc{1152.10 (+0.00)} & \gc{1} \\
& & HeteroLoRA    
& 67.45\% (+0.00\%) & 1 
& 277.94 (+115.80) & 4 
& 123737.58 (+0.00) & 1 
& 13593.57 (+5663.76) & 4 
& 2571.94 (+0.00) & 1 \\
& & \gc{FedAvg+LoRA}   
& \gc{68.42\% (+0.00\%)} & \gc{1} 
& \gc{274.18 (+114.17)} & \gc{3} 
& \gc{158985.00 (+0.00)} & \gc{1} 
& \gc{13409.35 (+5583.64)} & \gc{3} 
& \gc{3431.24 (+0.00)} & \gc{1} \\
& & FedProx+LoRA  
& 68.42\% (+0.00\%) & 1 
& 272.12 (+113.33) & 2 
& 158985.00 (+0.00) & 1 
& 13308.80 (+5542.88) & 2 
& 3474.71 (+0.00) & 1 \\

\midrule

\multirow{8}{*}{Reason}
& \multirow{4}{*}{ARC-E}
& \gc{SplitLoRA}    
& \gc{77.37\% (+0.00\%)} & \gc{1} 
& \gc{3.72 (+1.52)} & \gc{1} 
& \gc{49111.44 (+0.00)} & \gc{1} 
& \gc{179.89 (+73.81)} & \gc{1} 
& \gc{1147.42 (+0.00)} & \gc{1} \\
& & HeteroLoRA    
& 77.54\% (+0.00\%) & 1 
& 198.11 (+82.50) & 4 
& 87673.36 (+0.00) & 1 
& 9592.96 (+3994.92) & 4 
& 2554.31 (+0.00) & 1 \\
& & \gc{FedAvg+LoRA}   
& \gc{79.47\% (+0.00\%)} & \gc{1} 
& \gc{156.21 (+65.00)} & \gc{3} 
& \gc{90257.11 (+0.00)} & \gc{1} 
& \gc{7564.40 (+3147.51)} & \gc{3} 
& \gc{3422.95 (+0.00)} & \gc{1} \\
& & FedProx+LoRA  
& 79.47\% (+0.00\%) & 1 
& 155.23 (+64.58) & 2 
& 90257.11 (+0.00) & 1 
& 7516.90 (+3127.34) & 2 
& 3460.78 (+0.00) & 1 \\

\cmidrule(lr){2-13}

& \multirow{4}{*}{WinoGrande}
& \gc{SplitLoRA}    
& \gc{62.43\% (+0.00\%)} & \gc{1} 
& \gc{2.21 (+0.83)} & \gc{1} 
& \gc{40036.50 (+0.00)} & \gc{1} 
& \gc{111.99 (+42.15)} & \gc{1} 
& \gc{1139.52 (+0.00)} & \gc{1} \\
& & HeteroLoRA    
& 61.17\% (+0.00\%) & 1 
& 250.28 (+104.17) & 3 
& 110679.84 (+0.00) & 1 
& 12660.52 (+5269.37) & 3 
& 2535.09 (+0.00) & 1 \\
& & \gc{FedAvg+LoRA}   
& \gc{63.46\% (+0.00\%)} & \gc{1} 
& \gc{224.22 (+93.33)} & \gc{2} 
& \gc{130831.41 (+0.00)} & \gc{1} 
& \gc{11342.17 (+4721.35)} & \gc{2} 
& \gc{3392.58 (+0.00)} & \gc{1} \\
& & FedProx+LoRA  
& 63.61\% (+0.00\%) & 1 
& 266.85 (+111.18) & 4 
& 155672.81 (+0.00) & 1 
& 13498.93 (+5623.93) & 4 
& 3457.41 (+0.00) & 1 \\

\bottomrule
\end{tabular}
}
\end{table*}

\subsection{Results of Gemma 3-270M}

Table~\ref{tab:protocol_c_gemma270m_comm_fluctuation} shows a similar trend for Gemma 3-270M. 
The final testing accuracy is stable under communication fluctuation, and both communication volume and peak memory remain nearly unchanged across methods. 
This confirms that the perturbation mainly changes the effective communication delay rather than the amount of transmitted data or the memory requirement. 
Nevertheless, the impact on wall-clock time and energy consumption is method-dependent. 
SplitLoRA again achieves the smallest perturbation across almost all tasks, showing strong robustness to unstable bandwidth.

\begin{table*}[h]
\centering
\caption{Complete Protocol C results of Gemma 3-270M under communication fluctuation. Values in parentheses denote changes relative to the non-fluctuation setting. Ranks are computed by the absolute value of the change, where smaller change indicates better robustness}
\label{tab:protocol_c_gemma270m_comm_fluctuation}
\renewcommand{\arraystretch}{1.08}
\setlength{\tabcolsep}{3pt}
\scriptsize
\providecommand{\gc}[1]{\cellcolor{gray!10}#1}
\resizebox{\textwidth}{!}{
\begin{tabular}{lllcccccccccc}
\toprule
\textbf{Task Type} 
& \textbf{Task} 
& \textbf{Methods} 
& \textbf{Testing Accuracy} 
& \textbf{Rank} 
& \makecell{\textbf{Wall-clock}\\\textbf{time (h)}} 
& \textbf{Rank} 
& \makecell{\textbf{Communication}\\\textbf{volume (MB)}} 
& \textbf{Rank} 
& \makecell{\textbf{Energy}\\\textbf{consumption (kJ)}} 
& \textbf{Rank} 
& \makecell{\textbf{Peak}\\\textbf{memory (MB)}} 
& \textbf{Rank} \\
\midrule

\multirow{8}{*}{Verify}
& \multirow{4}{*}{BoolQ}
& \gc{SplitLoRA}    
& \gc{62.72\% (+0.00\%)} & \gc{1} 
& \gc{4.19 (+1.67)} & \gc{1} 
& \gc{61407.17 (+0.00)} & \gc{1} 
& \gc{217.62 (+86.62)} & \gc{1} 
& \gc{1010.90 (+0.00)} & \gc{1} \\
& & HeteroLoRA    
& 62.20\% (+0.00\%) & 1 
& 112.15 (+46.67) & 2 
& 59382.42 (+0.00) & 1 
& 5828.63 (+2425.33) & 2 
& 1414.19 (+0.00) & 1 \\
& & \gc{FedAvg+LoRA}   
& \gc{62.94\% (+0.00\%)} & \gc{1} 
& \gc{372.85 (+155.34)} & \gc{4} 
& \gc{101096.54 (+0.00)} & \gc{1} 
& \gc{19377.49 (+8073.30)} & \gc{4} 
& \gc{1350.13 (+0.00)} & \gc{1} \\
& & FedProx+LoRA  
& 64.53\% (+0.00\%) & 1 
& 128.15 (+53.33) & 3 
& 88106.48 (+0.00) & 1 
& 6660.06 (+2771.80) & 3 
& 2343.31 (+0.00) & 1 \\

\cmidrule(lr){2-13}

& \multirow{4}{*}{QNLI}
& \gc{SplitLoRA}    
& \gc{71.46\% (+0.00\%)} & \gc{1} 
& \gc{3.34 (+1.26)} & \gc{1} 
& \gc{40817.95 (+0.00)} & \gc{1} 
& \gc{174.60 (+65.66)} & \gc{1} 
& \gc{1024.00 (+0.00)} & \gc{1} \\
& & HeteroLoRA    
& 63.87\% (+0.00\%) & 1 
& 151.79 (+63.23) & 3 
& 80590.43 (+0.00) & 1 
& 7928.40 (+3302.58) & 3 
& 1459.04 (+0.00) & 1 \\
& & \gc{FedAvg+LoRA}   
& \gc{68.26\% (+0.00\%)} & \gc{1} 
& \gc{126.44 (+52.58)} & \gc{2} 
& \gc{87541.70 (+0.00)} & \gc{1} 
& \gc{6604.48 (+2746.51)} & \gc{2} 
& \gc{2378.84 (+0.00)} & \gc{1} \\
& & FedProx+LoRA  
& 66.98\% (+0.00\%) & 1 
& 157.72 (+65.69) & 4 
& 109003.54 (+0.00) & 1 
& 8238.39 (+3431.46) & 4 
& 2396.10 (+0.00) & 1 \\

\midrule

\multirow{12}{*}{Choose}
& \multirow{4}{*}{PIQA}
& \gc{SplitLoRA}    
& \gc{53.32\% (+0.00\%)} & \gc{1} 
& \gc{0.44 (+0.08)} & \gc{1} 
& \gc{7629.52 (+0.00)} & \gc{1} 
& \gc{20.54 (+3.83)} & \gc{1} 
& \gc{1012.20 (+0.00)} & \gc{1} \\
& & HeteroLoRA    
& 52.18\% (+0.00\%) & 1 
& 123.32 (+51.25) & 3 
& 95394.95 (+0.00) & 1 
& 5709.59 (+2372.77) & 3 
& 1349.86 (+0.00) & 1 \\
& & \gc{FedAvg+LoRA}   
& \gc{53.43\% (+0.00\%)} & \gc{1} 
& \gc{98.77 (+41.14)} & \gc{2} 
& \gc{68235.31 (+0.00)} & \gc{1} 
& \gc{4573.02 (+1904.55)} & \gc{2} 
& \gc{2353.70 (+0.00)} & \gc{1} \\
& & FedProx+LoRA  
& 51.69\% (+0.00\%) & 1 
& 150.02 (+62.50) & 4 
& 68799.24 (+0.00) & 1 
& 6945.68 (+2893.62) & 4 
& 2327.10 (+0.00) & 1 \\

\cmidrule(lr){2-13}

& \multirow{4}{*}{HellaSwag}
& \gc{SplitLoRA}    
& \gc{24.88\% (+0.00\%)} & \gc{1} 
& \gc{22.10 (+9.17)} & \gc{1} 
& \gc{70954.58 (+0.00)} & \gc{1} 
& \gc{1021.62 (+423.66)} & \gc{1} 
& \gc{946.00 (+0.00)} & \gc{1} \\
& & HeteroLoRA    
& 24.93\% (+0.00\%) & 1 
& 157.70 (+65.68) & 4 
& 83983.71 (+0.00) & 1 
& 7288.49 (+3035.72) & 4 
& 1377.03 (+0.00) & 1 \\
& & \gc{FedAvg+LoRA}   
& \gc{24.84\% (+0.00\%)} & \gc{1} 
& \gc{61.26 (+25.42)} & \gc{2} 
& \gc{42358.89 (+0.00)} & \gc{1} 
& \gc{2831.30 (+1174.69)} & \gc{2} 
& \gc{2336.51 (+0.00)} & \gc{1} \\
& & FedProx+LoRA  
& 24.84\% (+0.00\%) & 1 
& 61.34 (+25.42) & 3 
& 42358.89 (+0.00) & 1 
& 2835.02 (+1174.96) & 3 
& 2349.48 (+0.00) & 1 \\

\cmidrule(lr){2-13}

& \multirow{4}{*}{SocialIQA}
& \gc{SplitLoRA}    
& \gc{40.84\% (+0.00\%)} & \gc{1} 
& \gc{9.14 (+3.75)} & \gc{1} 
& \gc{75913.77 (+0.00)} & \gc{1} 
& \gc{405.15 (+166.21)} & \gc{1} 
& \gc{967.49 (+0.00)} & \gc{1} \\
& & HeteroLoRA    
& 38.79\% (+0.00\%) & 1 
& 153.14 (+63.75) & 2 
& 81438.75 (+0.00) & 1 
& 6787.49 (+2825.62) & 2 
& 1341.58 (+0.00) & 1 \\
& & \gc{FedAvg+LoRA}   
& \gc{40.94\% (+0.00\%)} & \gc{1} 
& \gc{155.04 (+64.58)} & \gc{3} 
& \gc{106179.61 (+0.00)} & \gc{1} 
& \gc{6871.89 (+2862.56)} & \gc{3} 
& \gc{2325.37 (+0.00)} & \gc{1} \\
& & FedProx+LoRA  
& 40.38\% (+0.00\%) & 1 
& 163.02 (+67.92) & 4 
& 111827.46 (+0.00) & 1 
& 7225.40 (+3010.30) & 4 
& 2316.62 (+0.00) & 1 \\

\midrule

\multirow{8}{*}{Reason}
& \multirow{4}{*}{ARC-E}
& \gc{SplitLoRA}    
& \gc{29.30\% (+0.00\%)} & \gc{1} 
& \gc{3.43 (+1.32)} & \gc{1} 
& \gc{29373.67 (+0.00)} & \gc{1} 
& \gc{152.54 (+58.82)} & \gc{1} 
& \gc{963.12 (+0.00)} & \gc{1} \\
& & HeteroLoRA    
& 30.00\% (+0.00\%) & 1 
& 270.77 (+112.80) & 4 
& 69986.43 (+0.00) & 1 
& 12040.74 (+5016.12) & 4 
& 1371.14 (+0.00) & 1 \\
& & \gc{FedAvg+LoRA}   
& \gc{30.18\% (+0.00\%)} & \gc{1} 
& \gc{63.44 (+26.33)} & \gc{2} 
& \gc{44053.24 (+0.00)} & \gc{1} 
& \gc{2821.07 (+1170.85)} & \gc{2} 
& \gc{2250.70 (+0.00)} & \gc{1} \\
& & FedProx+LoRA  
& 30.18\% (+0.00\%) & 1 
& 64.00 (+26.67) & 3 
& 44053.24 (+0.00) & 1 
& 2846.01 (+1185.83) & 3 
& 2286.99 (+0.00) & 1 \\

\cmidrule(lr){2-13}

& \multirow{4}{*}{WinoGrande}
& \gc{SplitLoRA}    
& \gc{50.43\% (+0.00\%)} & \gc{1} 
& \gc{0.28 (+0.00)} & \gc{1} 
& \gc{4959.19 (+0.00)} & \gc{1} 
& \gc{12.62 (+0.00)} & \gc{1} 
& \gc{966.91 (+0.00)} & \gc{1} \\
& & HeteroLoRA    
& 50.51\% (+0.00\%) & 1 
& 18.23 (+7.50) & 3 
& 9755.68 (+0.00) & 1 
& 825.82 (+339.80) & 3 
& 1373.49 (+0.00) & 1 \\
& & \gc{FedAvg+LoRA}   
& \gc{50.51\% (+0.00\%)} & \gc{1} 
& \gc{9.80 (+4.07)} & \gc{2} 
& \gc{6777.42 (+0.00)} & \gc{1} 
& \gc{444.02 (+184.26)} & \gc{2} 
& \gc{2340.61 (+0.00)} & \gc{1} \\
& & FedProx+LoRA  
& 51.30\% (+0.00\%) & 1 
& 65.69 (+27.34) & 4 
& 45182.81 (+0.00) & 1 
& 2976.11 (+1238.87) & 4 
& 2343.87 (+0.00) & 1 \\

\bottomrule
\end{tabular}
}
\end{table*}

\subsection{Results of Gemma 3-1B}

Different from the smaller models, only SplitLoRA is executable under the current edge-system budget, while FedAvg+LoRA, FedProx+LoRA, and HeteroLoRA remain infeasible due to out-of-memory failures. 
Under communication fluctuation, SplitLoRA preserves the same testing accuracy across all tasks, and its communication volume and peak memory remain unchanged. 
Its peak memory stays around $1.32$--$1.33$ GB, showing that the client-side memory footprint remains small even for the larger Gemma 3-1B backbone. 
Although wall-clock time and energy consumption increase under fluctuating bandwidth, SplitLoRA still remains deployable and stable across different tasks. 

\begin{table*}[h]
\centering
\caption{Complete Protocol C results of Gemma 3-1B under communication fluctuation. Values in parentheses denote changes relative to the non-fluctuation setting. ``-'' indicates that the method is not executable under the system budget due to out-of-memory. Ranks are computed by the absolute value of the change, where smaller change indicates better robustness.}
\label{tab:protocol_c_gemma1b_comm_fluctuation}
\renewcommand{\arraystretch}{1.08}
\setlength{\tabcolsep}{3pt}
\scriptsize
\providecommand{\gc}[1]{\cellcolor{gray!10}#1}
\resizebox{\textwidth}{!}{
\begin{tabular}{lllcccccccccc}
\toprule
\textbf{Task Type} 
& \textbf{Task} 
& \textbf{Methods} 
& \textbf{Testing Accuracy} 
& \textbf{Rank} 
& \makecell{\textbf{Wall-clock}\\\textbf{time (h)}} 
& \textbf{Rank} 
& \makecell{\textbf{Communication}\\\textbf{volume (MB)}} 
& \textbf{Rank} 
& \makecell{\textbf{Energy}\\\textbf{consumption (kJ)}} 
& \textbf{Rank} 
& \makecell{\textbf{Peak}\\\textbf{memory (MB)}} 
& \textbf{Rank} \\
\midrule

\multirow{8}{*}{Verify}
& \multirow{4}{*}{BoolQ}
& \gc{SplitLoRA}    
& \gc{64.25\% (+0.00\%)} & \gc{1} 
& \gc{2.32 (+0.83)} & \gc{1} 
& \gc{63401.95 (+0.00)} & \gc{1} 
& \gc{69.52 (+24.93)} & \gc{1} 
& \gc{1323.60 (+0.00)} & \gc{1} \\
& & HeteroLoRA    
& - & - 
& - & - 
& - & - 
& - & - 
& - & - \\
& & \gc{FedAvg+LoRA}   
& \gc{-} & \gc{-} 
& \gc{-} & \gc{-} 
& \gc{-} & \gc{-} 
& \gc{-} & \gc{-} 
& \gc{-} & \gc{-} \\
& & FedProx+LoRA  
& - & - 
& - & - 
& - & - 
& - & - 
& - & - \\

\cmidrule(lr){2-13}

& \multirow{4}{*}{QNLI}
& \gc{SplitLoRA}    
& \gc{64.52\% (+0.00\%)} & \gc{1} 
& \gc{3.13 (+1.25)} & \gc{1} 
& \gc{75478.51 (+0.00)} & \gc{1} 
& \gc{100.92 (+40.28)} & \gc{1} 
& \gc{1331.26 (+0.00)} & \gc{1} \\
& & HeteroLoRA    
& - & - 
& - & - 
& - & - 
& - & - 
& - & - \\
& & \gc{FedAvg+LoRA}   
& \gc{-} & \gc{-} 
& \gc{-} & \gc{-} 
& \gc{-} & \gc{-} 
& \gc{-} & \gc{-} 
& \gc{-} & \gc{-} \\
& & FedProx+LoRA  
& - & - 
& - & - 
& - & - 
& - & - 
& - & - \\

\midrule

\multirow{12}{*}{Choose}
& \multirow{4}{*}{PIQA}
& \gc{SplitLoRA}    
& \gc{53.81\% (+0.00\%)} & \gc{1} 
& \gc{3.80 (+1.57)} & \gc{1} 
& \gc{95606.11 (+0.00)} & \gc{1} 
& \gc{112.52 (+46.40)} & \gc{1} 
& \gc{1324.37 (+0.00)} & \gc{1} \\
& & HeteroLoRA    
& - & - 
& - & - 
& - & - 
& - & - 
& - & - \\
& & \gc{FedAvg+LoRA}   
& \gc{-} & \gc{-} 
& \gc{-} & \gc{-} 
& \gc{-} & \gc{-} 
& \gc{-} & \gc{-} 
& \gc{-} & \gc{-} \\
& & FedProx+LoRA  
& - & - 
& - & - 
& - & - 
& - & - 
& - & - \\

\cmidrule(lr){2-13}

& \multirow{4}{*}{HellaSwag}
& \gc{SplitLoRA}    
& \gc{25.33\% (+0.00\%)} & \gc{1} 
& \gc{2.15 (+0.83)} & \gc{1} 
& \gc{55854.10 (+0.00)} & \gc{1} 
& \gc{65.81 (+25.54)} & \gc{1} 
& \gc{1327.05 (+0.00)} & \gc{1} \\
& & HeteroLoRA    
& - & - 
& - & - 
& - & - 
& - & - 
& - & - \\
& & \gc{FedAvg+LoRA}   
& \gc{-} & \gc{-} 
& \gc{-} & \gc{-} 
& \gc{-} & \gc{-} 
& \gc{-} & \gc{-} 
& \gc{-} & \gc{-} \\
& & FedProx+LoRA  
& - & - 
& - & - 
& - & - 
& - & - 
& - & - \\

\cmidrule(lr){2-13}

& \multirow{4}{*}{SocialIQA}
& \gc{SplitLoRA}    
& \gc{45.70\% (+0.00\%)} & \gc{1} 
& \gc{2.81 (+1.16)} & \gc{1} 
& \gc{67930.66 (+0.00)} & \gc{1} 
& \gc{85.03 (+34.96)} & \gc{1} 
& \gc{1325.86 (+0.00)} & \gc{1} \\
& & HeteroLoRA    
& - & - 
& - & - 
& - & - 
& - & - 
& - & - \\
& & \gc{FedAvg+LoRA}   
& \gc{-} & \gc{-} 
& \gc{-} & \gc{-} 
& \gc{-} & \gc{-} 
& \gc{-} & \gc{-} 
& \gc{-} & \gc{-} \\
& & FedProx+LoRA  
& - & - 
& - & - 
& - & - 
& - & - 
& - & - \\

\midrule

\multirow{8}{*}{Reason}
& \multirow{4}{*}{ARC-E}
& \gc{SplitLoRA}    
& \gc{28.07\% (+0.00\%)} & \gc{1} 
& \gc{0.73 (+0.28)} & \gc{1} 
& \gc{19121.22 (+0.00)} & \gc{1} 
& \gc{21.46 (+8.27)} & \gc{1} 
& \gc{1323.40 (+0.00)} & \gc{1} \\
& & HeteroLoRA    
& - & - 
& - & - 
& - & - 
& - & - 
& - & - \\
& & \gc{FedAvg+LoRA}   
& \gc{-} & \gc{-} 
& \gc{-} & \gc{-} 
& \gc{-} & \gc{-} 
& \gc{-} & \gc{-} 
& \gc{-} & \gc{-} \\
& & FedProx+LoRA  
& - & - 
& - & - 
& - & - 
& - & - 
& - & - \\

\cmidrule(lr){2-13}

& \multirow{4}{*}{WinoGrande}
& \gc{SplitLoRA}    
& \gc{51.62\% (+0.00\%)} & \gc{1} 
& \gc{3.09 (+1.25)} & \gc{1} 
& \gc{80007.22 (+0.00)} & \gc{1} 
& \gc{92.40 (+37.32)} & \gc{1} 
& \gc{1325.58 (+0.00)} & \gc{1} \\
& & HeteroLoRA    
& - & - 
& - & - 
& - & - 
& - & - 
& - & - \\
& & \gc{FedAvg+LoRA}   
& \gc{-} & \gc{-} 
& \gc{-} & \gc{-} 
& \gc{-} & \gc{-} 
& \gc{-} & \gc{-} 
& \gc{-} & \gc{-} \\
& & FedProx+LoRA  
& - & - 
& - & - 
& - & - 
& - & - 
& - & - \\

\bottomrule
\end{tabular}
}
\end{table*}

\subsection{Overall ranking of methods under protocol C}

\begin{figure*}[h]
\centering
\includegraphics[width=\textwidth]{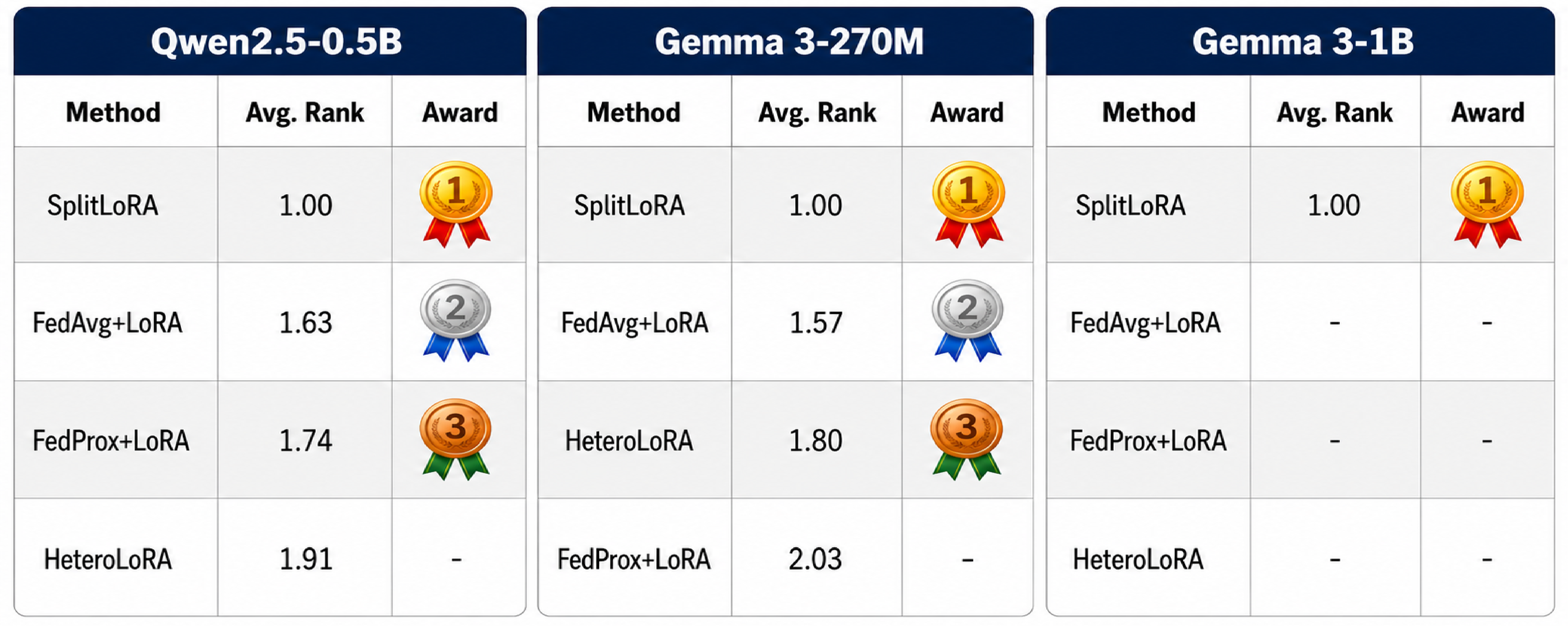}
\caption{Overall ranking of methods under protocol C. }
\label{fig:protocolCranking}
\vspace{-5mm}
\end{figure*}

Figure~\ref{fig:protocolCranking} summarizes the overall robustness ranking of different methods under Protocol C. 
Unlike Protocol B, which ranks methods by their absolute system cost to reach target accuracy, Protocol C ranks methods by the magnitude of performance variation under system perturbation, where a smaller change indicates stronger robustness. 
Overall, SplitLoRA achieves the best average rank across all three model scales, showing that its split architecture is less sensitive to communication fluctuation and can maintain stable testing accuracy, wall-clock time, energy consumption, and memory footprint under unstable edge conditions. 
For Qwen2.5-0.5B and Gemma 3-270M, FedAvg+LoRA ranks second overall, while FedProx+LoRA and HeteroLoRA show larger performance variations depending on the model scale and task. 
This indicates that methods with strong final accuracy or low cost under stable settings do not necessarily remain robust under system perturbations. 

Therefore, Figure~\ref{fig:protocolCranking} highlights a key Protocol C insight: \textit{Edge-system perturbations affect different methods differently, leading to method-dependent robustness in both model quality and system efficiency.}

\section{Discussion on overall ranking across methods}
\label{AppendixD}

\begin{figure*}[h]
\centering
\includegraphics[width=\textwidth]{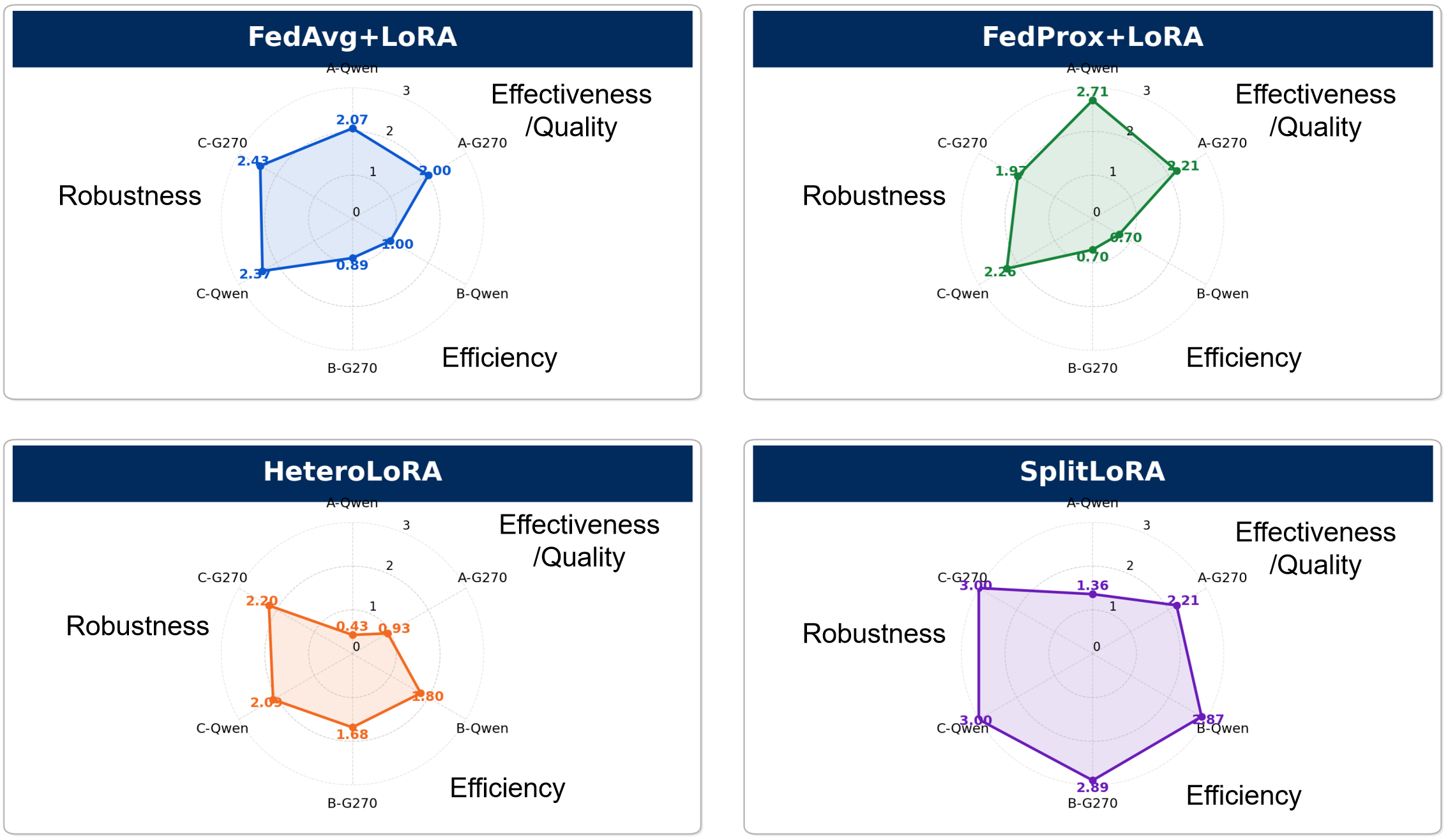}
\caption{Overall ranking across methods. 
For each method, the rank under each protocol--model setting is first obtained from the corresponding overall-ranking table, and then averaged to summarize its cross-protocol performance. 
In the radar chart, each vertex corresponds to one protocol--model pair: \textit{A-Qwen}, \textit{A-G270}, \textit{B-Qwen}, \textit{B-G270}, \textit{C-Qwen}, and \textit{C-G270}, where A/B/C denote Protocols A/B/C, \textit{Qwen} denotes Qwen2.5-0.5B, and \textit{G270} denotes Gemma 3-270M. 
The plotted value at each vertex is transformed from the average rank as $4-\mathrm{Avg.\ Rank}$, so a larger radius indicates a better overall ranking. 
Thus, methods with better ranks form larger hexagons. 
Gemma 3-1B is not included in the hexagon axes because only SplitLoRA is executable under our system budget, while the other methods encounter out-of-memory errors.}
\label{fig:ranking}
\end{figure*}

Figure~\ref{fig:ranking} provides a cross-protocol comparison of the four evaluated federated LLM fine-tuning methods. 

The results reveal clear trade-offs among methods. 
FedProx+LoRA achieves strong ranks under Protocol A, because the proximal regularizer stabilizes local LoRA updates and improves final quality. 
However, its advantage in final accuracy does not translate into system efficiency: under Protocol B, FedProx+LoRA shows much lower radar values, indicating that it often requires longer wall-clock time and higher energy consumption to reach the same target accuracy. 
FedAvg+LoRA shows a similar but slightly more balanced pattern: it is competitive in final quality and relatively stable under communication fluctuation, but it is not the most efficient method when cost-to-target is considered.

HeteroLoRA shows moderate performance under Protocol B and Protocol C, but its Protocol A ranking is relatively weak. 
This is consistent with the fact that heterogeneous LoRA aggregation introduces additional approximation noise, especially when adapters with different ranks are aligned or padded before aggregation. 
As a result, HeteroLoRA can improve adaptability to heterogeneous clients, but this does not always lead to the best model quality or the lowest system cost.

SplitLoRA shows the most balanced and deployment-friendly behavior across protocols. 
Although it is not always the best method in final accuracy under Protocol A, it achieves the strongest ranks under Protocol B and Protocol C. 
This indicates that SplitLoRA is particularly effective when the evaluation objective shifts from accuracy-only performance to practical edge deployment constraints. 
Moreover, for Gemma 3-1B, SplitLoRA is the only executable method under our system budget, while FedAvg+LoRA, FedProx+LoRA, and HeteroLoRA encounter out-of-memory errors. 

Overall, these cross-protocol results lead to a key conclusion: \textit{methods that appear favorable under accuracy-only evaluation may become suboptimal once realistic edge constraints and system perturbations are taken into account.}

\section{Detailed dynamics of fine-tuning process}
\label{AppendixE}

This section provides detailed fine-tuning trajectories under the experimental settings of Protocols A and B. 
While the previous sections mainly report testing results, including testing loss, testing accuracy, and system-level costs, here we further examine the training dynamics of different methods. 
Specifically, we plot the training loss over wall-clock time for four federated LLM fine-tuning methods, three backbone models, and seven benchmark datasets. 
These curves provide a more fine-grained view of how each method progresses during fine-tuning, revealing differences in convergence speed, training stability, and execution efficiency that may not be fully reflected by the final testing metrics alone. 
By comparing training loss against wall-clock time, we can directly observe whether a method reaches lower loss quickly, whether it suffers from slow system execution, and how the training behavior changes across model scales and task types.

\begin{figure*}[h]
\centering

\begin{subfigure}[t]{0.48\textwidth}
    \centering
    \includegraphics[width=\textwidth]{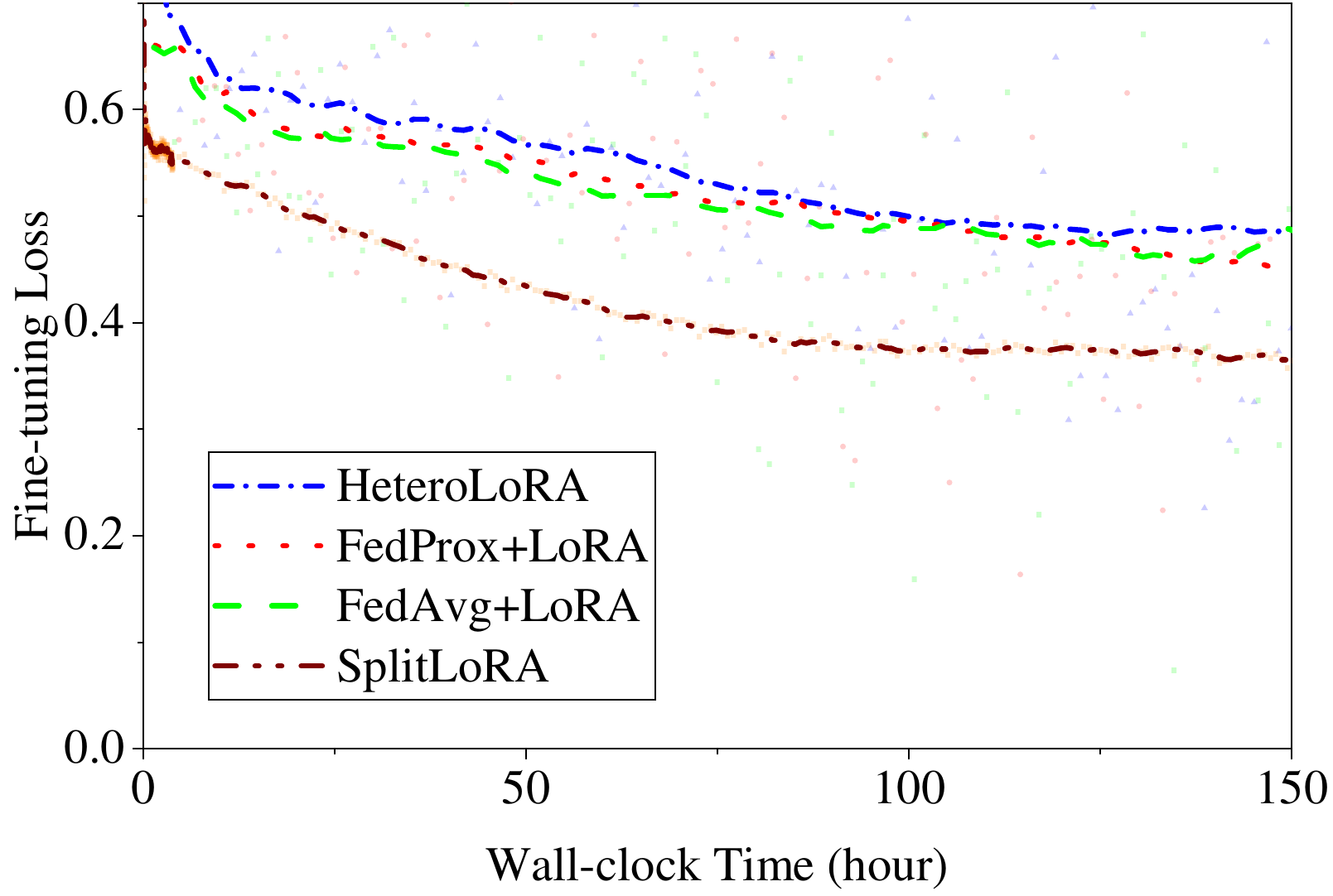}
    \caption{BoolQ}
    \label{fig:qwen25_boolq_loss}
\end{subfigure}
\hfill
\begin{subfigure}[t]{0.48\textwidth}
    \centering
    \includegraphics[width=\textwidth]{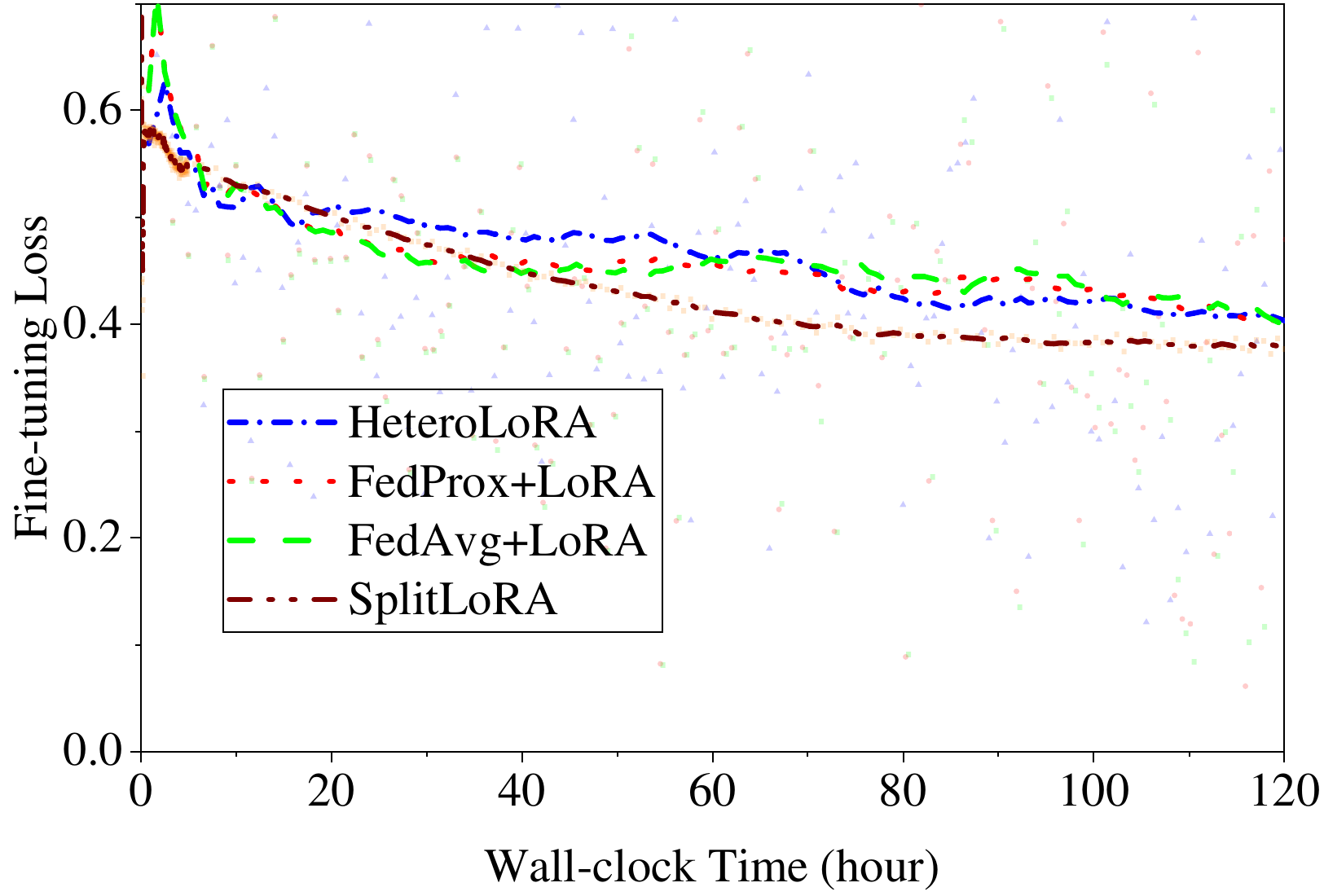}
    \caption{QNLI}
    \label{fig:qwen25_qnli_loss}
\end{subfigure}

\vspace{2mm}

\begin{subfigure}[t]{0.48\textwidth}
    \centering
    \includegraphics[width=\textwidth]{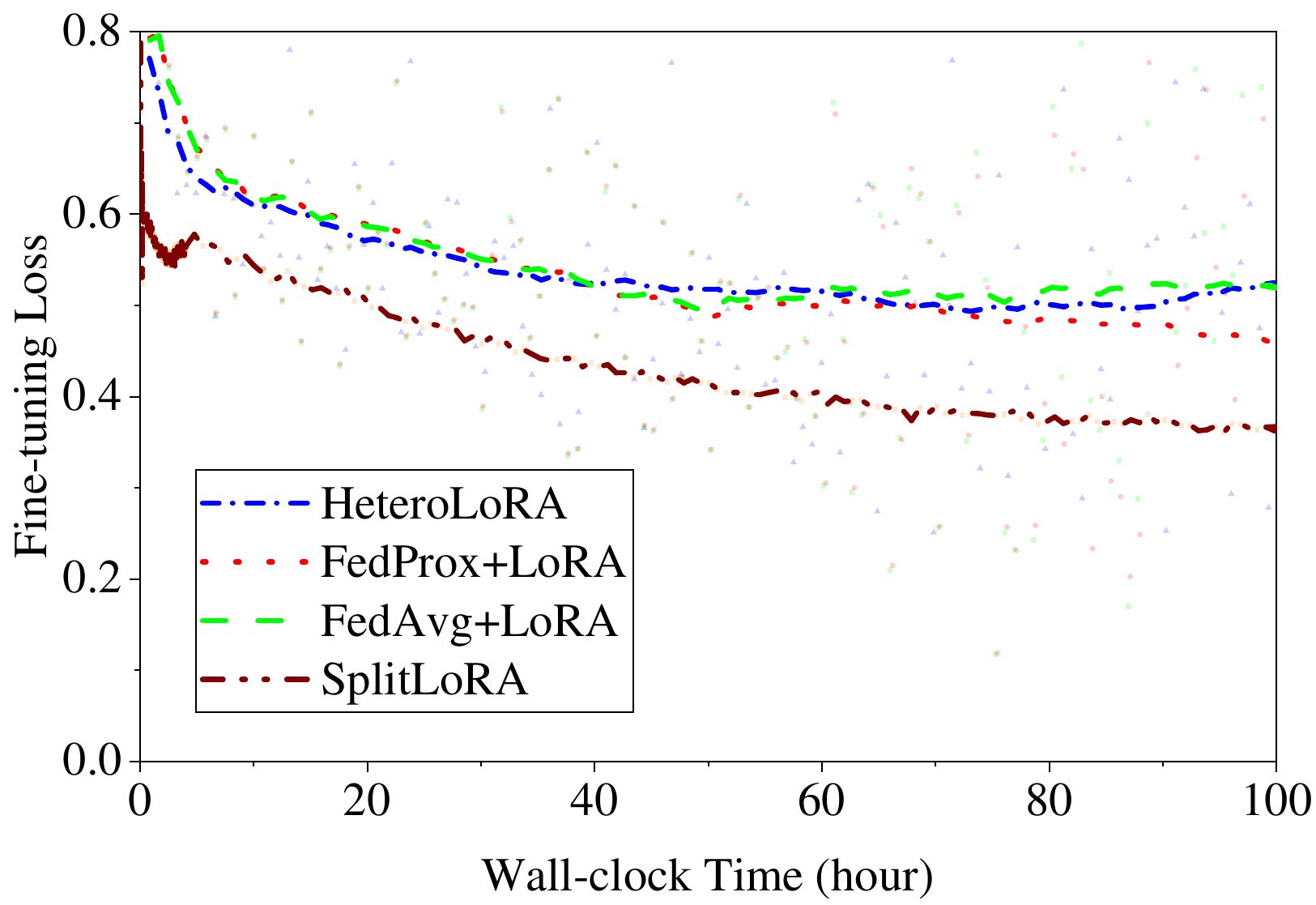}
    \caption{PIQA}
    \label{fig:qwen25_piqa_loss}
\end{subfigure}
\hfill
\begin{subfigure}[t]{0.48\textwidth}
    \centering
    \includegraphics[width=\textwidth]{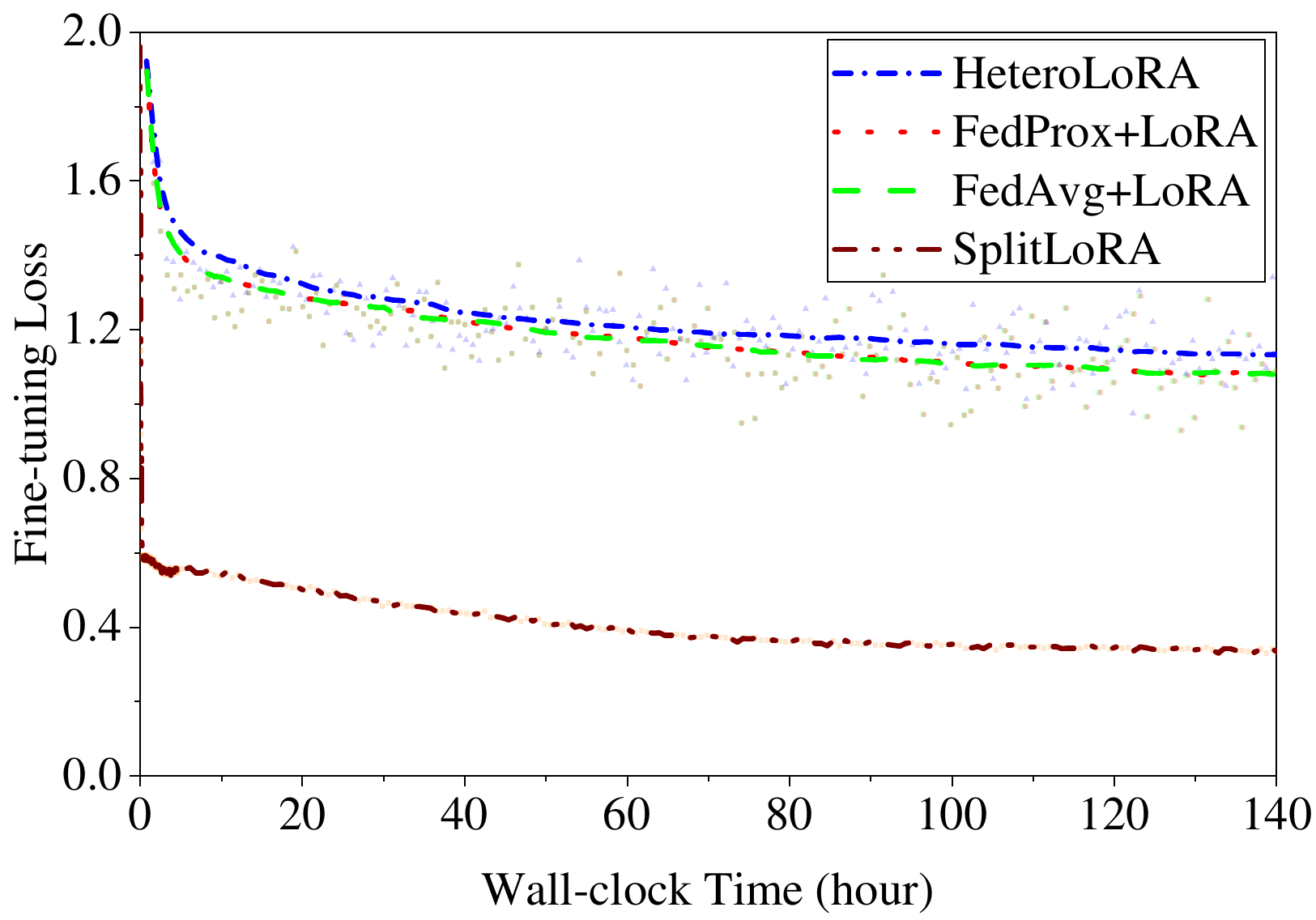}
    \caption{HellaSwag}
    \label{fig:qwen25_hellaswag_loss}
\end{subfigure}

\vspace{2mm}

\begin{subfigure}[t]{0.48\textwidth}
    \centering
    \includegraphics[width=\textwidth]{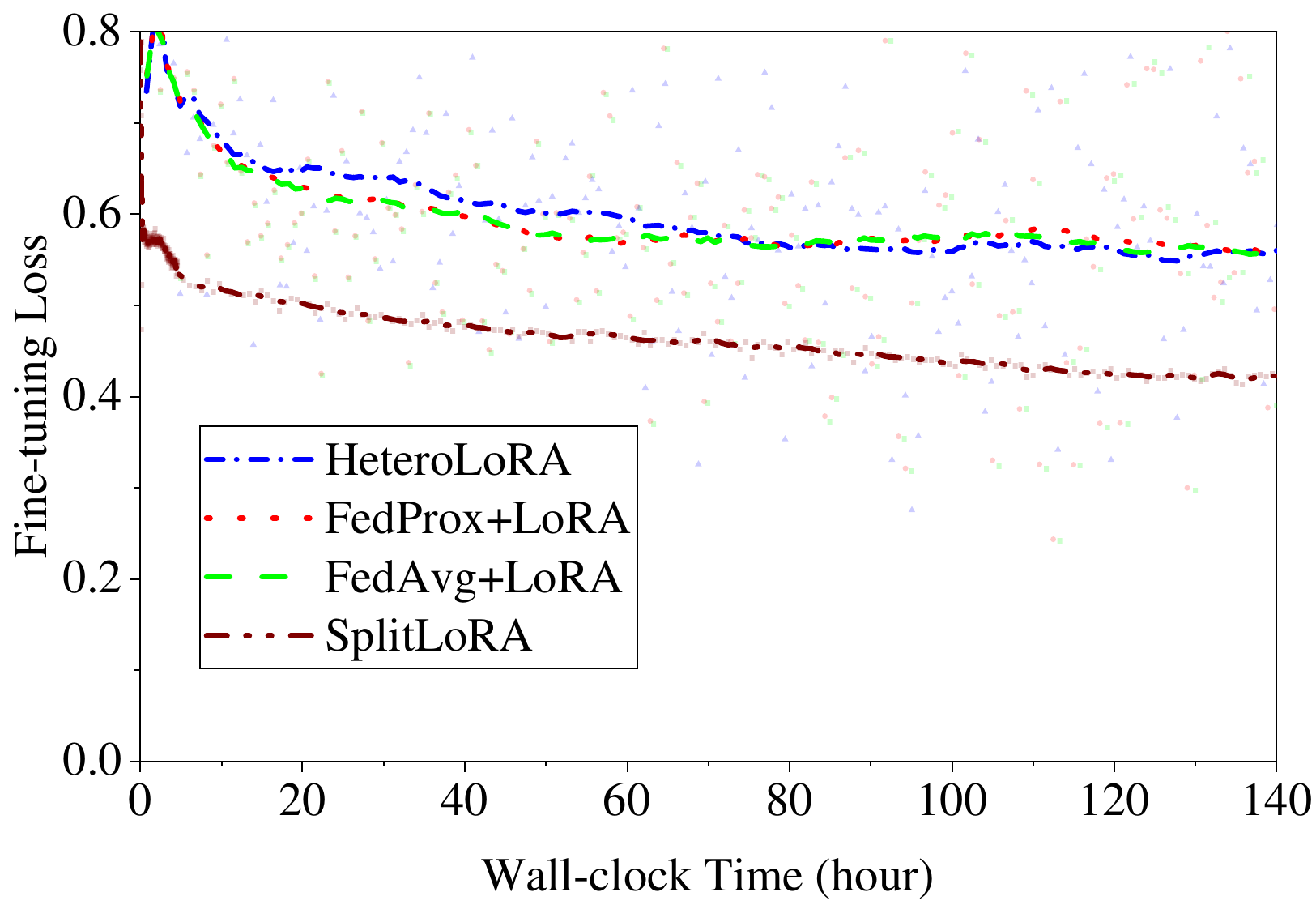}
    \caption{SocialIQA}
    \label{fig:qwen25_socialiqa_loss}
\end{subfigure}
\hfill
\begin{subfigure}[t]{0.48\textwidth}
    \centering
    \includegraphics[width=\textwidth]{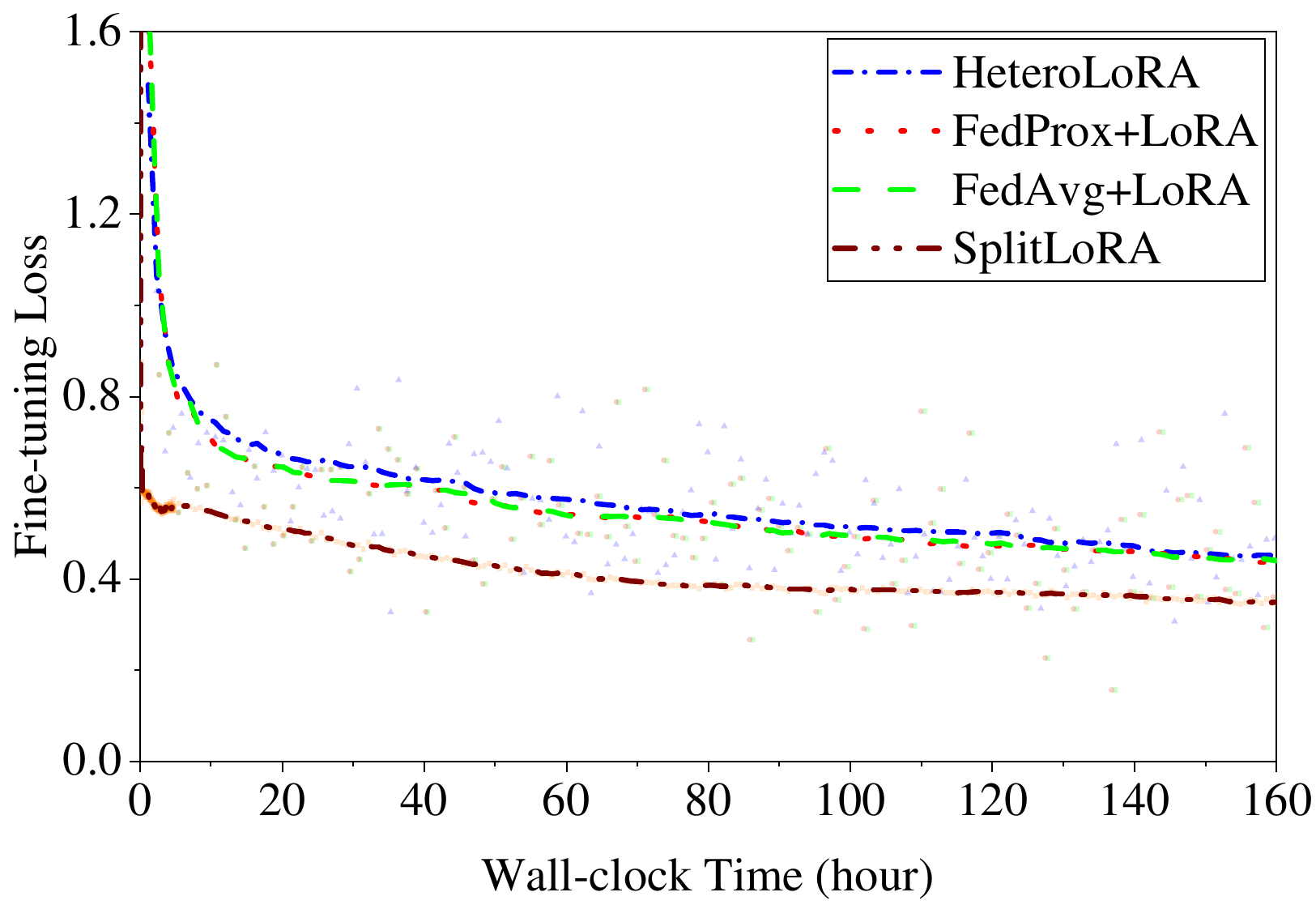}
    \caption{ARC-E}
    \label{fig:qwen25_arce_loss}
\end{subfigure}

\vspace{2mm}

\begin{subfigure}[t]{0.48\textwidth}
    \centering
    \includegraphics[width=\textwidth]{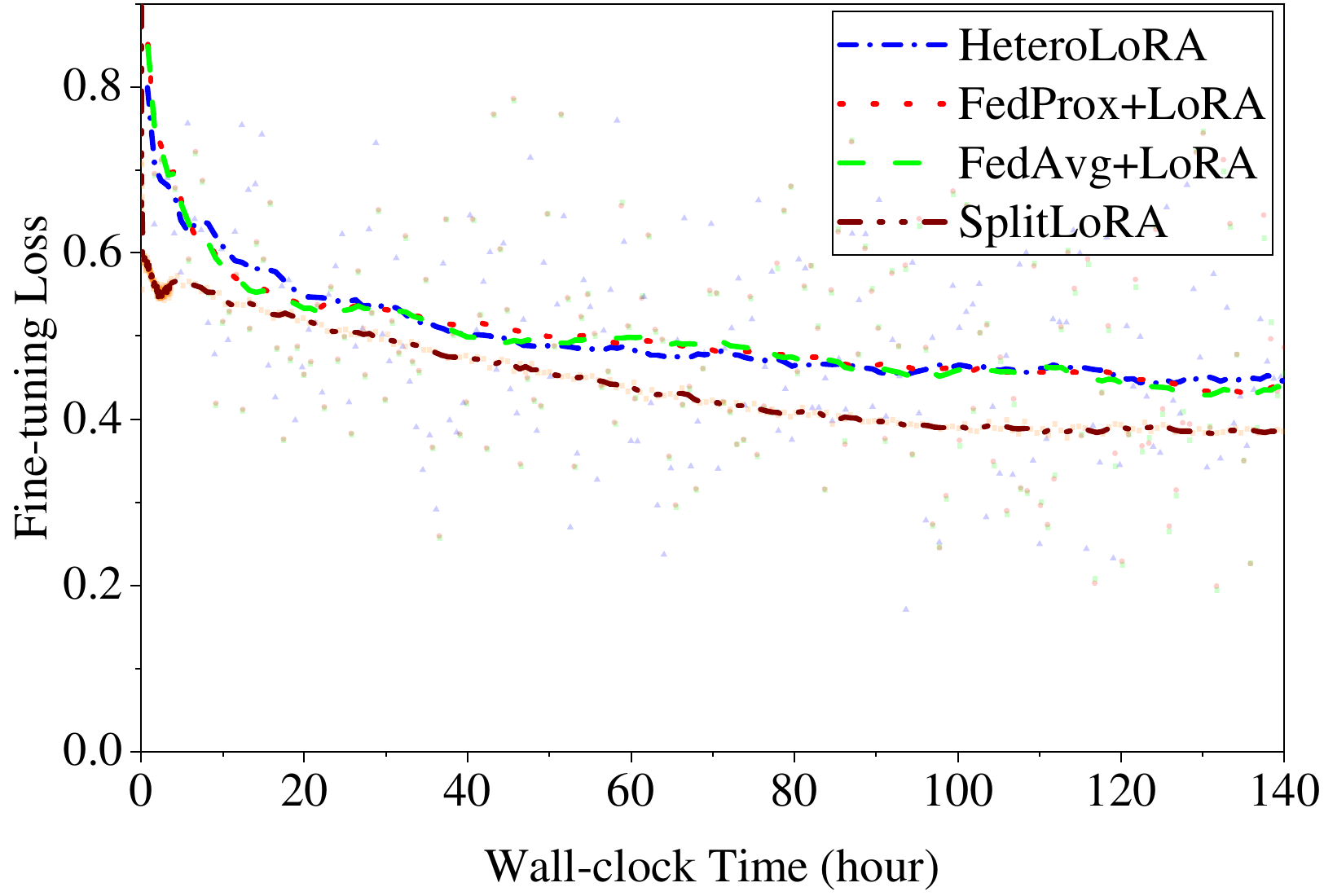}
    \caption{WinoGrande}
    \label{fig:qwen25_winogrande_loss}
\end{subfigure}

\caption{Training loss curves of the four federated fine-tuning methods on Qwen2.5-0.5B across seven datasets under the experimental settings of Protocols A and B. The curves show the evolution of training loss with respect to wall-clock time.}
\label{fig:qwen25_loss_curves}
\vspace{-4mm}
\end{figure*}

\begin{figure*}[h]
\centering

\begin{subfigure}[t]{0.48\textwidth}
    \centering
    \includegraphics[width=\textwidth]{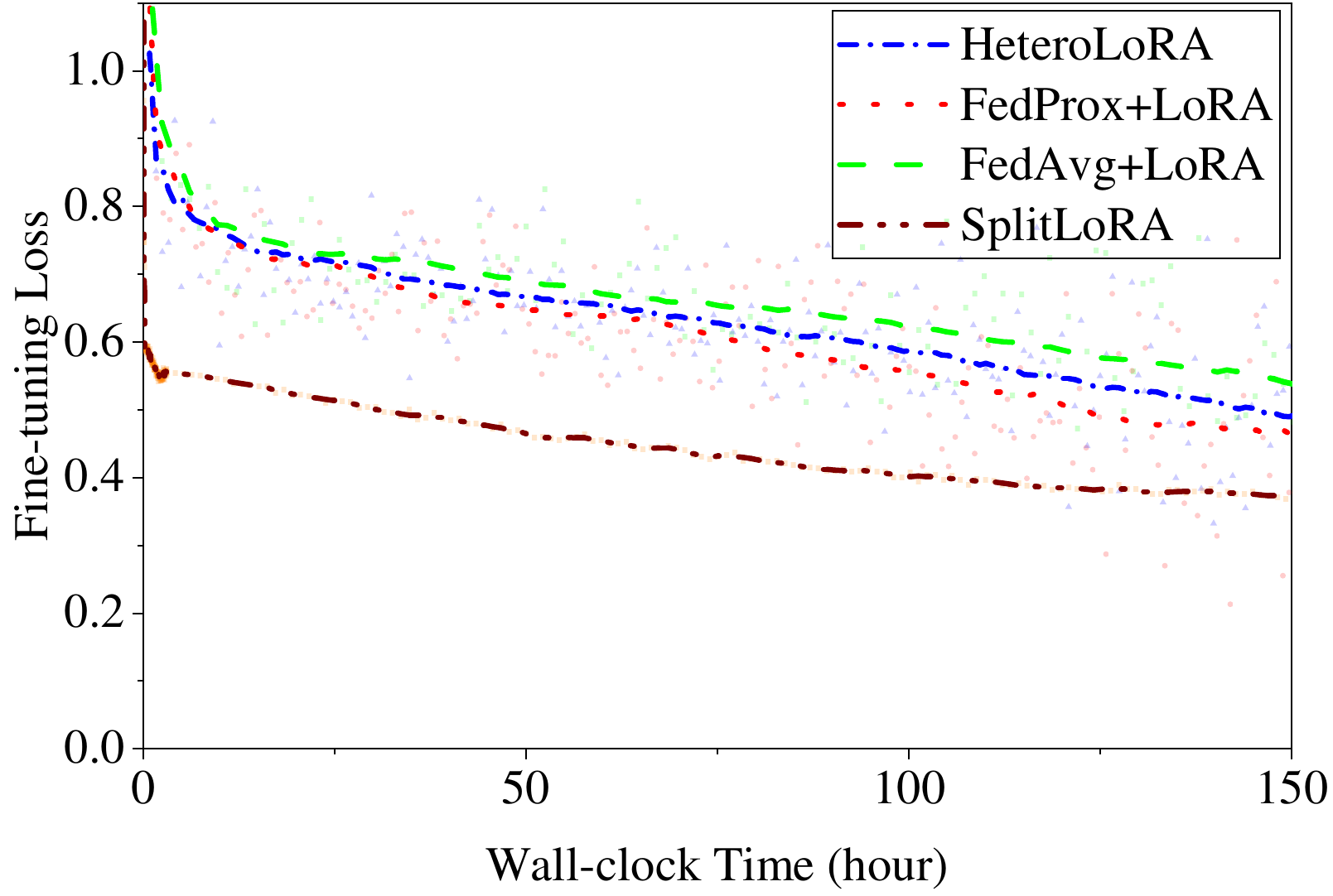}
    \caption{BoolQ}
    \label{fig:gemma270_boolq_loss}
\end{subfigure}
\hfill
\begin{subfigure}[t]{0.48\textwidth}
    \centering
    \includegraphics[width=\textwidth]{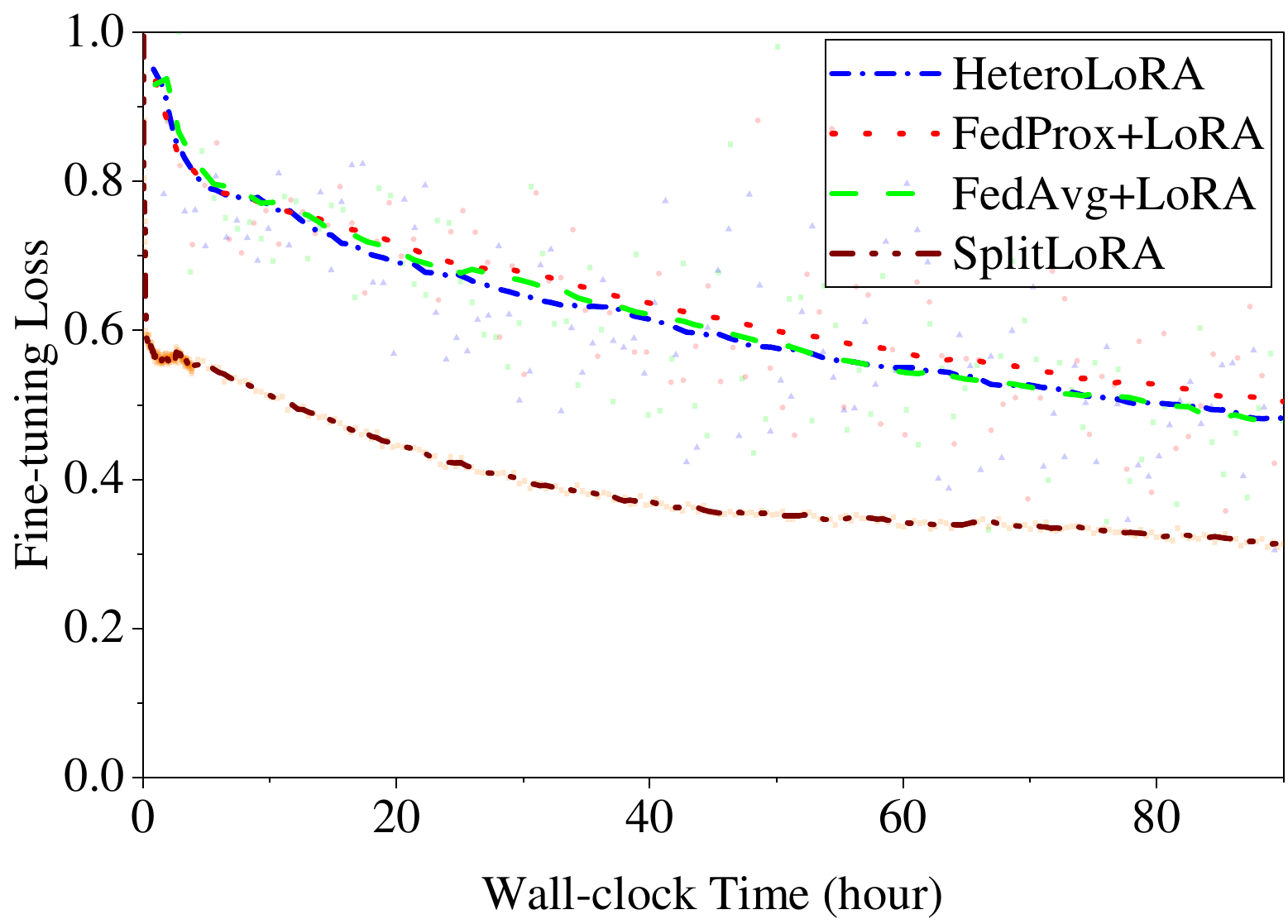}
    \caption{QNLI}
    \label{fig:gemma270_qnli_loss}
\end{subfigure}

\vspace{2mm}

\begin{subfigure}[t]{0.48\textwidth}
    \centering
    \includegraphics[width=\textwidth]{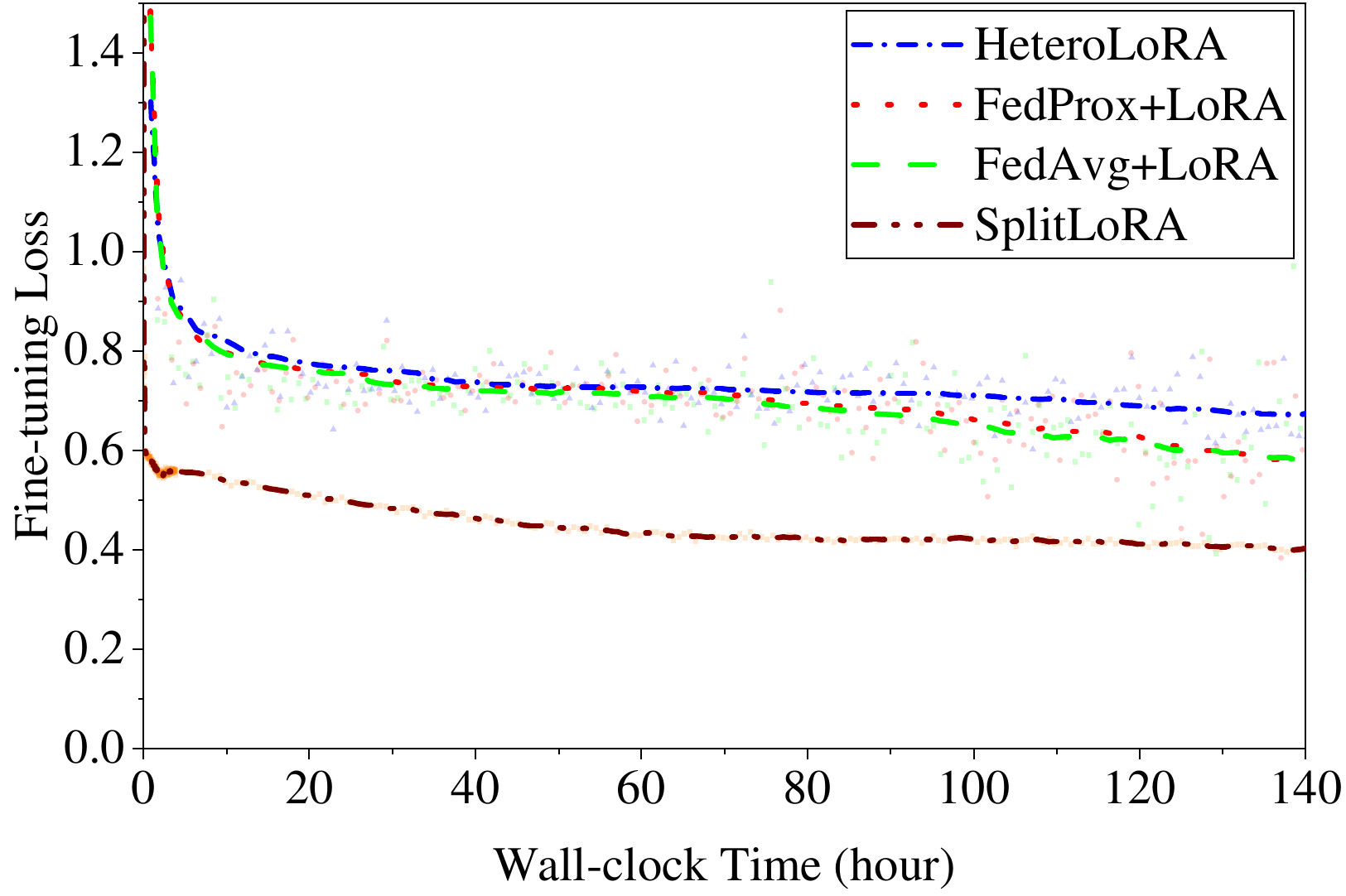}
    \caption{PIQA}
    \label{fig:gemma270_piqa_loss}
\end{subfigure}
\hfill
\begin{subfigure}[t]{0.48\textwidth}
    \centering
    \includegraphics[width=\textwidth]{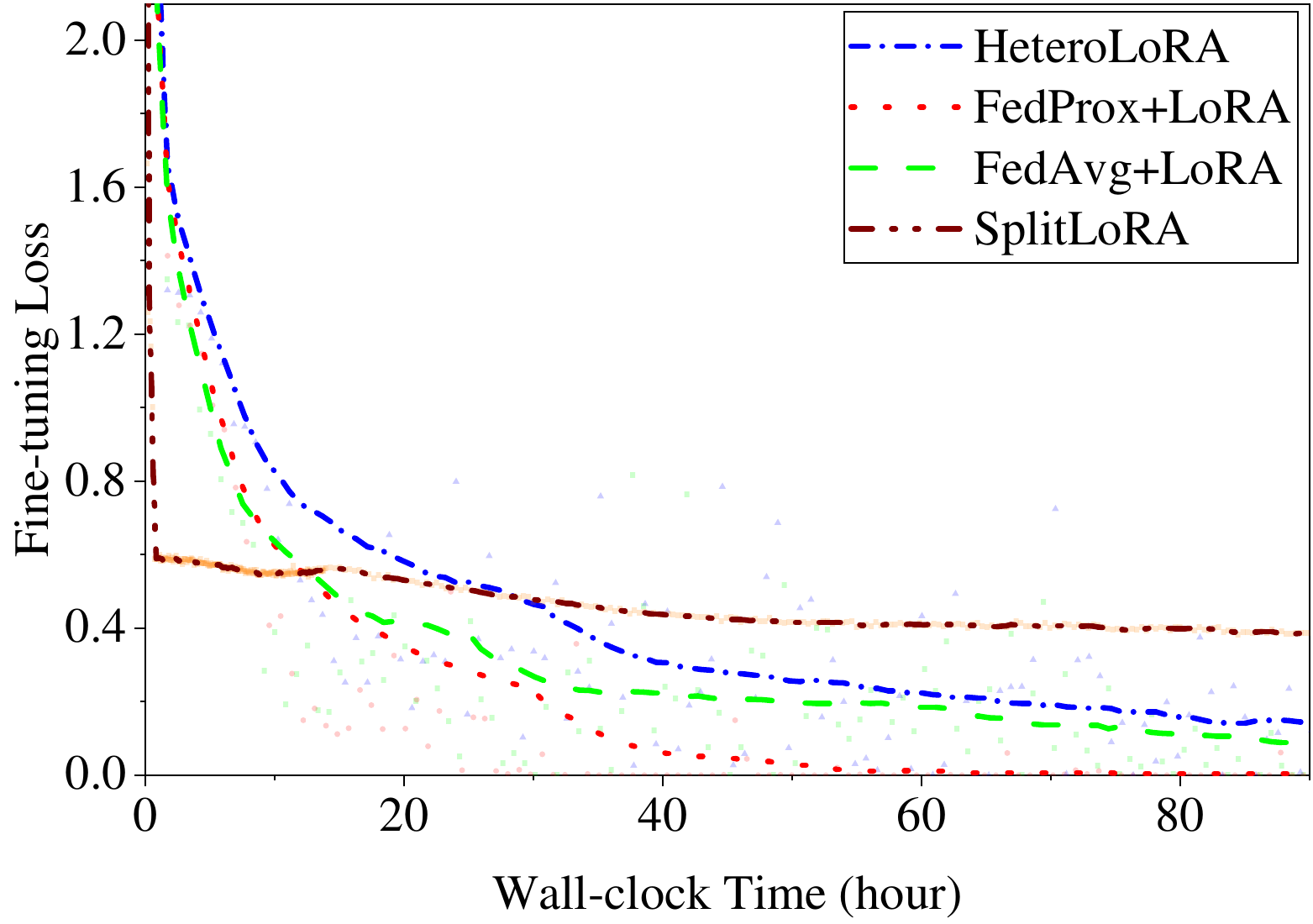}
    \caption{HellaSwag}
    \label{fig:gemma270_hellaswag_loss}
\end{subfigure}

\vspace{2mm}

\begin{subfigure}[t]{0.48\textwidth}
    \centering
    \includegraphics[width=\textwidth]{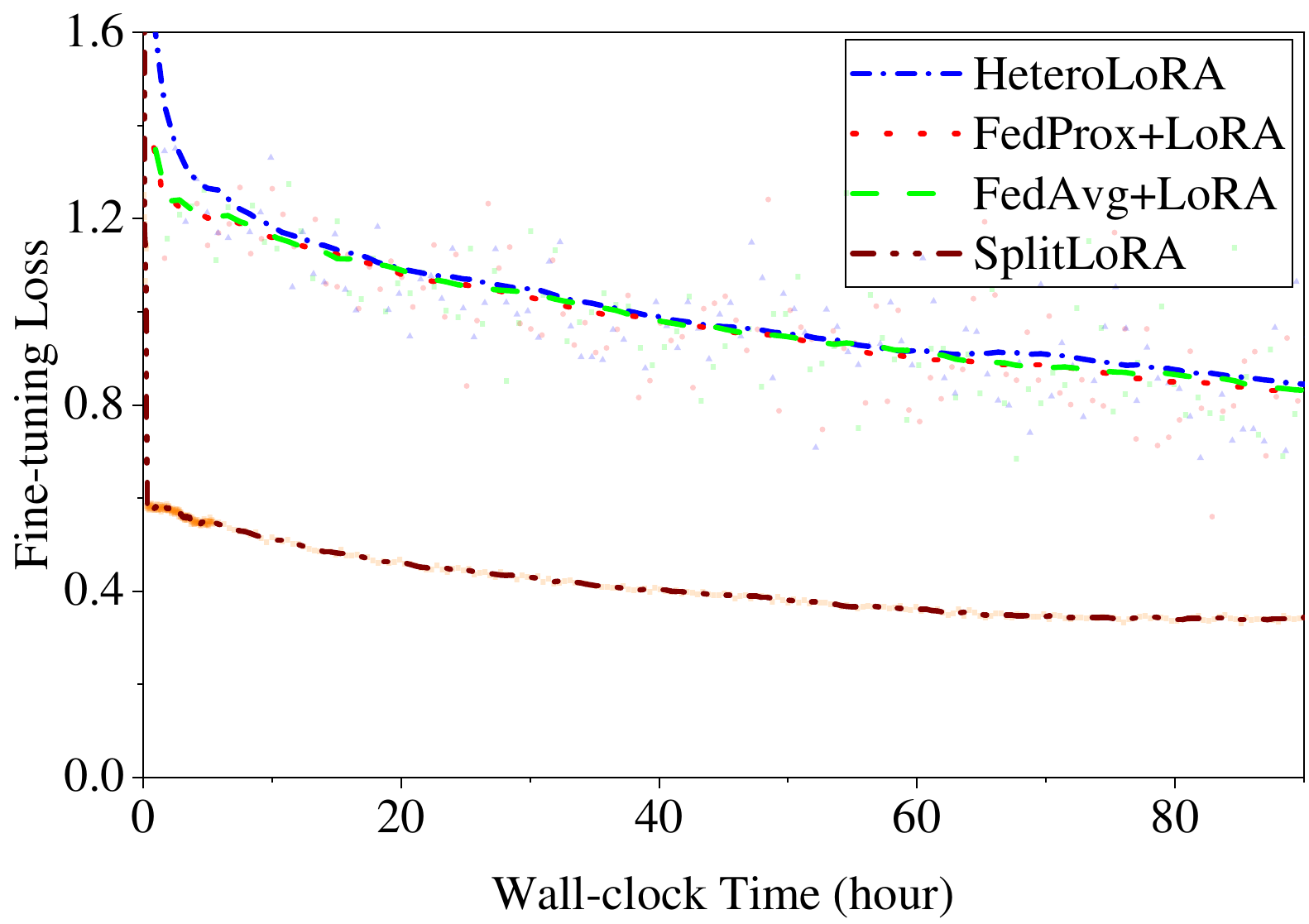}
    \caption{SocialIQA}
    \label{fig:gemma270_socialiqa_loss}
\end{subfigure}
\hfill
\begin{subfigure}[t]{0.48\textwidth}
    \centering
    \includegraphics[width=\textwidth]{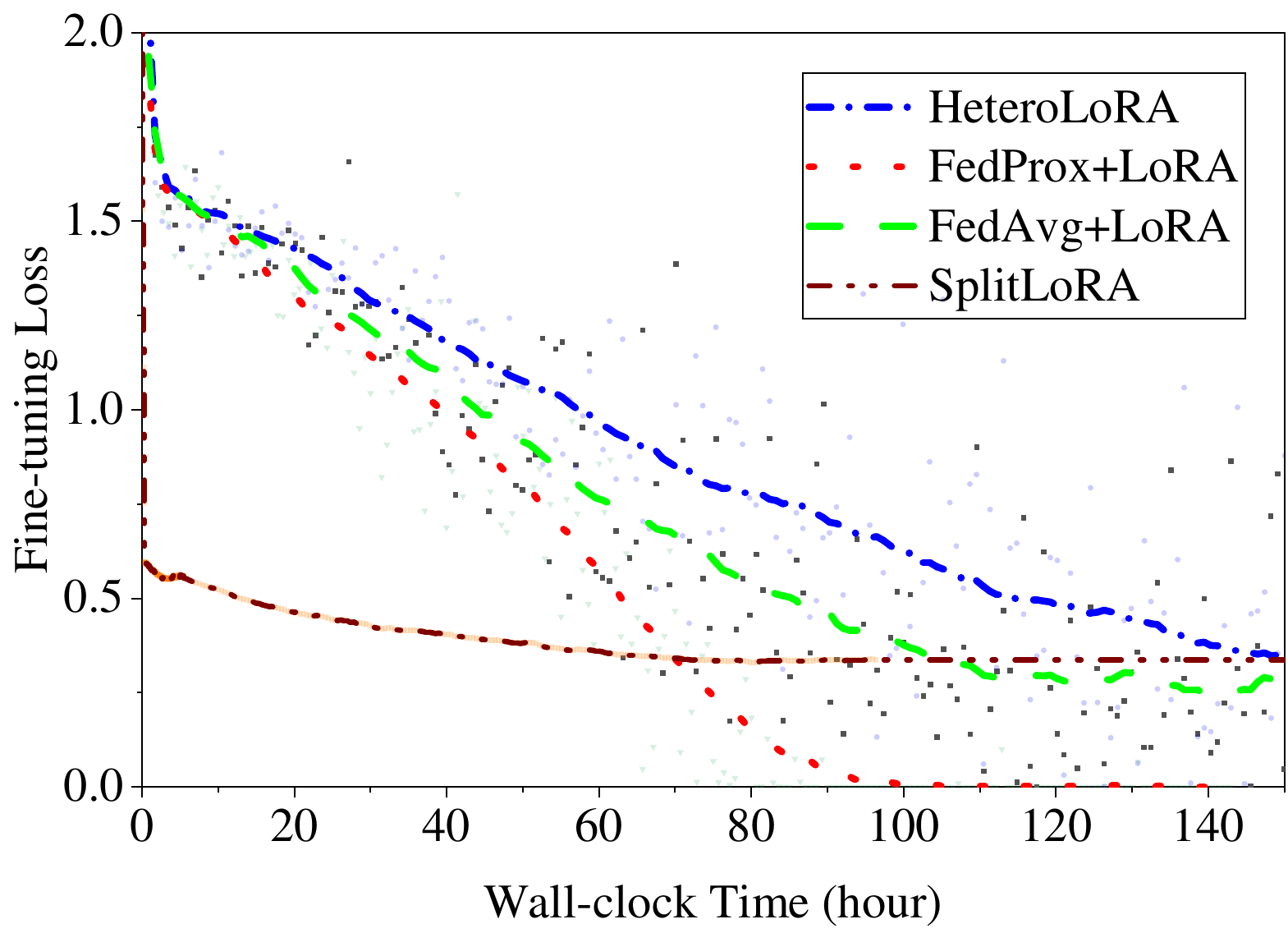}
    \caption{ARC-E}
    \label{fig:gemma270_arce_loss}
\end{subfigure}

\vspace{2mm}

\begin{subfigure}[t]{0.48\textwidth}
    \centering
    \includegraphics[width=\textwidth]{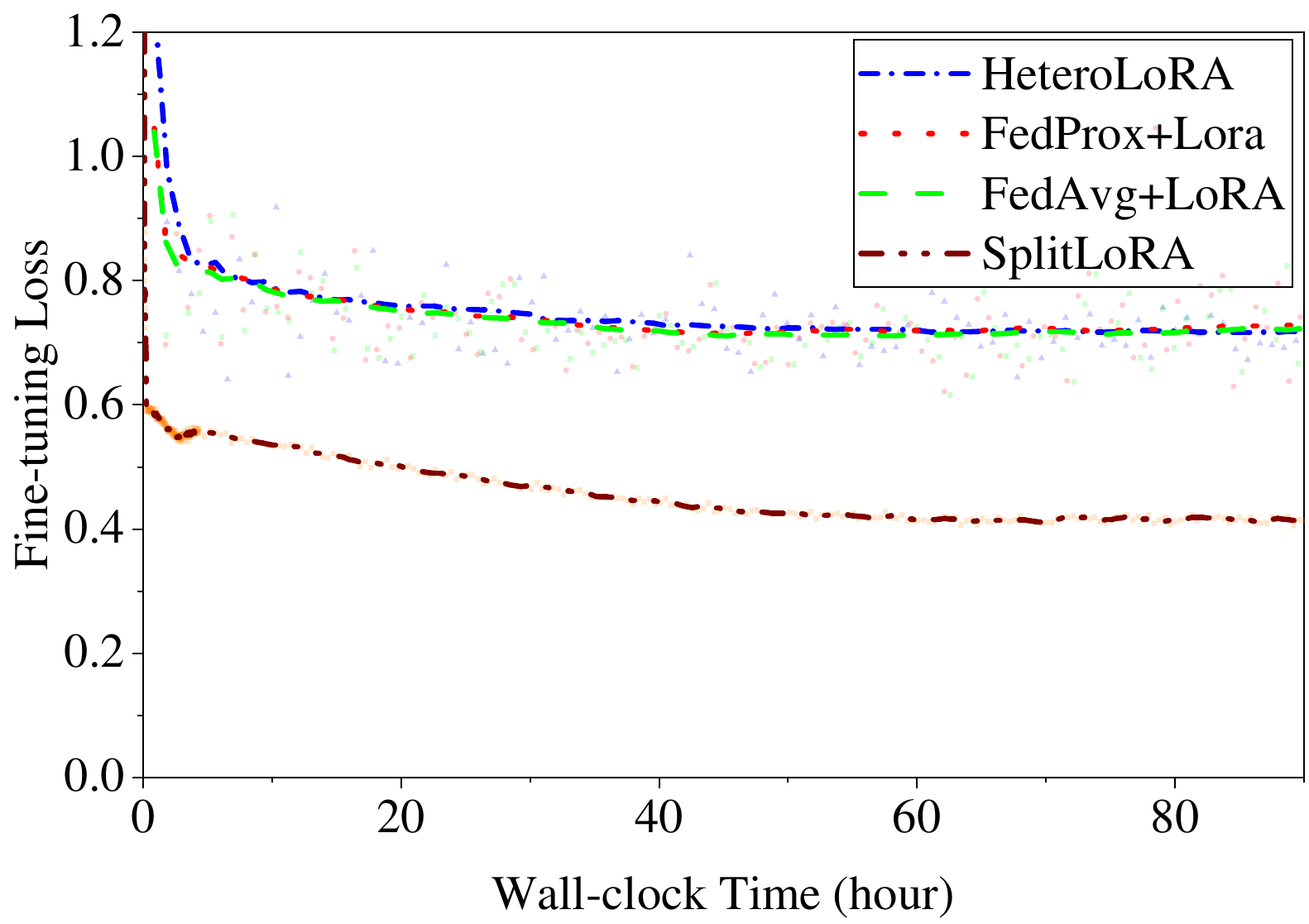}
    \caption{WinoGrande}
    \label{fig:gemma270_winogrande_loss}
\end{subfigure}

\caption{Training loss curves of the four federated fine-tuning methods on Gemma 3-270M across seven datasets under the experimental settings of Protocols A and B. The curves show the evolution of training loss with respect to wall-clock time.}
\label{fig:gemma270_loss_curves}
\vspace{-4mm}
\end{figure*}

\begin{figure*}[t]
\centering

\begin{subfigure}[t]{0.48\textwidth}
    \centering
    \includegraphics[width=\textwidth]{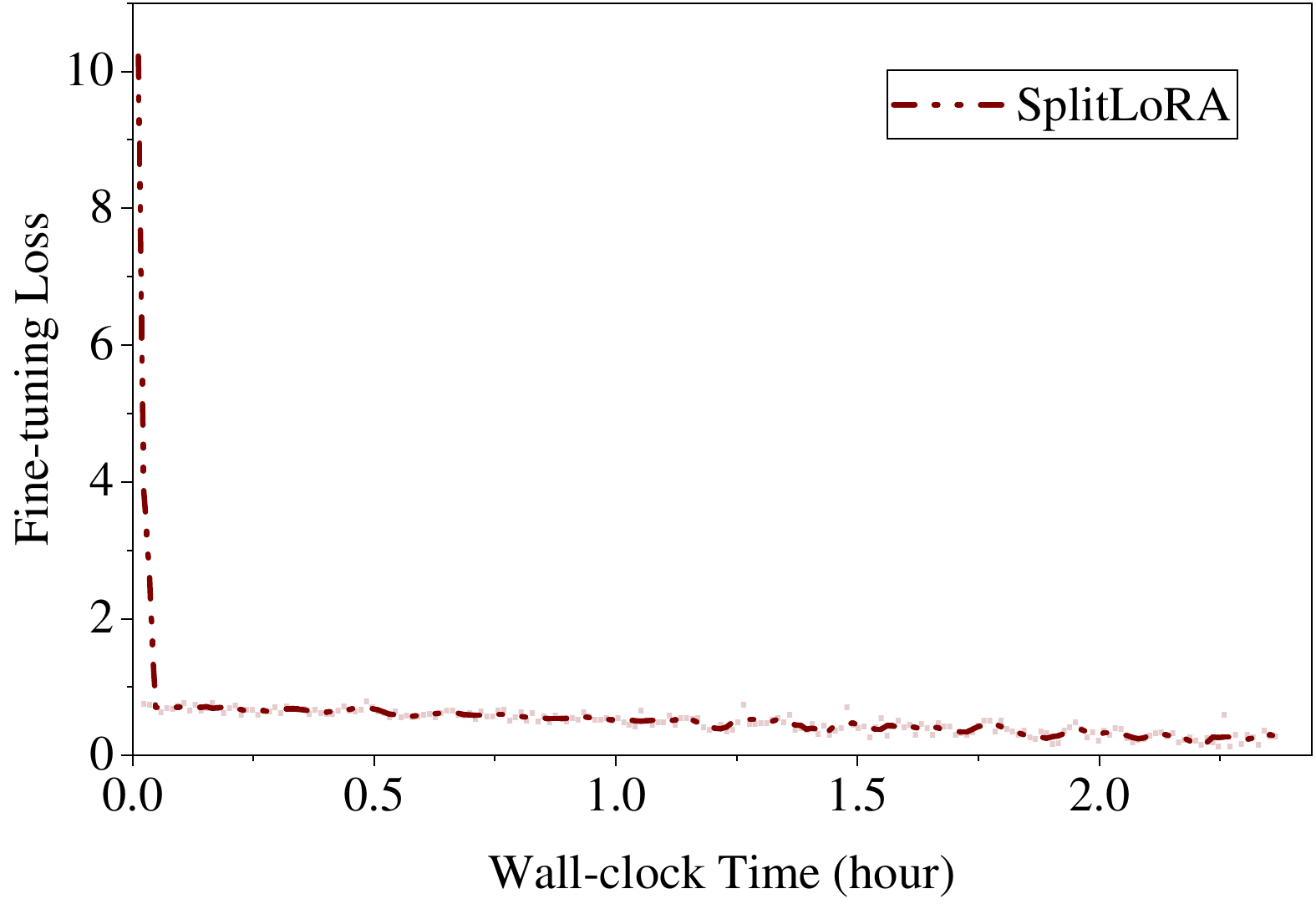}
    \caption{BoolQ}
    \label{fig:gemma1b_boolq_loss}
\end{subfigure}
\hfill
\begin{subfigure}[t]{0.48\textwidth}
    \centering
    \includegraphics[width=\textwidth]{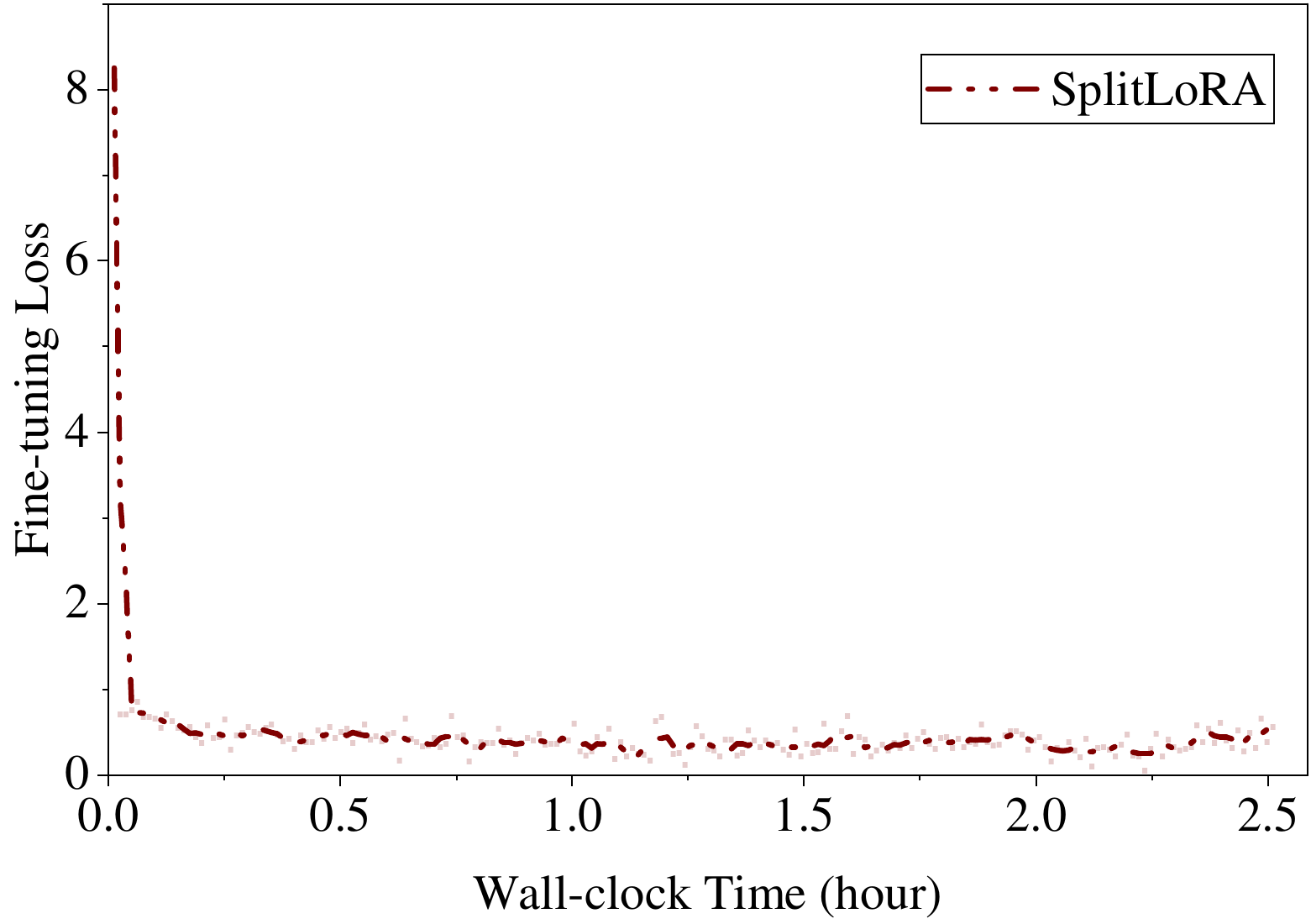}
    \caption{QNLI}
    \label{fig:gemma1b_qnli_loss}
\end{subfigure}

\vspace{2mm}

\begin{subfigure}[t]{0.48\textwidth}
    \centering
    \includegraphics[width=\textwidth]{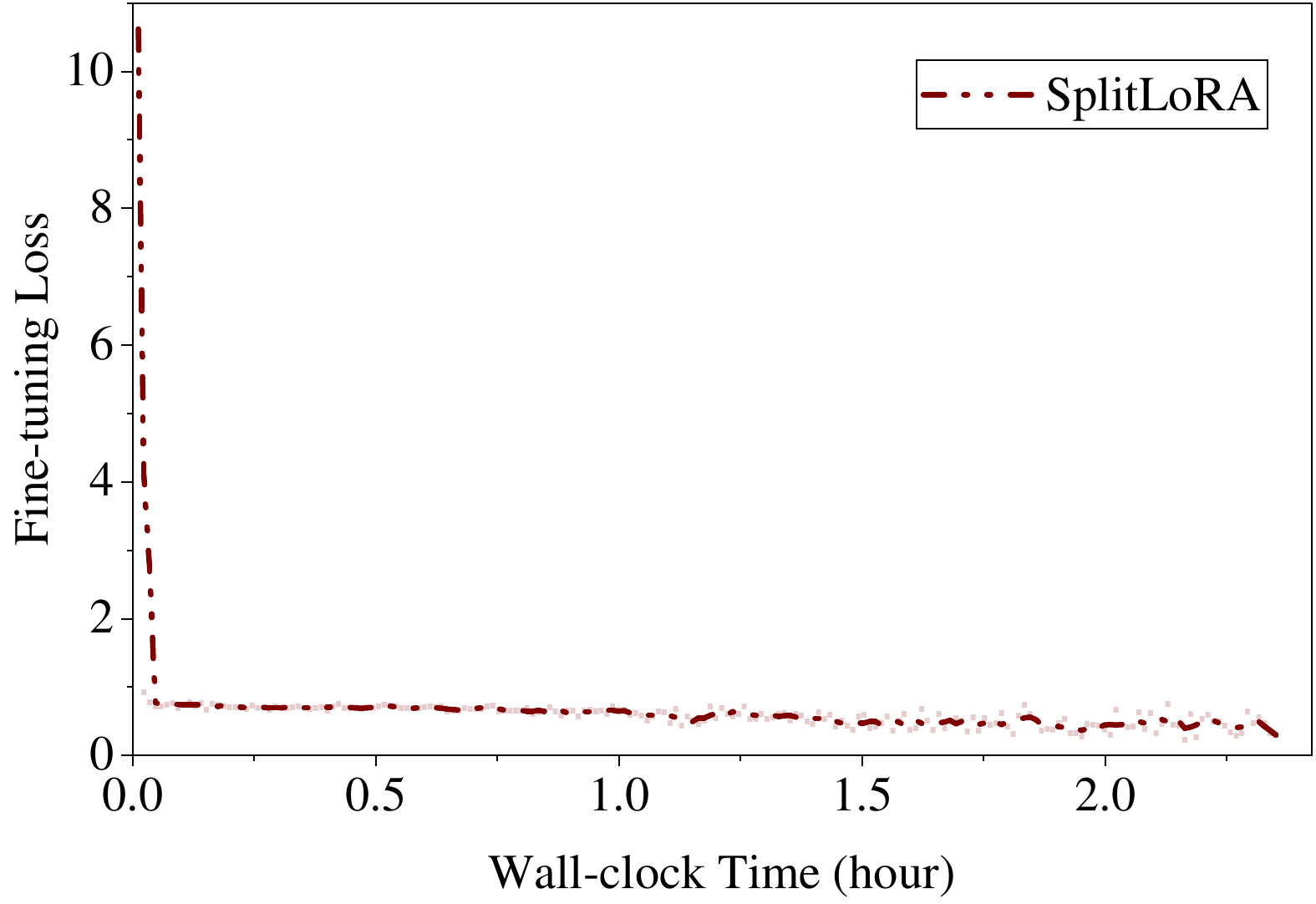}
    \caption{PIQA}
    \label{fig:gemma1b_piqa_loss}
\end{subfigure}
\hfill
\begin{subfigure}[t]{0.48\textwidth}
    \centering
    \includegraphics[width=\textwidth]{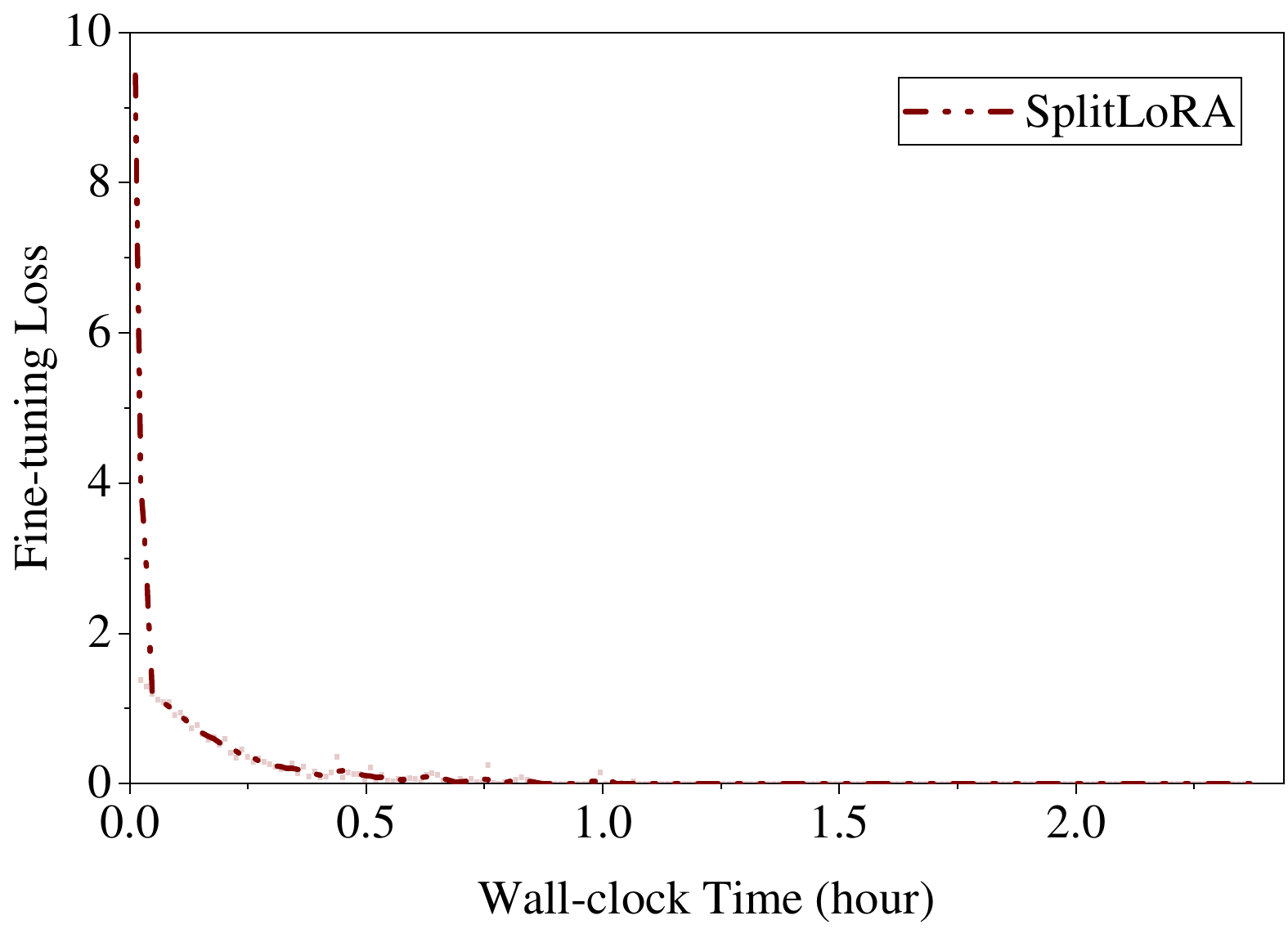}
    \caption{HellaSwag}
    \label{fig:gemma1b_hellaswag_loss}
\end{subfigure}

\vspace{2mm}

\begin{subfigure}[t]{0.48\textwidth}
    \centering
    \includegraphics[width=\textwidth]{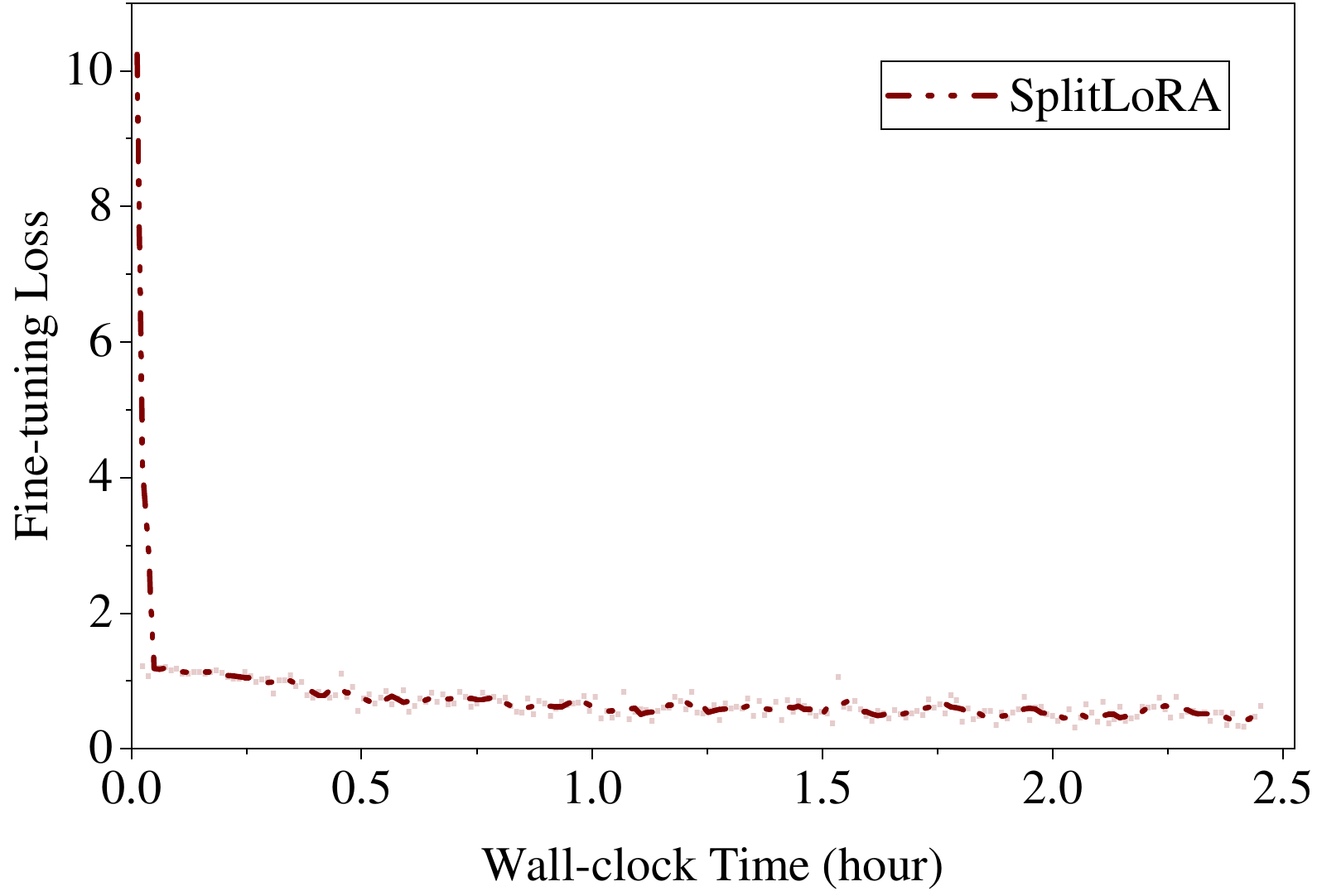}
    \caption{SocialIQA}
    \label{fig:gemma1b_socialiqa_loss}
\end{subfigure}
\hfill
\begin{subfigure}[t]{0.48\textwidth}
    \centering
    \includegraphics[width=\textwidth]{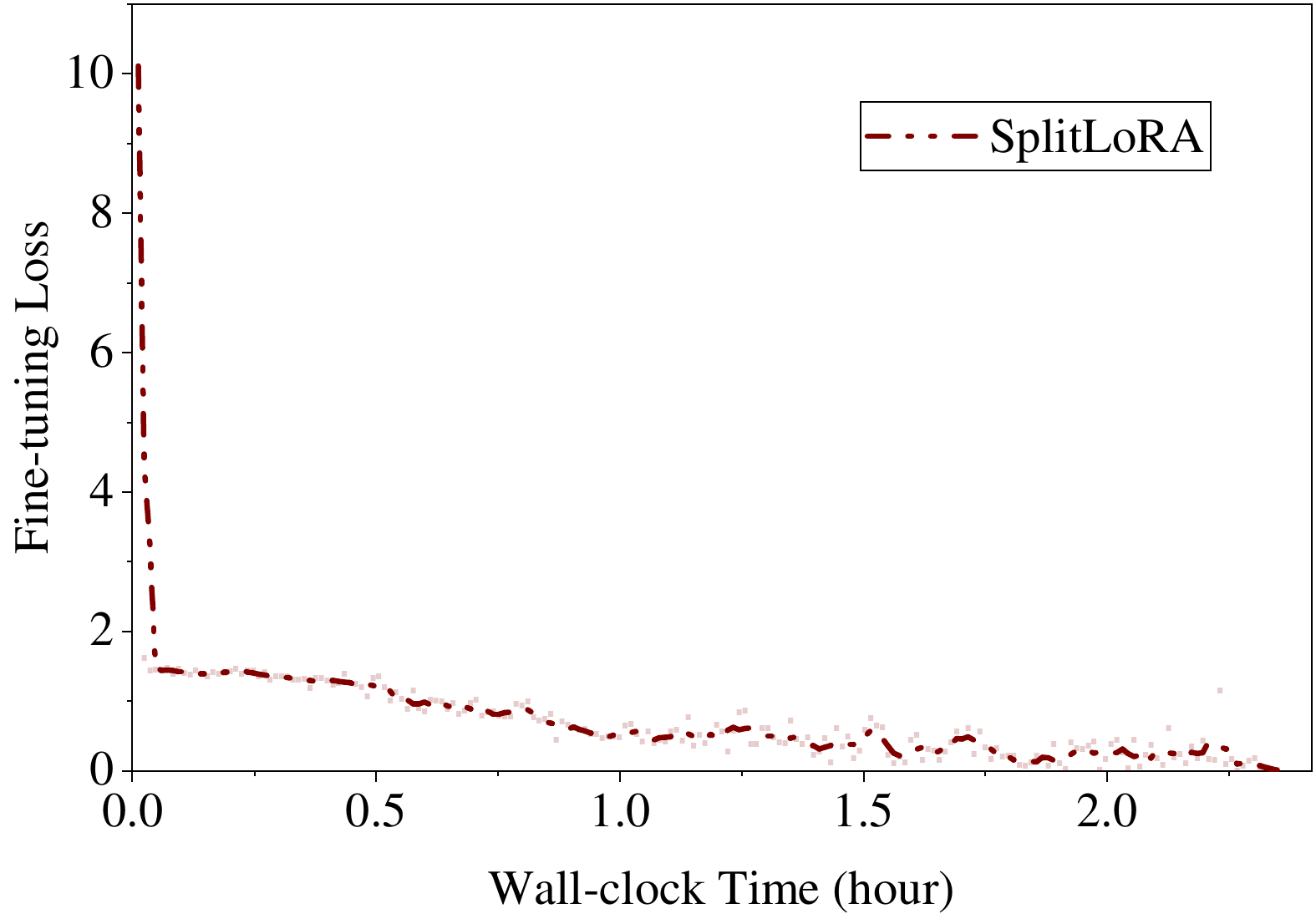}
    \caption{ARC-E}
    \label{fig:gemma1b_arce_loss}
\end{subfigure}

\vspace{2mm}

\begin{subfigure}[t]{0.48\textwidth}
    \centering
    \includegraphics[width=\textwidth]{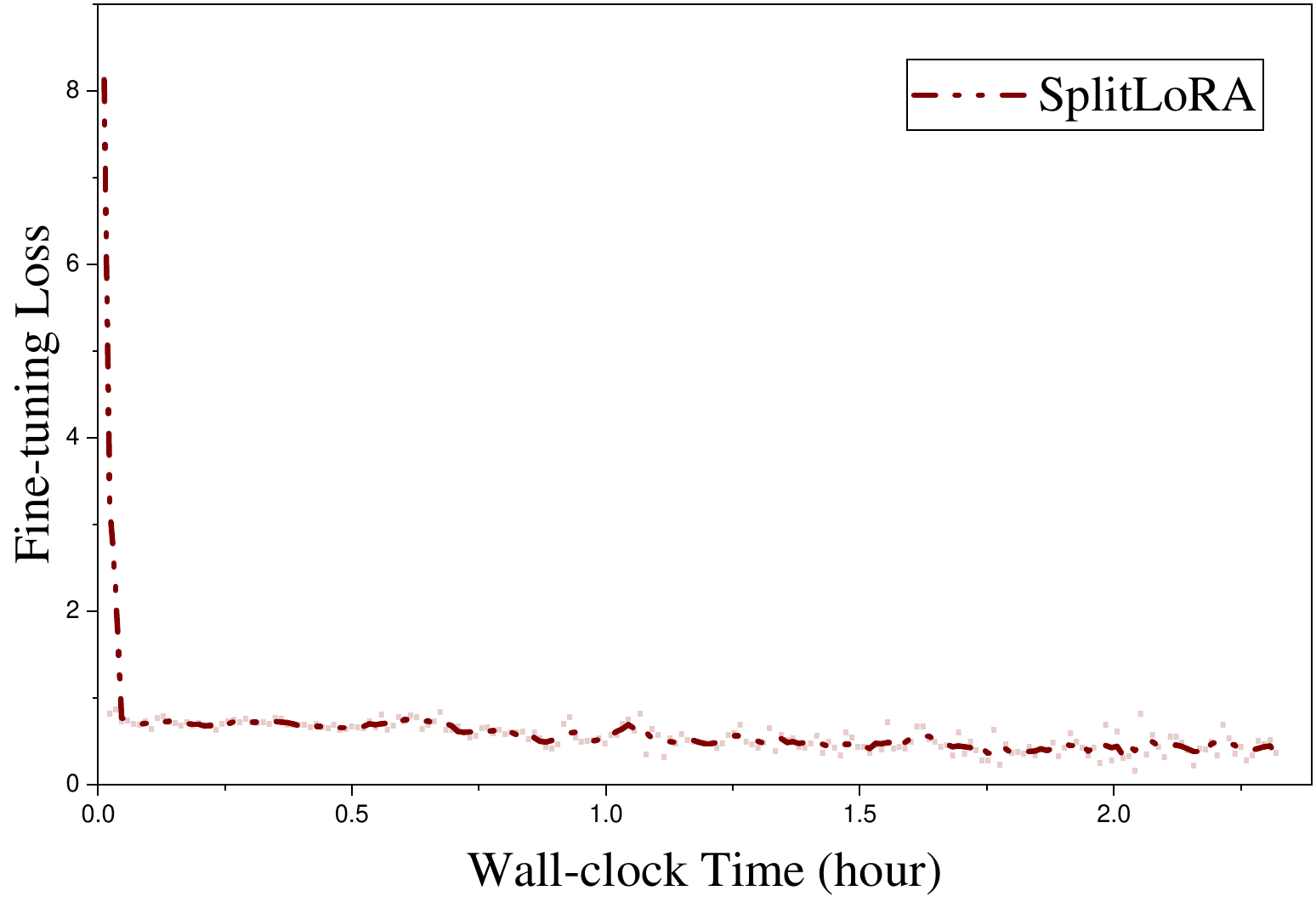}
    \caption{WinoGrande}
    \label{fig:gemma1b_winogrande_loss}
\end{subfigure}

\caption{Training loss curves of the federated fine-tuning methods on Gemma 3-1B across seven datasets under the experimental settings of Protocols A and B. The curves show the evolution of training loss with respect to wall-clock time.}
\label{fig:gemma1b_loss_curves}
\vspace{-4mm}
\end{figure*}


\end{document}